\documentclass[10pt, letterpaper, twoside]{article}

\usepackage[margin=1.25in]{geometry}

\usepackage{pratik}
\usepackage{times}

\def \mid {\,|\,}

\def \ehn {\hat{e}(h_w, D_n)}
\def \lhn {\Breve{e}(h_w, D_n)}
\def \eh {e(h_w)}
\def \eqn {\hat{e}(Q, D_n)}
\def \lqn {\Breve{e}(Q, D_n)}
\def \eq {e(Q)}

\def \kl {\text{kl}}
\def \KL {\text{KL}}

\def \Lbar {{\bar{\L}_w}}
\def \lbar {\bar{\l}}

\def \spec {\text{spec}}
\def \ev {\text{Ev}}

\def \tr {\text{tr}}

\graphicspath{{./fig/}}
\usepackage{pythonhighlight}
\usepackage[round]{natbib}
\usepackage{float}

\usepackage[color=gray!30,textsize=tiny,disable]{todonotes}

\makeatletter
\DeclareMathSizes{\@xpt}{9}{7}{5}
\makeatother

\def \mytitle {Does the Data Induce Capacity Control in Deep Learning?\xspace}

\title{\mytitle}
\author[1]{Rubing Yang}
\author[1]{Jialin Mao}
\author[2,3]{Pratik Chaudhari}

\affil[1]{\normalsize Applied Mathematics and Computational Sciences, University of Pennsylvania\vspace*{0.25em}}
\affil[2]{\normalsize Electrical and Systems Engineering, University of Pennsylvania\vspace*{0.25em}}
\affil[3]{\normalsize Computer and Information Science, University of Pennsylvania\vspace*{0.25em}}
\affil[ ]{\normalsize Email: \href{mailto:rubingy@sas.upenn.edu}{rubingy@sas.upenn.edu}, \href{mailto:jmao@sas.upenn.edu}{jmao@sas.upenn.edu}, \href{mailto:pratikac@seas.upenn.edu}{pratikac@seas.upenn.edu}}
\date{}

\begin{document}





\maketitle

\begin{abstract}
We show that the input correlation matrix of typical classification datasets has an eigenspectrum where, after a sharp initial drop, a large number of small eigenvalues are distributed uniformly over an exponentially large range. This structure is mirrored in a network trained on this data: we show that the Hessian and the Fisher Information Matrix (FIM) have eigenvalues that are spread uniformly over exponentially large ranges. We call such eigenspectra ``sloppy'' because sets of weights corresponding to small eigenvalues can be changed by large magnitudes without affecting the loss. Networks trained on atypical datasets with non-sloppy inputs do not share these traits and deep networks trained on such datasets generalize poorly. Inspired by this, we study the hypothesis that sloppiness of inputs aids generalization in deep networks. We show that if the Hessian is sloppy, we can compute non-vacuous PAC-Bayes generalization bounds analytically. By exploiting our empirical observation that training predominantly takes place in the non-sloppy subspace of the FIM, we develop data-distribution dependent PAC-Bayes priors that lead to accurate generalization bounds using numerical optimization.
\footnote{Proceedings of the 39$^{\text{th}}$ International Conference on Machine Learning, Baltimore, Maryland, USA. Copyright 2022 by the authors.}
\end{abstract}


\section{Introduction}
\label{s:intro}

\begin{figure}[!htpb]
\centering
\includegraphics[width=0.72\linewidth]{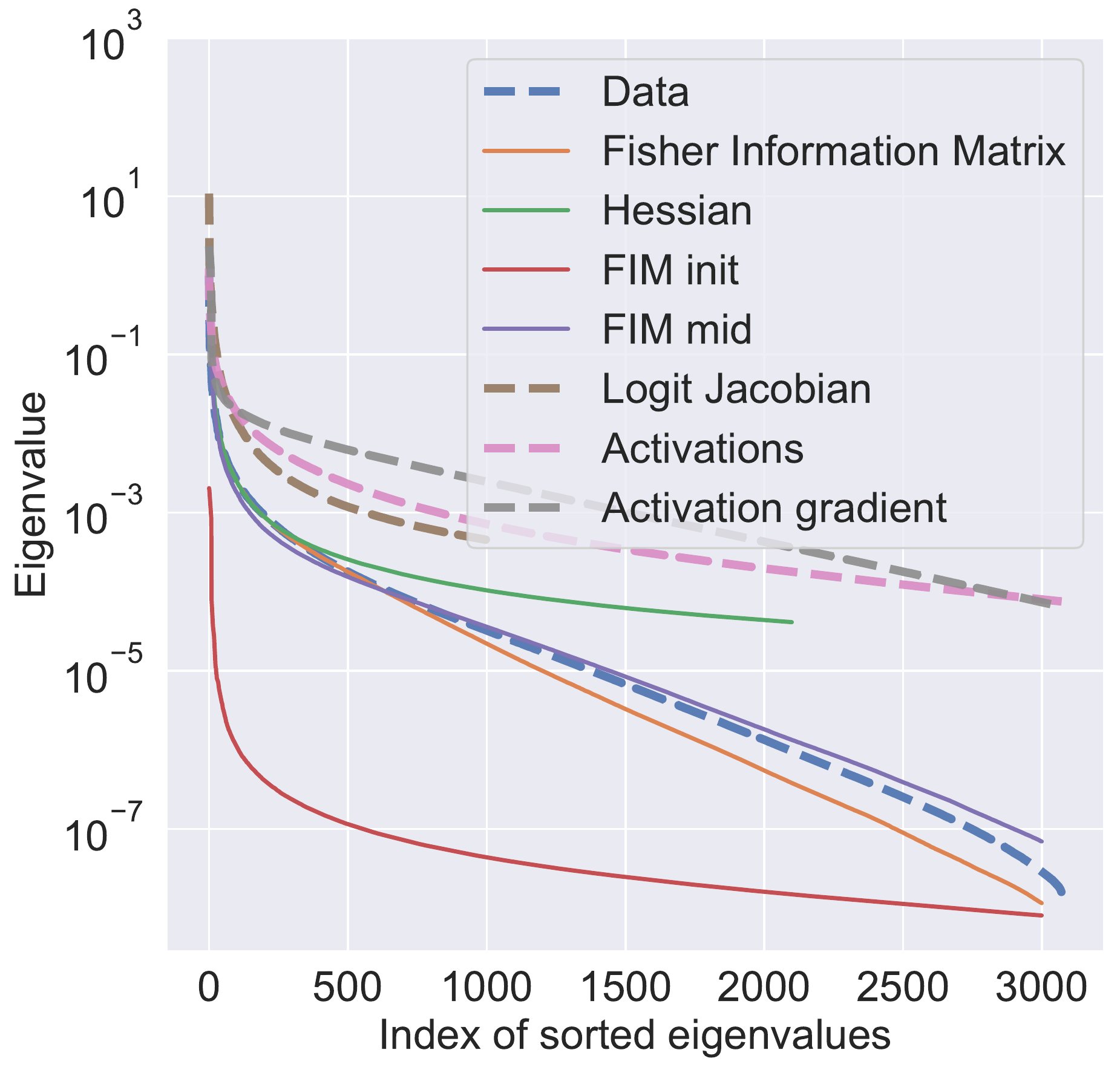}

\includegraphics[width=0.46\linewidth]{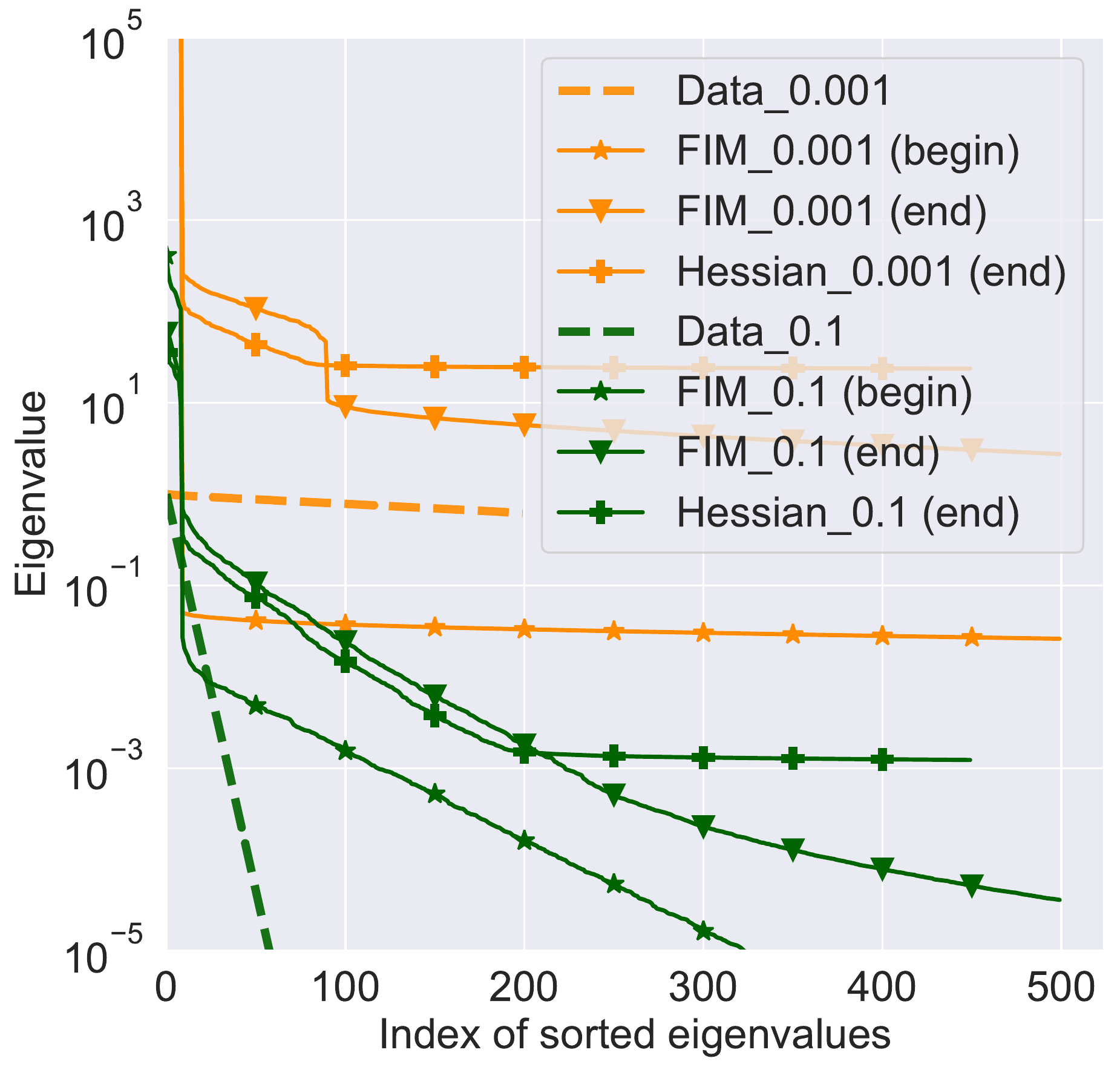}
\includegraphics[width=0.46\linewidth]{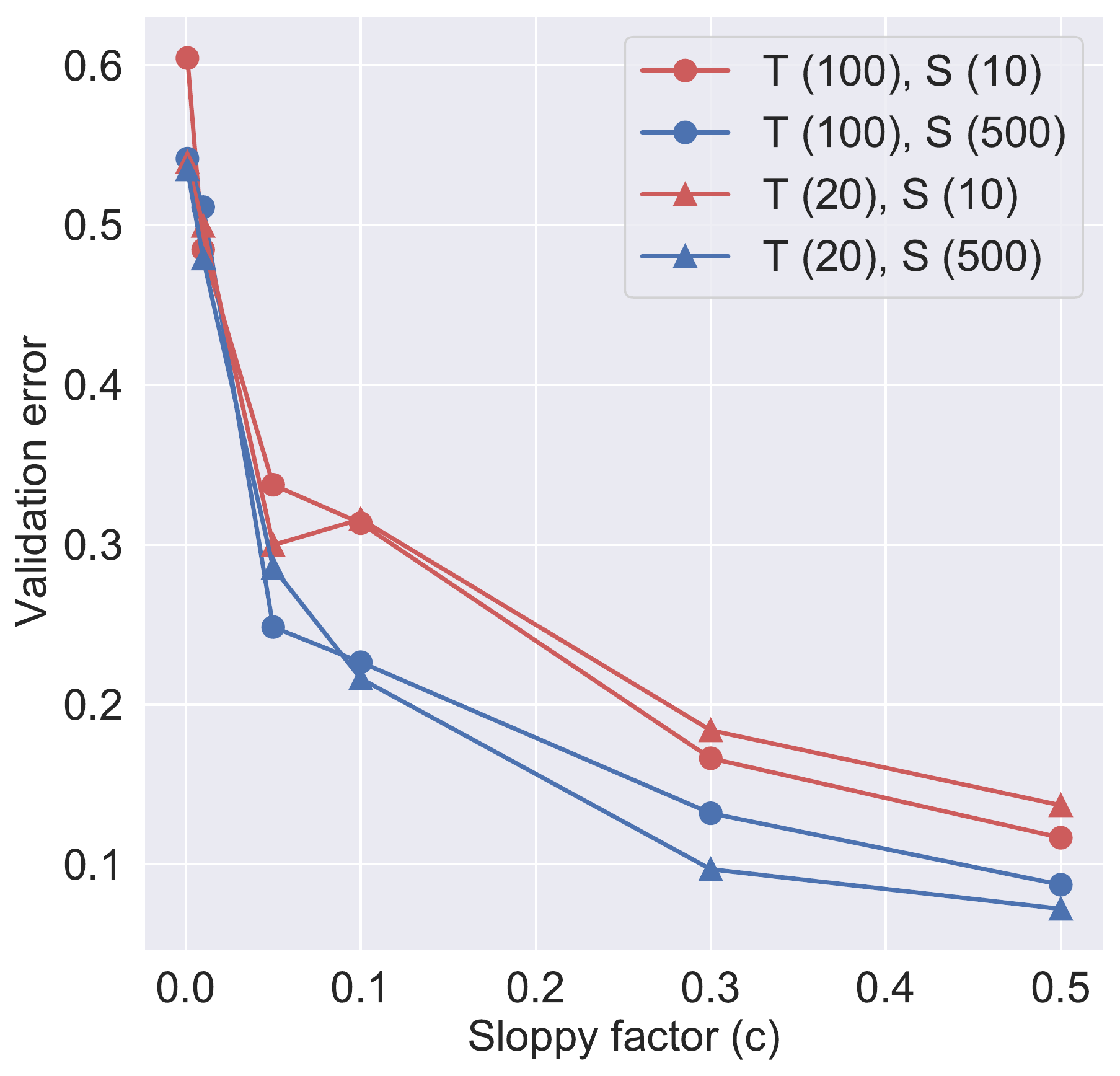}
\caption{
\textbf{Top:} Eigenspectra of the correlations of the inputs, activations and activation gradients, logit Jacobians and the FIM, Hessian at the end of training; for FIM we also calculate the spectra at initialization and middle of training. All eigenspectra are scaled by the largest eigenvalue of the input correlations (activation gradients are scaled up by $10^{12}$). Eigenspectra corresponding to activations/activation gradients of all layers of the network, and logit Jacobians of all logits are very similar (see~\cref{s:app:further_expt}). Eigenspectra of these quantities are also qualitatively the same at initialization, at the middle of training (see~\cref{s:app:eig_mid} \cref{fig: eig_mid}). This plot is drawn for a wide residual network with 10 layers on CIFAR-10 (WRN-10-8~\citet{zagoruyko2016wide}), eigenspectra of other networks/datasets are qualitatively the same (see~\cref{fig:eig_mnist_fc_600_2,s:app:further_expt}).\\
\textbf{Bottom Left:} Eigenspectra of the input correlation matix, FIM and Hessian at begining and end of training for sloppy factor (slope of the sloppy eigenvalue decay) $c = 10^{-3}$ (orange) and $c=10^{-1}$ (green). If inputs are not sloppy (small $c$) then even if there is a sharp drop after the top few eigenvalues (around 100 for orange lines), the eigenspectrum is flat. In comparison, the FIM/Hessian decay by about 3 orders of magnitude for $c=0.1$. The details of the experiments can be found at ~\cref{s:app:setup}.\\
\textbf{Bottom Right:} Validation error of a student (S) network on synthetic datasets of different sloppiness (X-axis) labeled by a teacher network (T). Numbers in brackets indicate number of hidden neurons in two-layer teachers/students. All students in this plot interpolate the training data perfectly. For non-sloppy inputs, interpolation leads to poor generalization, whereas interpolation is not detrimental to generalization for sloppy inputs. As the number of student neurons increases, fixed the teacher's size and the sloppiness factor, the validation error is better. Fixed teacher size, say 20, if inputs are sloppier (sloppy factor of 0.5 vs. 0.1) then we can generalize---roughly equally well---even if the student is smaller (10 vs. 500).
}
\label{fig:intro}
\end{figure}

In~\cref{fig:intro} (top), for a wide residual network (with 10 layers) on CIFAR-10, we calculated the eigenspectrum of the input correlation matrix ($n^{-1} X X^\top$ where each column of $X$ is one input image) and compared it to the eigenspectra of the Fisher Information Matrix (FIM) and the Hessian. We find that this decay pattern for the input correlation matrix is mirrored in that of the FIM and the Hessian. There are very few (less than 5\% of the input dimensionality) large eigenvalues (stiff) after which there is a sharp drop and a long tail of small eigenvalues (we call them sloppy as defined in \cref{def:sloppy}). Other quantities, e.g., correlations of activations of different layers, Jacobians of different logits with respect to the weights, and gradients of the loss with respect to activations of different layers, have a similar decay pattern. Eigenvalues span exponentially large ranges---about 7 orders of magnitude in this experiment. Sloppy eigenvalues are distributed uniformly across such exponentially large ranges.

Eigenspectra of many typical datasets and networks are similar. In~\cref{fig:intro} bottom, we created synthetic inputs with varying slopes for the decay of sloppy eigenvalues. We labeled such inputs using a teacher network with randomly generated, but fixed, weights and trained different student networks on such datasets. Each student was trained to have zero training error, i.e., it interpolated its training dataset perfectly. We find in~\cref{fig:intro} (left) that, again, the decay pattern of the inputs is mirrored in the FIM/Hessian of the students---sloppier the inputs, sloppier the FIM and the Hessian. Sloppier the input correlations, better the generalization error of the student (\cref{fig:intro} bottom right).

The Hessian governs the local geometry of the loss function in the weight space; small eigenvalues correspond to directions along which the loss is insensitive to changes in the weights. The FIM governs the local geometry in the prediction space; if we think of a network as a parameterized distribution $p_w(y \mid x)$, eigenvectors corresponding to small eigenvalues of the FIM correspond to sets of weights which can be changed significantly without affecting the distribution $p_w(y \mid x)$ much. A sloppy eigenspectrum for these matrices indicates that the trained network is in some sense, ``simple'': few sets of weights dominate its predictions while there exists a large set of sets that improve the predictions marginally. Both these matrices play a role in determining the generalization error of a neural network.

This paper investigates how sloppiness of the inputs causes the sloppiness of the FIM and the Hessian and how such sloppiness aids generalization.

\subsection{Contributions}

\begin{enumerate}[(1), nosep, wide, labelindent=0ex, itemsep=0.1em]
\item We show that \textbf{for typical datasets and deep networks, eigenspectra of correlation matrices of the inputs, activations of different layers, Jacobian of logits with respect to the weights, gradients of the loss with respect to the activations, as also the Hessian and the FIM, are sloppy}. These eigenspectra consists of few large eigenvalues and a large number of small eigenvalues that are distributed uniformly across an exponentially large range. We call such eigenspectra (or the corresponding quantities) ``sloppy'' and define this notion in~\cref{def:sloppy}. Synthetic datasets can be constructed where these quantities are not sloppy; interpolating networks do not generalize well for such datasets. We prove that (a) the trace of the correlation of the activations, logit Jacobians, Hessian and the FIM can be upper bounded by the trace of the input correlation matrix, (b) if we assume that the activations are sloppy then the eigenspectrum of a block-diagonal approximation of the FIM is also sloppy, (c) under the assumption of weights with bounded norm, eigenvalues of activations decays faster than $\mathcal{O}(1/i)$.

\item For a Gaussian isotropic prior $N(w_0, \e^{-1} I)$ centered at the initialized weights $w_0$, we calculate the optimal covariance of a Gaussian posterior $N(w, \S_q)$ (where $w$ are weights of the trained network) that minimizes a PAC-Bayes generalization bound. If the Hessian at $w$ is sloppy, then we obtain a non-vacuous generalization bound. For example, for MNIST, we get a bound of 32.4\% for a fully-connected network and 5.7\% for LeNet. This indicates that sloppiness of inputs controls the capacity of the model. \textbf{To our knowledge, this is the only analytical, non-vacuous generalization bound for deep networks that does not use weight compression.}


\item We characterize the \textbf{effective dimensionality of a deep network as}
\[
    \textstyle p(n, \e) = \sum_{i=1}^p \ind{\abr{\l_i} > \f{\e}{2 (n-1)}},
\]
where $\e$ is the inverse covariance of the PAC-Bayes prior and $n$ is the number of samples. Roughly speaking, $\e/(2 (n-1))$ is the elbow of the eigenspectrum in~\cref{fig:intro} (top); eigenvalues of the optimal PAC-Bayes posterior beyond this threshold are dominated by the complexity term in a PAC-Bayes bound while eigenvalues before this threshold are dominated by the training error. \textbf{For sloppy eigenspectra, this dimensionality is typically a tiny fraction of the number of weights}, e.g., it is less than 0.5\% of the number of weights for all networks/datasets considered in this paper, and much smaller than, say the VC-dimension.

\item We find that the \textbf{stiff sub-space of the FIM at initialization has a strong overlap with its counterpart at the end of training, and weight updates during training primarily happen in this stiff subspace}. We exploit this observation to numerically compute a PAC-Bayes bound using a Gaussian prior whose covariance is proportional to the FIM and a Gaussian posterior whose eigenvectors are the same as those of the FIM at initialization. This is a remarkably accurate estimate of generalization gap, e.g., for LeNet on MNIST, it is 0.9\% whereas the gap is about 0.5\%.
\end{enumerate}

All the code for experiments in this paper is provided at \href{https://github.com/grasp-lyrl/sloppy}{https://github.com/grasp-lyrl/sloppy}.


\section{Background}
\label{s:background}

Consider a dataset $D_n = \cbr{(x_i, y_i)}_{i=1}^n$ with $n$ samples, $x_i \in X \subset \reals^d$ and $y_i \in Y = \cbr{1,\ldots, m}$. We assume that this dataset is drawn from a joint distribution $D$ on $X \times Y$. A classifier $h_w: X \mapsto [0,1]^m$ parameterized by weights $w \in \reals^p$ belongs to a hypothesis space $\cbr{h_w: w \in \reals^p}$; this classifier maps inputs $x \in X$ to $m$-dimensional categorical distributions $p_w(y \mid x) \in [0,1]^m$. Let $Q$ be a distribution on hypotheses, which is implicitly a measure on $\reals^p$. We define
\begin{enumerate}[(a), nosep]
    \item training error of a hypothesis $\ehn = \f 1 n \sum_{i=1}^n {\bf 1} \{y_i \neq \argmax_y(p_w(y \mid x_i)) \}$;
    \item population error $\eh = \E_{D_n \sim D^n} \sbr{\ehn}$;
    \item training loss is $\lhn = -\f 1 {n\log (2)} \sum_{i=1}^n \log p_w(y_i \mid x_i)$;
    \item empirical error and loss of the distribution $Q$ of hypotheses $\eqn = \E_{w \sim Q} \sbr{\ehn}$ and $\lqn = \E_{w \sim Q} \sbr{\lhn}$, respectively;
    \item population error of distribution $Q$ given by $\eq = \E_{D_n \sim D^n} \sbr{\eqn}$; and
    \item population loss is $\Breve{e}(Q) = \mathbb{E}_{D_n\sim D^n}[\Breve{e}(Q, D_n)]$.
\end{enumerate}

\paragraph{Hessian and Fisher Information Matrix (FIM)} The Hessian $H \in \reals^{p \times p}$ is the second derivative of the empirical loss with respect to the weights $w$, i.e., $ H_{ij} = \partial_i \partial_j \lhn$. The Fisher Information Matrix (FIM) $F \in \reals^{p \times p}$ has entries
\beqs{
    F_{ij} = \f 1 n \sum_{k=1}^n \sum_{y=1}^m p_w(y \mid x_k) \partial_i \log p_w(y \mid x_k) \partial_j \log p_w(y \mid x_k).
    \label{eq:fim}
}
It is important to note the expectation over the outputs $y$. The empirical FIM is an approximation of the FIM where one sets $y = y_k$. Both the Hessian and FIM are large matrices and it is difficult to compute them for modern deep networks. Therefore some of our experiments use a Kronecker-factor approximation  ~\citep{martensOptimizingNeuralNetworks2016} of a block diagonal Hessian and FIM where cross-terms $\partial_i \partial_j$ across different layers of a deep network are set to zero.

\subsection{PAC-Bayes Generalization Bounds}
\label{s:pac_bayes}

The PAC-Bayesian framework developed in~\citet{langford2001bounds,mcallester1999pac} allows us to estimate the population error of a randomized hypothesis with distribution $Q$ using its empirical error and its Kullback-Leibler (KL) divergence with respect to some prior distribution $P$. For any $\delta > 0$,  with probability at least $1-\delta$ over draws of the dataset $D_n$, we have
\beq{
    \kl(\eqn, \eq)  \leq \f{\KL(Q, P) + \log(n/\delta)}{(n-1)},
    \label{eq:pac_bayes}
}
where $\KL(Q, P) = \int \dd{Q(w)} \log(\dd{Q}/ \dd{P})(w)$. We will also define a KL divergence between two Bernoulli random variables with parameters $b, a$ as $\kl(b, a) = b \log(b/a) + (1-b) \log((1-b)/(1-a))$. The right hand-side of this inequality can be minimized to compute a distribution $Q$ that has a small generalization error~\citep{langford2002not,dziugaiteComputingNonvacuousGeneralization2017}. Typically, we pick a simple form for distributions $Q$ and $P$, say Gaussian. We can also have hyper-parameters for the prior $P$, say the scale $\e$ of the covariance of $P$ and search over this scale while optimizing the bound. See~\cref{s:app:effective_dim} for details.

\subsection{Data-dependent PAC-Bayes Priors}

The posterior $Q$ in~\cref{eq:pac_bayes} may depend upon the training samples $D_n$, e.g., it could be the distribution on the weight space induced by a randomized training algorithm like stochastic gradient descent (SGD). The prior $P$ can depend upon the data distribution $D$, but not the samples $D_n$ themselves. Although it is common to use priors that do not depend upon the data at all, it is has been increasingly noticed that data-distribution dependent priors may provide tighter bounds~\citep{dziugaite2018data}. To gain intuition, recall that in the expression for the KL-divergence between two Gaussians $Q = N(w, \S_q)$ and $P = N(w_0, \S_p)$, we have a term of the form $(w-w_0)^\top \S_p^{-1} (w-w_0)$ that depends upon the distance between trained weights $w$ and the initialization $w_0$. Priors $P$ that do not depend upon the data may therefore incur a large KL-term.

\noindent \textbf{FIM and Hessian-dependent priors}
We can pick a prior using a subset of the training samples~\citep{ambroladze2007tighter}, e.g., we can center the Gaussian prior on weights trained on this subset, to obtain a better PAC-Bayes bound---the theory allows this. Doing so leads to a worse denominator in~\cref{eq:pac_bayes}, although this may be mitigated by a smaller numerator. \citet{parrado2012pac} also define expectation-priors, i.e., where we choose a prior that depends on the data \emph{distribution} and, in practice, evaluate this prior using samples in the training dataset in lieu of the distribution. For example, PAC-Bayes theory allows both picking the prior covariance $\S_p$ to be $\S_p \propto F_{w_0}$ and $\S_p \propto \tilde{H}_{w_0}$ where $\tilde{H}$ is the Gauss-Newton approximation of the Hessian. But while we may use all the  samples to compute the FIM, we should compute the Hessian on a separate subset of the data.

\section{Theoretical Results}
\label{s:methods}

We prove how sloppiness in the Hessian and the FIM is related to sloppiness of the correlations of the activations (\cref{s:sloppy_input_sloppy_fim_hessian}) and the inputs (\cref{s:special_cases}). We then exploit sloppiness to compute PAC-Bayes generalization bounds (\cref{s:pac_bayes_bounds}) and develop an expression for the effective dimensionality of a deep network (\cref{s:effective_dimensionality}). We exploit sloppiness to get effective methods for optimizing PAC-Bayes bounds (~\cref{s:non_analytical_pac_bayes}). All proofs are provided in~\cref{s:app:proofs}. The theory in this section applies for general deep networks; we will remark when restrictions are in place.

\subsection{Sloppy Input Correlation Matrix Leads to a Sloppy FIM and Hessian}
\label{s:sloppy_input_sloppy_fim_hessian}

Consider a deep network with $L$ layers with weights $w = (w^0, w^1, \ldots, w^L)$. Activations of the $k^{\text{th}}$ layer are given by
\(
    h^k = \s (w^{k-1} h^{k-1}),
\)
and we set $h^0 \equiv x$. The non-linearity $\s$ acts element-wise upon its argument and we assume that it has a bounded derivative $\abr{\s'(x)} \leq a$ with $\s(x) = 0$ in which case $\abr{\s(x)} \leq a \abr{x}$; ReLU, leaky ReLUs and tanh satisfy this assumption. Preactivations (before nonlinearities) will be denoted by $u^k = w^{k-1} h^{k-1}$ for $k = 1, \cdots, L+1$, and for clarity, we use a special notation $z \equiv u^{L+1}$ to denote the logits of the network. The dimensionality of these quantities is $h^k \in \reals^{d_k}$, $w^k \in \reals^{d_{k+1} \times d_k}$ and $w^L \in \reals^{m \times d_L}$. The linear map represented by $w^{k}$ can model both fully-connected layers and convolutional layers.  For the sake of exposition, we set all the bias terms to zero.

\begin{theorem}[Trace of the FIM and Hessian are bounded by that of the input correlation matrix]
\label{lem:sloppy_input_sloppy_fim_hessian}
For any weights, the trace of the FIM $F_w$ and the Gauss-Newton approximation of the Hessian $\tilde{H}_w$ are both upper-bounded by
\beq{
\aed{
    2 ma^{2L} \tr \rbr{\mathbb{E}[xx^\top]} \prod_{j=0}^L \norm{w^j}_2^2 \rbr{\sum_{j=0}^L \norm{w^j}_2^{-2}}
    \label{eq:sloppy_input_sloppy_fim_hessian}.
    }
}
\end{theorem}

The Gauss-Newton approximation which neglects the so-called $H$ terms of the Hessian~\citep{papyan2019measurements} is good towards the end of training when the logits have a small entropy. For the FIM, the above bound is remarkable however, it indicates that the trace of FIM is controlled by that the input correlations and multiplicative terms that depend upon the $\ell_2$ norm of the weights.

We can also go beyond the trace and control the entire eigenspectrum. But this is difficult to do in general because both FIM and Hessian are a result of multiple nonlinear operations on the inputs. We therefore bound the eigenvalues of a block-diagonal approximation of the FIM in terms of the eigenvalues of the activations.

\begin{lemma}[Block-diagonal approximation of the FIM is sloppy if the activations are sloppy]
\label{lem:fim_sloppy}
Let $\spec(A)$ denote the eigenvalues $(\l_1(A), \ldots, \l_p(A))$ of a matrix $A$ in descending order. For a constant $c$, let $\spec(A) \preceq  c\ \spec(B)$ denote that $\l_i(A) \leq c \l_i(B)$ for all $i \leq p$. For any logit $z_i$, for all layers $k \leq L$, we have
\beq{
    \aed{
            \spec &\rbr{\E \sbr{\dv{z_i}{w^k} {\dv{z_i}{w^k}}^\top}} \preceq\ a^{2(L-k)} \prod_{j=k+1}^L \norm{w^j}_2^2
            \\
            &\qquad\spec(I_{d_{k+1}}) \otimes \spec\rbr{\E\sbr{h^k {h^k}^\top}},
        }
    \label{eq:fim_sloppy}
}
with $\prod_{j=L+1}^L \norm{w_j}_2^2 \equiv 1$. A similar result also holds for the sum of logits $\sum_{i=1}^m z_i$ as in ~\cref{lem:fim_sloppy} (see ~\cref{cor: fim_sloppy}). The proof of this lemma also shows that a block-diagonal approximation of the Gauss-Newton approximation of the Hessian is sloppy if the activations are sloppy.
\end{lemma}

This lemma indicates that the eigenspectrum of the block-diagonal approximation of the FIM (concatenation of the eigenspectra of different blocks) is controlled by the eigenspectrum of the activation correlations of different layers. Our experiments show that activations of all layers (except the logits) of a trained deep network are sloppy.

\subsection{Special Cases Where Sloppy Inputs Lead to Sloppy Activations and Thereby Sloppy FIM and Hessian}
\label{s:special_cases}

Although our experiments show that activations are sloppy if the inputs are, it seems rather difficult to prove in general. We therefore discuss two special cases where this holds. The first case is for a kernel machine with an inner product kernel while the second case assumes that the width of the network goes to infinity and weights remain bounded in $\ell_2$ norm.

\begin{remark}[Eigenspectrum of inner product kernel is controlled by that of its inputs]
Let $x_i \in \reals^{d}$ for $i \leq n$ be iid random vectors. \citet[Theorem 2.1]{Karoui2010TheSO} shows that the Gram matrix of an inner product kernel $M_{i, j} = f \rbr{\frac{x_i^\top x_j}{d}}$ for some function $f$ can be approximated by
\[
    \aed{
    K &= \rbr{f(0) + f''(0)\frac{\tr (\S_d^2)}{2d^2}}11^\top + f(0) \frac{X X^\top}{d} + v_{d} I_n\\
    \text{where }& v_d = f \rbr{\frac{\tr(\S_d)}{d}} - f(0) - f'(0)\frac{\tr (\S_d)}{d}.
    }
\]
More precisely $\norm{M-K}_2 \to 0$ in probability when $d, n \to \infty$ for a fixed ratio $d / n$. Note that $v_{d}$ is small when $\frac{\tr(\S_{d})}{d}$ is small. Hence, we can see that the eigenspectrum of $K$, and thereby $M$, is controlled directly by that of $XX^\top$.

Note that this argument cannot directly be used for a deep network because correlations of activations in the network are not an inner product kernel. But this indicates that even for such a kernel machine, sloppiness of the inputs leads to sloppiness of the FIM.

\end{remark}

\begin{remark}[Infinitely wide network with bounded weight norm]
\label{rem:inf_wide_activations}
If the $\ell_2$ norm of the weights is bounded, we show in~\cref{lem:bound_xcorr_norm} that
\[
    \tr \left(\E \sbr{h^k {h^k}^\top} \right) \leq a^2 \norm{w^{k-1}}_2^2 \tr \left( \E \sbr{h^{k-1} {h^{k-1}}^\top}\right).
\]
If we iterate upon this inequality down to the last layer to $\tr \E[x x^\top]$ on the right hand-side (which is a constant). If the width of the $k^{\text{th}}$ layer goes to infinity, for the trace to be summable, we have that the eigenvalues of $\E [h^k{h^k}^\top]$ decay faster than $\mathcal{O}(1/i)$.
\end{remark}


\subsection{Analytical Bound on Generalization (Method 1)}
\label{s:pac_bayes_bounds}

Consider a deep network trained to minimize the loss $\Breve{e}(h_{w}, D_n)$. Assume that $w$ is a local minimum of the objective and thus the Hessian $H_w$ is positive semi-definite. We can write $H_w$ as its orthonormal decomposition $H_w = U_w \L_w U_w^\top$ where $\L_w = \diag(\l_1, \ldots,\l_p)$ with eigenvalues $\l_1, \geq \cdots \geq \l_p \geq 0$ arranged in descending order. Consider a Gaussian posterior $Q = N(\mu_q, \S_q)$ with the mean $\mu_q = w$ fixed. We would like to compute the best $\Sigma_q$ that gives a tight PAC-Bayes bound.

We use a loose version of the bound $\eq \leq L(\S_q) := \Breve{e}(Q, D_n) + \KL(Q, P)/(2 (n-1))$ to simplify the analytical calculation and show in~\cref{s:app:lbar_4nm1} that
\aeq{
    \S_q &= U_w (\Lbar)^{-1} U_w^\top,
    \label{eq:S_q_basic}\\
    \text{where }\lbar_i &= 2(n-1) \l_i + \e\quad \forall i \leq p.
    \label{eq:lbar_4nm1}
}

This posterior gives a non-vacuous bound on the generalization error (as explained in \cref{s:def_sloppy}) and to our knowledge, this is the only analytical bound that is non-vacuous and does not use weight compression (e.g.,~\cite{zhou2018non}). For example, the bound for a fully-connected network on MNIST with one hidden layer of 600 neurons is 0.32 while the test error $\eq$ is $\approx 0.089$. For comparison, \citet{dziugaiteComputingNonvacuousGeneralization2017} \emph{numerically optimize}~\cref{eq:pac_bayes} to get a bound of 0.161.



\begin{remark}[PAC-Bayes posterior is more spread out along sloppy eigenvectors]
\label{rem:analytical_pac_bayes}
In~\cref{eq:lbar_4nm1}, we can think of the scaled prior inverse variance $\e/(2(n-1))$ as a threshold beyond which the sloppy eigenvalues of the Hessian $\l_i$ are small enough and the loss changes so little that the optimal PAC-Bayes posterior in~\cref{eq:pac_bayes} focuses on accurately capturing the prior's covariance to obtain a small KL-term. For eigenvalues above this threshold, e.g., the stiff eigenvalues, the optimal posterior has to ensure that the empirical loss is not large. We will see in~\cref{s:expt:pac_bayes} that this phenomenon also holds for cases when posteriors are optimized.
\end{remark}


\section{Effective Dimensionality of a Deep Network}

\subsection{Definition of Effective Dimensionality}
\label{s:effective_dimensionality}

\begin{figure}[htpb]
\centering
\includegraphics[width=0.75\linewidth]{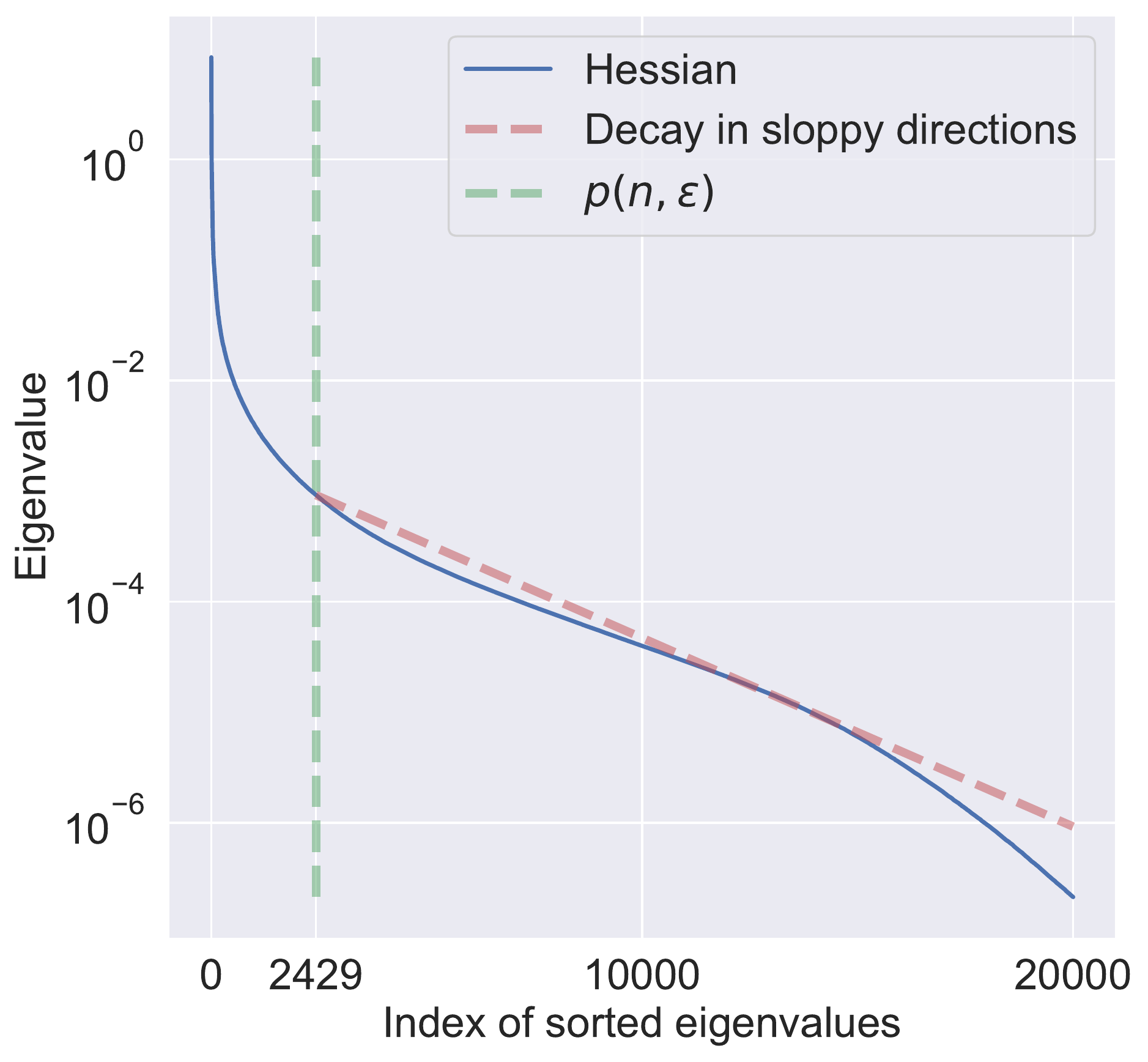}
\caption{
For two layer fully connected network (FC-600-2), we calculated the eigenspectrum (blue) of Kronecker-factored approximation of the Hessian at the mean of the posterior $Q$ (using numerical optimization of the PAC-Bayes bound using Method 3 in~\cref{s:non_analytical_pac_bayes} but that is not important at the present). The dimensionality $p(n, \e)$ (green) was calculated using the $\e$ obtained by the same procedure. The red line shows the linear decay of sloppy eigenvalues (slope is 0.0004). The green line is close to the elbow and effectively splits the stiff and sloppy eigenvalues.
}
\label{fig:effec_dim}
\end{figure}

Motivated by ~\cref{rem:analytical_pac_bayes}, we define the effective dimensionality for a deep network at weights $w$ as the number of eigenvalues of the Hessian $H_w$ with magnitude at least $\e/(2(n-1))$, i.e.,
\beq{
    \textstyle p(n, \e) = \sum_{i=1}^p \ind{\abr{\l_i(H_w)} \geq \f{\e}{2(n-1)}}.
    \label{eq:pne}
}
\cref{s:app:effective_dim_calculation} gives the calculation for why this is a good definition of the dimensionality. It indicates that the threshold $\e/(2(n-1))$ can be thought of as the ``elbow'' in the eigenspectra in~\cref{fig:intro} (top), which separates the stiff eigenvalues which decrease quickly and the sloppy eigenvalues. This gives an easy way to compute the effective dimensionality, e.g., for the purposes of model selection. This is also true if we use more sophisticated, numerical, methods for optimizing the PAC-Bayes bound as shown in~\cref{fig:effec_dim} and \cref{tab:eff dim}.

\begin{remark}[Why does the effective dimensionality depend on $\e$?]
\label{rem:effec_dim_eps}
Our definition in~\cref{eq:pne} may seem unusual because $\e$ is a user-chosen parameter but this is only an artifact of PAC-Bayes theory. As $\e \to 0$, the effective dimensionality converges to the number of weights $p$, but for non-zero values of $\e$, where the PAC-Bayes theory effectively restricts its predictions to a subset of the hypothesis space, this expression coupled with the analytical calculation in~\cref{eq:lbar_4nm1} may provide a useful way to perform model selection.
\end{remark}

\begin{remark}[Why does the effective dimensionality depend on $n$?]
\label{rem:effec_dim_n}
The fact that $p(n,\e)$ depends upon $n$ is reminiscent of the Bayesian Information Criterion (BIC) where the the model complexity term scales with $\log n$~\citep{schwarz1978estimating}. The dependence on $n$ in our cases also arises for similar reasons, from a balance between the training error $\eqn$ and the KL-term in~\cref{eq:pac_bayes}. As $n \to \infty$, we see that $p(n,\e) \to p$. This is because for inputs with sloppy dimensions the model needs to capture \emph{all} the dimensions to predict accurately.
\end{remark}

{
\renewcommand{\arraystretch}{1}
\begin{table}
\centering
\tiny
\rowcolors{1}{}{black!5}
\resizebox{0.7\linewidth}{!}{

\begin{tabular}{lrrr}
\toprule
Model & \#weights ($p$) & $p(n,\e)/p$ (\%) \\
\midrule
FC-600-1 & 472,202 &0.487 \\
FC-600-2 & 832,802 &0.292 \\
FC-1200-1 & 944,402 &0.245 \\
FC-1200-2 & 2,385,602 &0.095 \\
LeNet & 44,429 &0.184 \\
Synthetic ($c=10^{-1}$) & 211,010 &0.256 \\
Synthetic ($c=10^{-3}$) & 211,010 & 0.820 \\
\bottomrule
\end{tabular}
}
\caption{\textbf{Effective dimensionality of different models} calculated using the $\e$ and Hessian from Method 3 in~\cref{s:non_analytical_pac_bayes} for different networks on MNIST (except last two rows which use fully-connected networks for synthetic datasets created in~\cref{fig:intro} with different slopes of the eigenspectra $c$). We see that in all cases, $p(n,\e)$ is a very small fraction of the number of weights.}
\label{tab:eff dim}
\end{table}
}



\subsection{Definition of Sloppiness}
\label{s:def_sloppy}

We next build upon~\cref{s:effective_dimensionality} to define sloppiness.

\begin{definition}[Strength factor and sloppy factor]
\label{def:sloppy}
Let $\l_i(A)$ denote eigenvalues of a positive semi-definite matrix $A \in \reals^{p\times p}$ in descending order $\l_1 \geq \cdots \geq \l_p$. The strength factor for a model with effective dimensionality $p(n,\e)$ at a local minimum $w$ (where $H_w$ is positive semi-definite) is defined to be
\beq{
    \textstyle s(n, \e) = \sum_{i=1}^{p(n, \e)} 1+\log \rbr{\frac{2(n-1) \l_i(H_w)}{\e} + 1}.
    \label{eq:strength}
}
The strength factor characterizes the stiff eigenvalues of the eigenspectrum. For a matrix $A$, the sloppy factor for such a model at index $r$ is defined to be
\beq{
    c(A, r) = \sup \{c' \geq 0 : \l_i(A) \leq \l_r(A) e^{-c'(i-r)} \forall i \geq r \geq 1 \}
    \label{eq:def_sloppy}
}
This definition implicitly means that the small eigenvalues beyond $\l_r(A)$ are uniformly distributed across an exponentially large range $(\l_r, \l_p)$ if $c(A, r) > 0$. We will be primarily interested in setting the index $r$ to be simply $p(n,\e)$. Note that sloppiness is a phenomenon pertaining to the \emph{non-zero eigenvalues} of a matrix and is relevant even if the matrix is singular, e.g., the FIM loses rank for non-identifiable models like deep networks~\citep{amari2002geometrical}.
\end{definition}

\textbf{How do the strength and sloppy factor affect generalization?}
Let us simplify notation to write the sloppy factor as $c(n, \e) \equiv c(H_w, p(n,\e))$. Under the assumption that the $c(n,\e)$ is non-negative, when the training error $\breve{e}(h_w, D_n)$ is close to zero, we show in~\cref{s:app:effective_dim} a loose version of PAC-Bayes bound $\lqn + \KL(Q, P)/(2 (n-1))$ (this was also used in Method 1 in~\cref{s:pac_bayes_bounds}) is
\beq{
    \frac{s(n, \e) + 2/c(n, \e) + \e \norm{w-w_0}_2^2}{4(n-1)}.
    \label{eq:effective_dim_order}
}
Thus, the strength and sloppy factor together determine the generalization performance. If the Hessian $H_w$ is sloppy, then the effective dimensionality $p(n, \e)$ is small. This ensures that both $s(n, \e)$ and $1/c(n, \e)$ are small compare to $n$. The third term $\e \norm{w-w_0}^2$ comes from the the fact that the mean of $P$ and $Q$ are different. It is typically not large compared to $n$. For example, for a two-layer fully-connected network on MINST, $p(n, \e) = 2429$, $s(n, \e) = 6810$, $1/c(n, \e) = 2545$, and $\e\norm{w-w_0}^2 = 8526$, with $n=55000, \e = 101.3$). For comparison, if we have an isotropic Hessian $\l_i \equiv \l$, either $s(n, \e)$ or $1/c(n, \e)$ will be $\OO(p)$ and $p$ is about 0.8 million.

This suggests that \textbf{even if the hypothesis class of deep networks is very large, sloppiness of $H_w$, which is inherited from sloppiness of the input data, restricts the set of hypotheses that the trained model belongs to}. The three quantities that we have defined here $p(n, \e)$, $s(n, e)$ and $c(n, \e)$ together help understand this phenomenon; see~\cref{s:app:pac_bayes_full} their values for other models.

\section{Numerical Methods to Compute PAC-Bayes Bounds}
\label{s:non_analytical_pac_bayes}

We next discuss three methods to numerically optimize the PAC-Bayes bound. These methods exploit the observation in our experiments that there is a large overlap between the subspace spanned by the stiff eigenvectors of the FIM at the end of training with the corresponding subspace at the beginning of training (\cref{fig:overlap_fim_end_init}). Similarly, there is a large overlap between the subspace spanned by the stiff eigenvectors of the Hessian with that of the FIM (\cref{fig:full_vs_kfac}). We will use the notation $\ev(A)$ to denote the set of eigenvectors of the matrix $A$, arranged in decreasing order of eigenvalues.

\paragraph{Method 2: Eigenvectors of covariance of $Q$ are fixed to those of FIM at initialization} ($\ev(\S_q) = \ev(F_{w_0})$)
We pick the posterior covariance matrix to have the same eigenvectors as that of FIM at initialization and optimize only its eigenvalues, i.e., we set
\beq{
    P = N(w_0, \e^{-1} I),\ Q = N(w, \S_q = U_{w_0} \Lbar U_{w_0}^\top),
    \label{eq:Sq_fisher_init}
}
where $F_{w_0} = U_{w_0} \L U_{w_0}^\top$ is the orthonormal decomposition of the FIM at initialization $w_0$. We can optimize the bound in~\cref{eq:pac_bayes} numerically over the mean of the posterior $w$, eigenvalues of the covariance $\Lbar$ and the scale of the prior $\e$.

\paragraph{Method 3: Eigenvectors of covariance of $Q$ are the same as those of the Hessian} ($\ev(\S_q) = \ev(H_{w})$)
We show in~\cref{s:app:lbar_4nm1} that the covariance of the optimal Gaussian posterior has the same eigenvectors as those of the Hessian. Building upon this, we modify the eigenvectors of $\S_q$ while optimizing the bound as
\beq{
    P = N(w_0, \e^{-1} I),\ Q = N(w, \S_q = U_w \Lbar U_w^\top),
    \label{eq:Sq_hessian_moving}
}
where $H_w = U_w \L U_w^\top$ is the orthonormal decomposition of the Hessian at weights $w$. The variables of optimization are the mean of the posterior $w$, eigenvalues of the covariance $\Lbar$, and the scale of the prior $\e$. Note that this involves recomputing the Hessian at every candidate weight $w$ during optimization of the bound.

\paragraph{Method 4: Covariance of $P$ is proportional to FIM at initialization; eigenvectors of covariance of $Q$ are the same as those of FIM at initialization} ($\S_p = a F_{w_0} + \e^{-1};\ \ev(\S_q) = \ev(F_{w_0})$)
\label{s:method:4}
This is a data-distribution dependent prior. We set
\beq{
    P = N(w_0, a F_{w_0} + \e^{-1} I),\ Q = N(w, \S_q = U_{w_0} \Lbar U_{w_0}^\top),
    \label{eq:Sq_fisher_fisher}
}
where $F_{w_0} = U_{w_0} \L U_{w_0}^\top$ is the orthonormal decomposition of the FIM at initialization $w_0$. The variables of optimization of the bound are the posterior mean $w$, eigenvalues $\Lbar$, and scalar constants $a$ and $\e^{-1}$.

\section{Empirical Validation}
\label{s:expt}

We use fully-connected networks (of varying widths, and up to two hidden layers), convolutional networks (LeNet, ALL-CNN of~\citet{springenbergStrivingSimplicityAll2015} and wide residual network of~\citet{zagoruyko2016wide}) of varying sizes on MNIST~\citep{lecun1990handwritten} and CIFAR-10~\citep{krizhevsky2009learning} for empirical validation of our theoretical results. See~\cref{s:app:setup,s:app:technical_details_pac_bayes} for further details.

To be able to work with Hessian/FIM of large networks, in some cases, e.g.,~\cref{fig:intro} we compute fewer eigenvalues, but compute them exactly without any approximations. When the Hessian/FIM are used for optimizing the PAC-Bayes bound (e.g., Methods 3--4) we use Kronecker-factor (KFAC) approximation of the Gauss-Newton matrix in Backpack~\citep{dangel2020backpack}. We have also developed a trick in PyTorch (see~\cref{s:app:working_efficiently_bayesian}) that allows us to quickly estimate $\Breve{e}(Q, D_n)$ using a large number of samples (we use 150, for comparison~\citet{dziugaiteComputingNonvacuousGeneralization2017} use 1). This allows us to optimize the PAC-Bayes bound with much fewer iterations. This trick that we have developed is also useful for Bayesian deep learning~\citep{wilson2020case}.

\subsection{Sloppiness of the Hessian, FIM and Other Related Quantities in the Network}
\label{s:data_hessian_fim_sloppy}

\begin{figure}[htpb]
\centering
\begin{subfigure}[b]{0.49\linewidth}
\centering
\includegraphics[width=\linewidth]{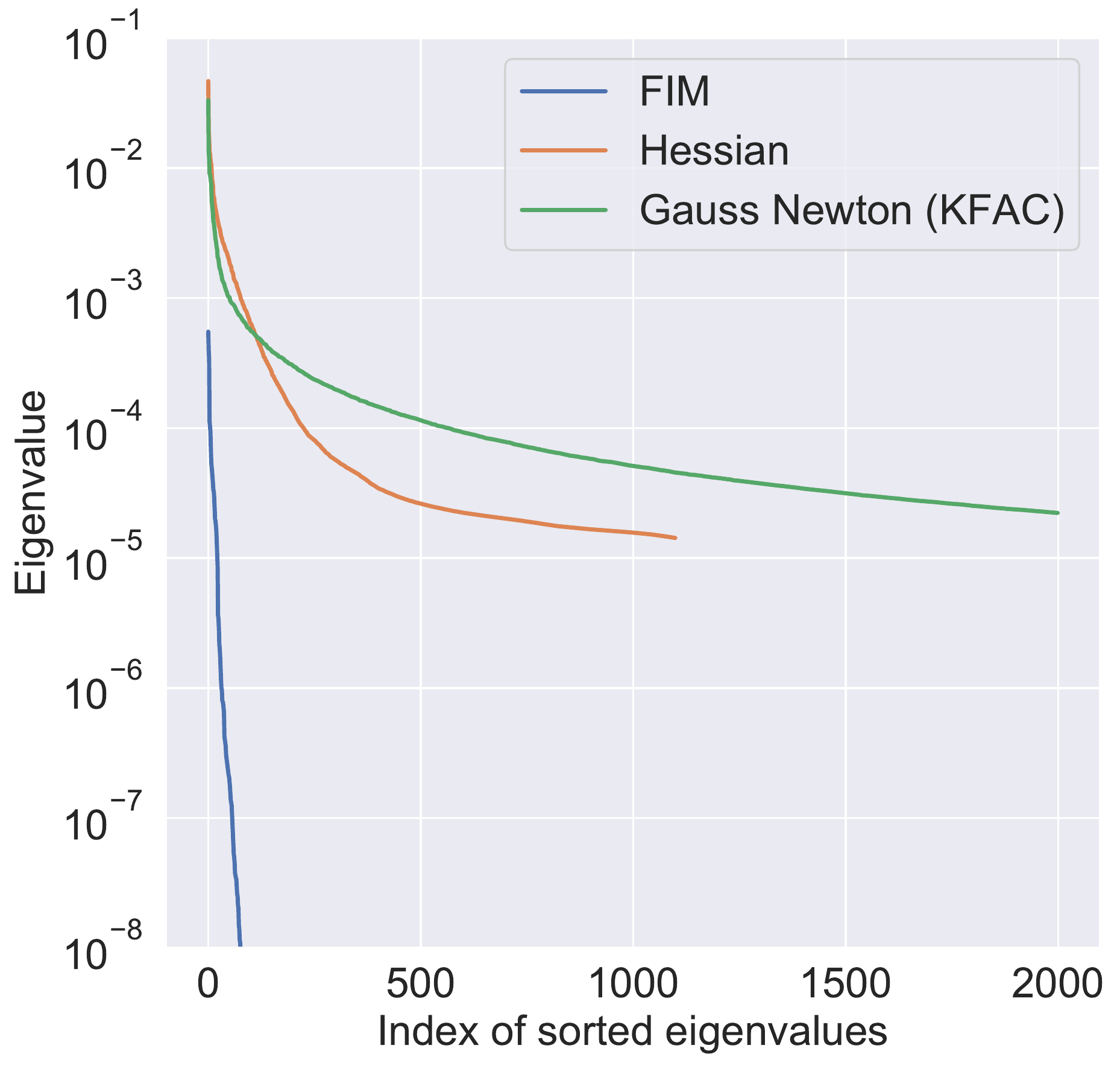}
\end{subfigure}
\begin{subfigure}[b]{0.49\linewidth}
\centering
\includegraphics[width=\linewidth]{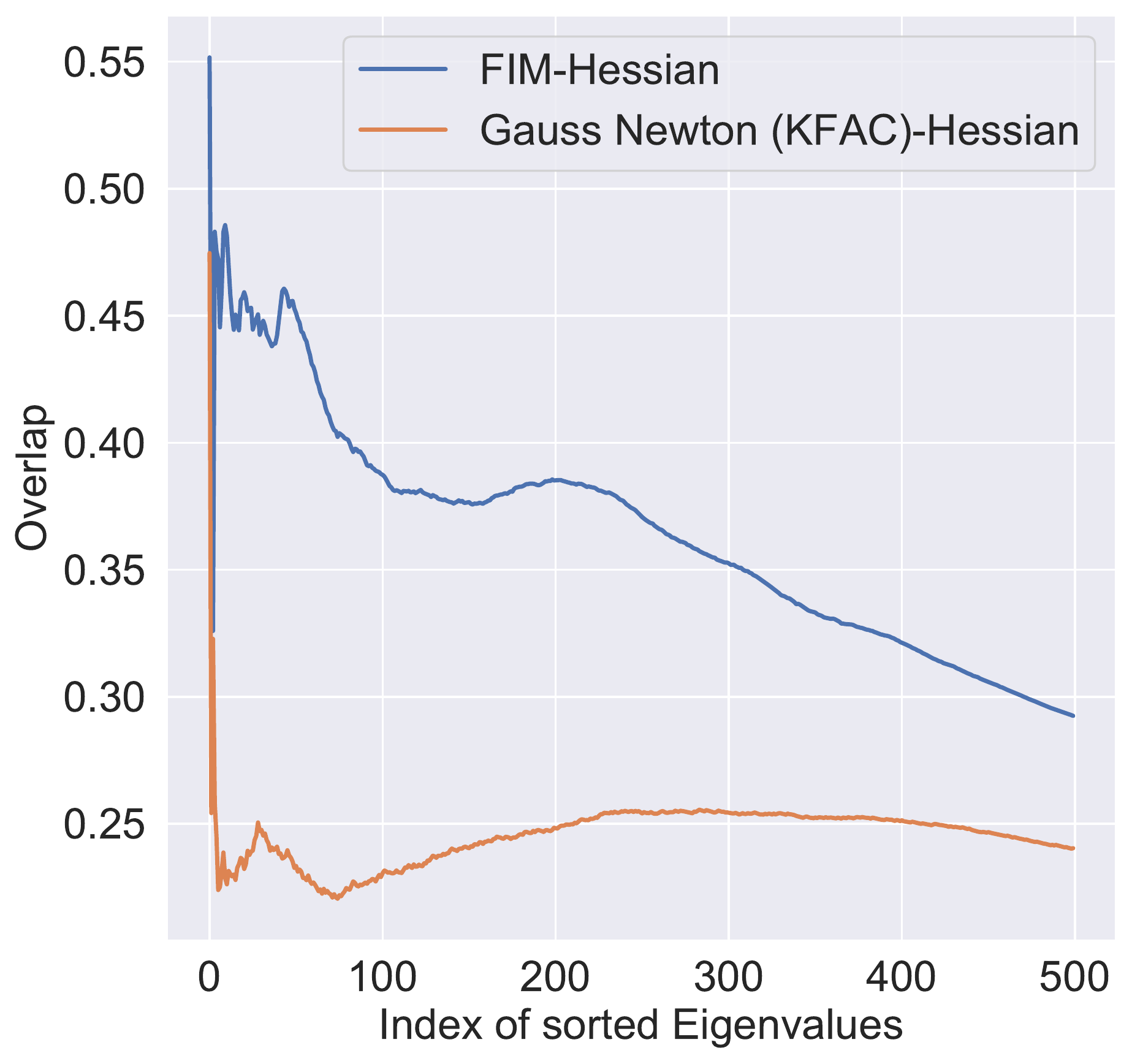}
\end{subfigure}
\caption{(Left) \textbf{Eigenspectra of FIM, Hessian and a KFAC approximation of the Gauss-Newton matrix for a two-layer fully-connected network on MNIST.} Even if FIM's eigenvalues are quite different, its eigenvectors have a large inner product with those of the Hessian (right), much larger than a random vector. KFAC is a good approximation for the eigenvalues of the Hessian but eigenvectors computed from KFAC are quite different from those of the Hessian. This also shows that eigenvectors of the FIM have a strong overlap with those of the Hessian.}
\label{fig:full_vs_kfac}
\end{figure}

\cref{s:app:additional:sloppiness} shows the eigenspectra of the Hessian, FIM and correlations of the activations, logit Jacobians and activation gradients for two and three-layer fully-connected networks on MNIST and All-CNN and a wide residual network on CIFAR-10;~\cref{s:app:eig_mid} shows some eigenspectra at the middle of training;~\cref{s:app:eig_random_end} shows eigenspectra for synthetic datasets. The eigenspectra are qualitatively the same as those in~\cref{fig:intro} so we do not repeat them in the main text. \cref{fig:full_vs_kfac} studies how eigenspectra of FIM and Hessian compare to their KFAC approximations.

\subsection{Overlap of the Stiff Subspaces of the FIM/Hessian at the End of Training with that at Initialization}

\begin{figure}[htpb]
\centering
\begin{subfigure}[c]{0.49\linewidth}
\centering
\includegraphics[width=\linewidth]{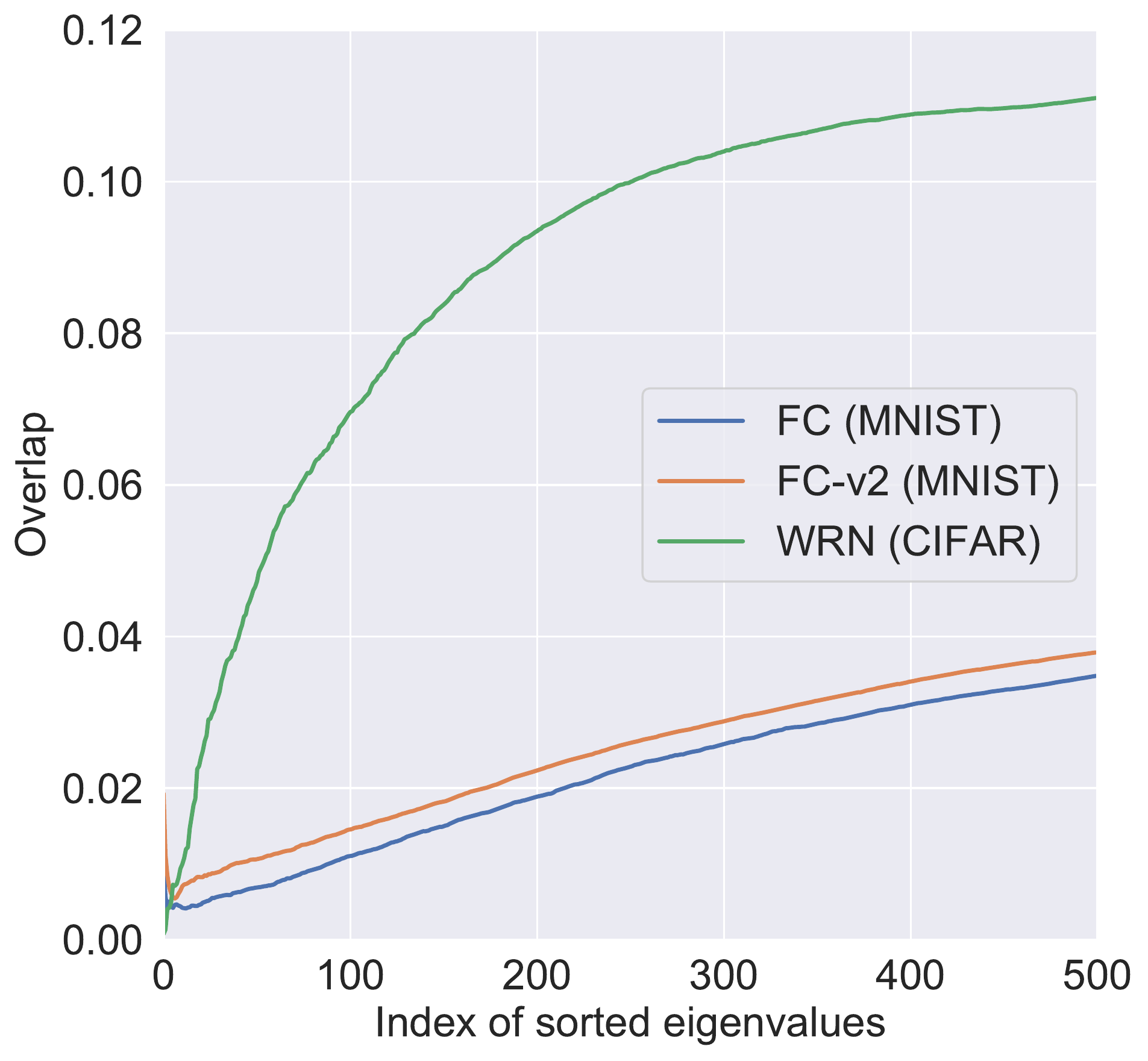}
\end{subfigure}
\begin{subfigure}[c]{0.49\linewidth}
\centering
\includegraphics[width=\linewidth]{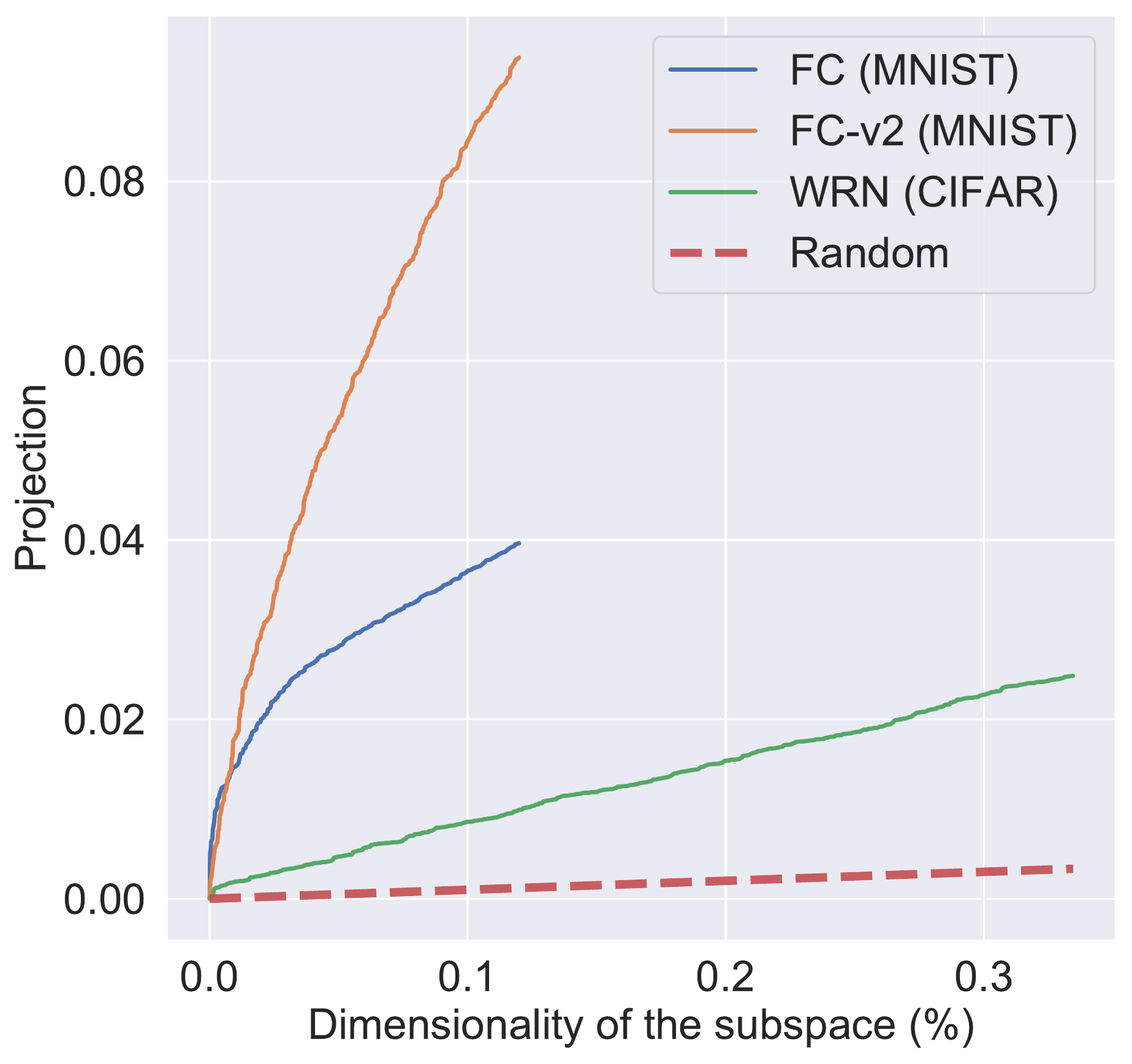}
\end{subfigure}
\caption{(Left) \textbf{Overlap between the subspace of the top $k$ eigenvectors (X-axis) of the FIM at the end of training with that at initialization} ($\norm{\ev_k(F_w)^\top \ev_k(F_{w_0})}_F^2 / k$) for fully-connected (FC) and convolutional networks (WRN) is far larger than overlap of two random subspaces in $\mathbb{R}^p$, which is approximately $10^{-6}$. (Right) \textbf{Projection} $\norm{\ev_k(F_{w_0}) \D w}_2^2 /\norm{\D w}_2^2$ \textbf{of the change in weights} (where $\D w = w-w_0$) \textbf{into the subspace of the top $k$} (shown as percentage of weights because different networks have different sizes) eigenvectors of the FIM is much larger than the projection into a random subspace.
}
\label{fig:overlap_fim_end_init}
\end{figure}

\cref{fig:overlap_fim_end_init} (left) computes the overlap of the subspace spanned by the top $k$ eigenvectors of the FIM at the end of training with that at the initialization. \cref{fig:overlap_fim_end_init} (right) shows a projection of the change in the weights (difference between weights at the and that at initialization) into the subspace spanned by the top $k$ eigenvectors of the FIM.  Both these overlaps are large, e.g., projection into a random subspace of dimension $k$ for the latter. This suggests that during training, weights change predominantly in the stiff subspace of the FIM at initialization.

We also constructed a third network (denoted -v2) as follows. Given a trained network $w$ from initialization $w_0$, we train $w$ for more epochs to minimize the training loss and a penalty $\norm{w-w_0}^2$ which pulls it closer to $w_0$---without changing the training/validation error much.
We find in~\cref{fig:overlap_fim_end_init} (right) that this variant has a much larger projection into the stiff eigenspace, and thereby a smaller overlap with the sloppy eigenvectors. Thus, weights can effectively ``come back'' towards the initialization in the sloppy subspace although they evolved during training in the stiff subspace. See \cref{fig:pruning} for more details.

\subsection{PAC-Bayes Bounds}
\label{s:expt:pac_bayes}

{
\renewcommand{\arraystretch}{1.25}
\begin{table}
    \centering
    \rowcolors{1}{}{black!5}
    \resizebox{0.7\linewidth}{!}{
    \begin{tabular}{l *{6}r}
    \rowcolor{white}
    \toprule
    Model & \multicolumn{6}{c}{Method}\\
    & 1 & 2 & 3 & 4 & A & B \\
    \midrule
    FC-600-1& 0.3241 &0.1590 &0.1357 &0.1323 &0.161 &0.1198 \\
    FC-600-2&  0.3794 &0.1767 &0.1540 &0.1397 &0.186 &0.1443   \\
    FC-1200-1& 0.3509 &0.1523 &0.1515 &0.1486 &0.179 &0.1413   \\
    FC-1200-2&  0.3915 &0.2017 &0.1817 &0.1702 &0.223 &-\\
    LeNet-5 &  0.0572 &0.0099 &0.0188 &0.0092 &- &-\\
    \bottomrule
    \end{tabular}
}
\caption{\textbf{PAC-Bayes bounds on MNIST for different methods.} Methods 1--4 are ours, described in \cref{s:pac_bayes_bounds,s:non_analytical_pac_bayes}. The prior for Method 4 is $\S_p = a F_{w_0} + \e^{-1}$; all other methods use $P = N(w_0, \e^{-1} I)$. The penultimate column (A) is from~\citet{dziugaiteComputingNonvacuousGeneralization2017} and optimizes the diagonal of the covariance of $Q$ numerically. The final column (B) which sets eigenvectors of covariance of $Q$ to be the same as that of the block-diagonal Hessian is from~\citet{wuDissectingHessianUnderstanding2021}. For fully-connected nets, the error $\eq$ ranges from 6--8 $\times 10^{-2}$ for Method 1 and 1--4 $\times 10^{-2}$ for all other methods. For LeNet-5 the error $\eq$ ranges from 1--2 $\times 10^{-2}$ for all methods. See~\cref{s:app:pac_bayes_full} for the extended version.\\}
\label{tab:pac_bayes}
\end{table}
}

\begin{figure}
\centering
\begin{subfigure}[c]{0.49\linewidth}
\centering
\includegraphics[width=\linewidth]{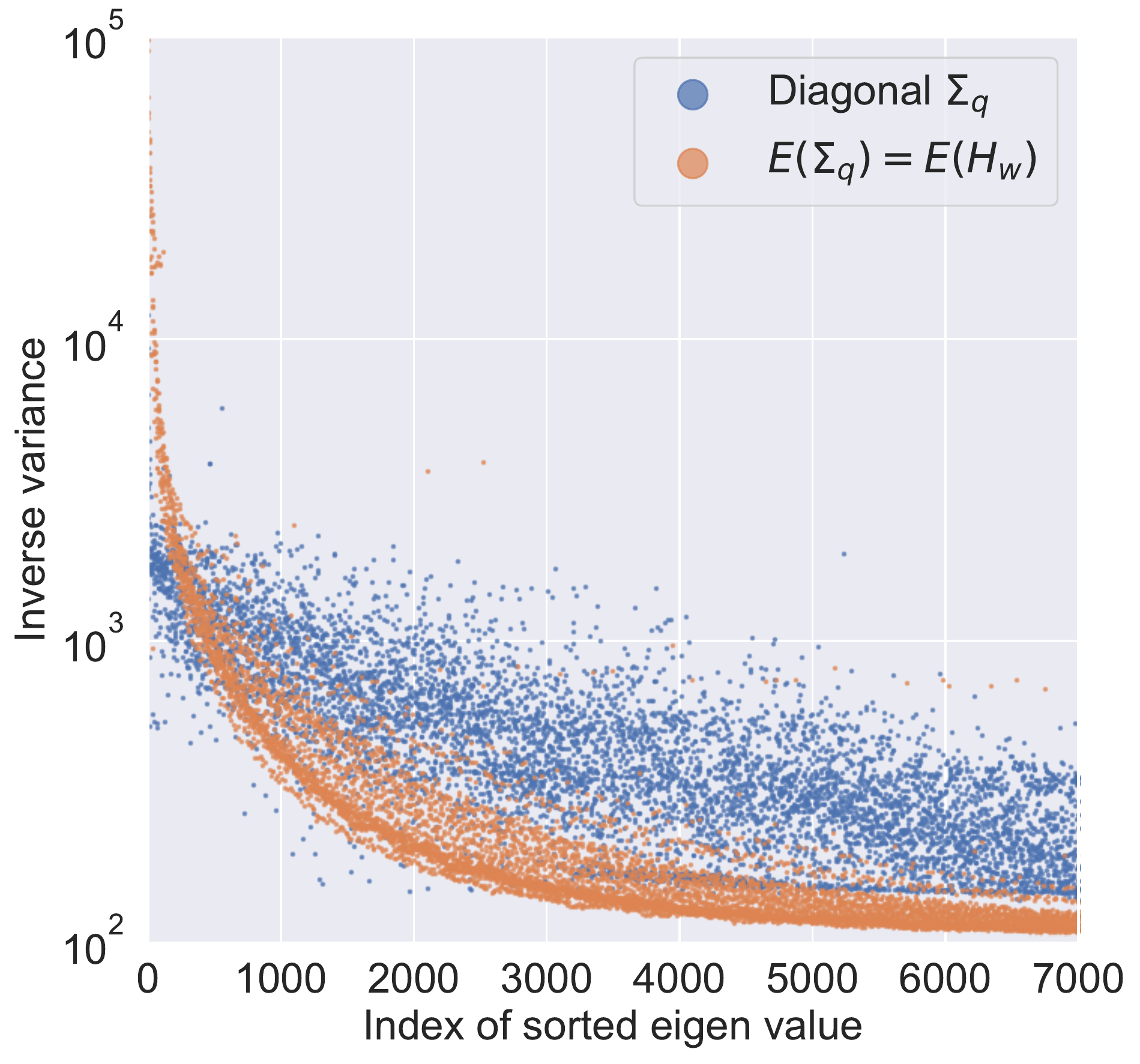}
\end{subfigure}
\begin{subfigure}[c]{0.49\linewidth}
\centering
\includegraphics[width=\linewidth]{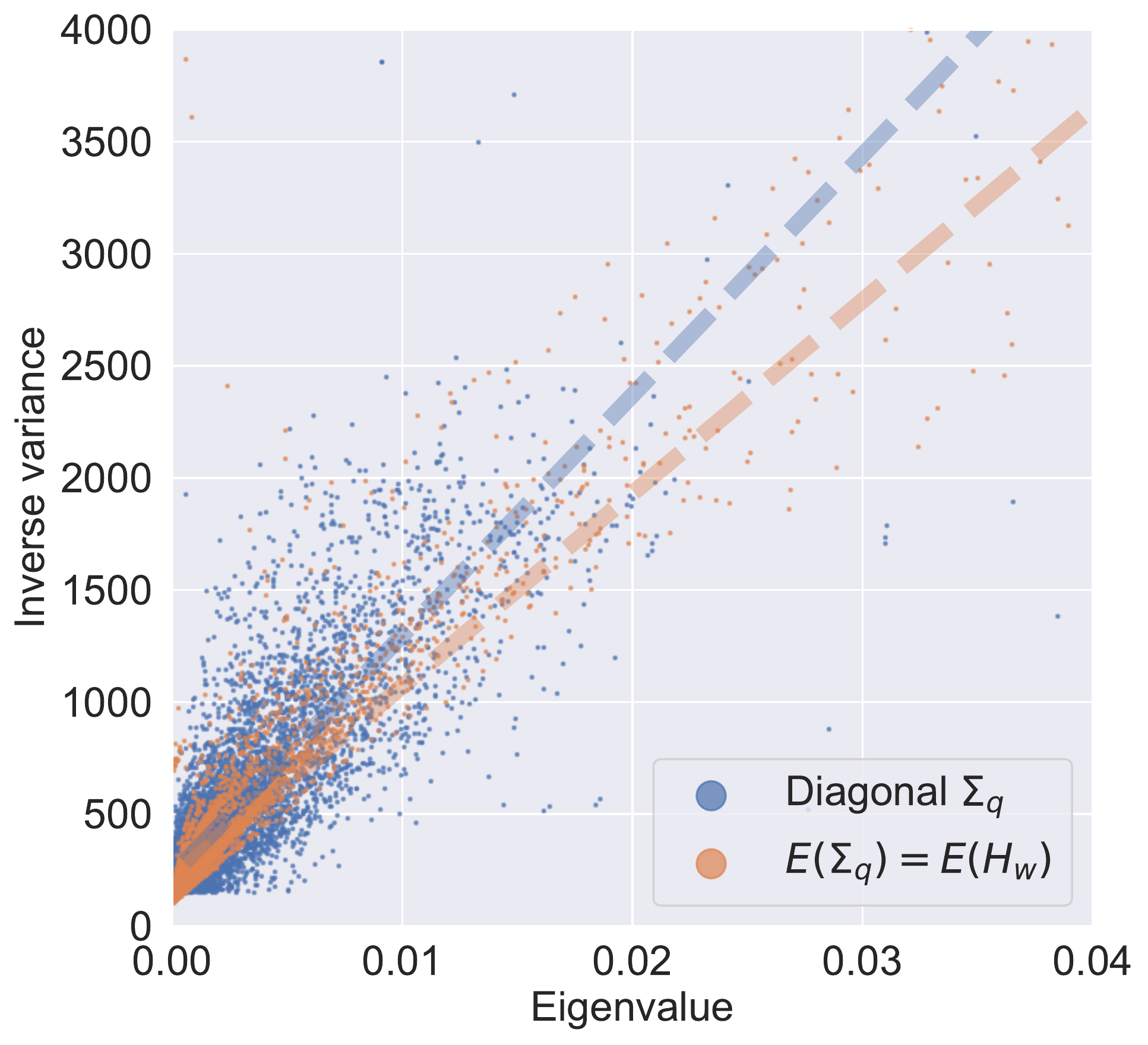}
\end{subfigure}
\caption{
\textbf{Posterior covariance computed by numerically optimizing PAC-Bayes bound is aligned with sloppy directions.} (Left) Inverse eigenvalues of the posterior covariance ($\lbar_i^{-1}$) \emph{indexed by eigenvalues} of the Hessian (X-axis) for $\ev(\S_q) = \ev(H_w)$ (orange, our Method 3) and a diagonal-$\S_q$ (blue, this is the method of~\citet{dziugaiteComputingNonvacuousGeneralization2017}). For the latter, abscissa are the indices of the sorted diagonal of the Hessian.
(Right) Inverse eigenvalues of the posterior covariance ($\lbar_i^{-1}$) \emph{plotted against eigenvalues} of the Hessian for $\ev(\S_q) = \ev(H_w)$ (orange) and for diagonal-$\S_q$ (blue). For the latter the X-axis is the corresponding entry on the diagonal of the Hessian. For the orange pointcloud, we obtain a surprisingly accurate fit for our analytical expression $(\lbar_i^{-1} = 2(n-1) \l_i + \e$: we get $n \sim 43,000$ (true $n=55,000$) and $\e \sim 210.6$ (true $\e = 101.3$) with $R^2 = 0.972$, which indicates  good fit for regression.
}
\label{fig:posterior_scatter}
\end{figure}

\cref{tab:pac_bayes} shows results of using different methods to calculate PAC-Bayes bounds. It is remarkable that for all networks, the analytical method using Method 1 obtains a non-vacuous bound. Methods 2--4 obtain bounds that are comparable to those of existing methods, e.g.,\citet{dziugaiteComputingNonvacuousGeneralization2017,wuDissectingHessianUnderstanding2021}. \cref{s:app:technical_details_pac_bayes} discusses a number of technical details in how each method are implemented numerically, e.g., sampling weights from posteriors whose covariance is represented as a KFAC approximation. As discussed in~\cref{s:non_analytical_pac_bayes}, our methods in~\cref{tab:pac_bayes} exploit the fact that the stiff subspaces of the Hessian/FIM have a strong overlap at the beginning and that training predominantly takes place in the stiff subspace of the FIM at initialization. This experiment therefore shows that we can exploit sloppiness to compute non-vacuous generalization bounds for deep networks.

\textbf{We next study how similar the optimal PAC-Bayes posterior covariance computed by numerical optimization and the one obtained analytically are.} We will contrast two numerical methods, our Method 3 with $\ev(\S_q) = \ev(H_w)$, and the method of~\citet{dziugaiteComputingNonvacuousGeneralization2017} in~\cref{tab:pac_bayes} which uses a diagonal posterior covariance. We find in~\cref{fig:posterior_scatter} that the eigenvalues of the posterior computed by Method 3 match remarkably well with our analytical expression in~\cref{eq:lbar_4nm1} for Method 1. This sheds light into why our analytical PAC-Bayes bound is non-vacuous---essentially numeric optimization finds a very similar posterior as that of the analytical method. It is however important to run even the numerical optimization in the appropriate basis. The bounds obtained by~\citet{dziugaiteComputingNonvacuousGeneralization2017} for a diagonal posterior are worse than Method 3 in~\cref{tab:pac_bayes} and as~\cref{fig:posterior_scatter} (left) indicates, this is because their posterior has a smaller variance in the sloppy subspace (blue cloud).

\section{Related Work}
\label{s:related}

\noindent \textbf{Sloppy models in physics and biology}
Our work is inspired by~\citet{brown2004statistical,gutenkunst2007universally} who noticed that regression models fitted to systems biology data have few stiff parameters that determine the outcome and a large number of sloppy parameters which only weakly determine the outcome. These authors have developed an elaborate geometric understanding of this phenomenon, see~\citet{transtrumGeometryNonlinearLeast2011} and references therein. While sloppiness is thought to be a universal property of parametric models~\citep{waterfallSloppyModelUniversalityClass2006}, the mechanism that causes models to be sloppy has not been studied yet. This work has also exclusively focused on the under-parameterized regime. We connect the sloppiness of a deep network to the sloppiness of inputs and show that if the inputs are sloppy, then key quantities pertaining to the model, e.g., activations, FIM and Hessian etc., are also sloppy.

\noindent \textbf{Hessian and the FIM of deep networks}
have been studied to understand the local geometry of the energy landscape and the behavior of SGD, see~\citet{hochreiterFlatMinima1997,chaudhari2016entropy,fortEmergentPropertiesLocal2019}, among others. FIM has been used to study optimization~\citep{6790500,martensOptimizingNeuralNetworks2016,karakidaUniversalStatisticsFisher2019}, gradient diversity~\citep{yin2018gradient,chaudhari2017stochastic}, and generalization~\citep{achille2019information}. A number of these works have pointed out that the Hessian and the FIM have spiky/large eigenvalues~\citep{papyan2019measurements} along with a bulk of near-zero eigenvalues~\citep{papyan2018full,penningtonGeometryNeuralNetwork}, and that this indicates that the energy landscape, or the prediction space, is locally flat. We focus on the decay pattern of the eigenspectra of these matrices and discover that it mirrors the decay pattern of the inputs for typical datasets. We see a strong overlap of the stiff subspace of the Hessian/FIM at initialization with that at the end of training; this is consistent with the analysis in~\citet{gur-ariGradientDescentHappens2018,chizat2019lazy}.

\noindent \textbf{Generalization}
PAC-Bayes bounds for deep networks have been obtained using the methods of~\cite{langford2002not} by~\citet{dziugaiteComputingNonvacuousGeneralization2017,dziugaite2020revisiting,zhou2018non}. While analytical generalization bounds are often vacuous~\citep{bartlett2017spectrally,bartlett2021deep,neyshaburExploringGeneralizationDeep2017}, we show that if we exploit the sloppiness of the Hessian, then we can obtain non-vacuous analytical bounds. We show that the posterior computed by the method of~\citet{dziugaiteComputingNonvacuousGeneralization2017} aligns well with sloppy eigenvalues of the Hessian/FIM. We build upon this work and show the benefits of sloppiness by providing data-distribution dependent PAC-Bayes bounds (also see~\citet{dziugaite2018data}).

Our Method 3 is related to the work of~\citet{wuDissectingHessianUnderstanding2021}. The difference is that they assume that the block-diagonal approximation of the Hessian decouples into a Kronecker product of the Hessian of the activations and the input correlation matrix; we instead optimize the PAC-Bayes prior using the top few eigenvectors of the full Hessian for some models (LeNet) and the Kronecker-factored approximation of the blocks for others. They analyze the case when the data matrix has rank 1 and Hessian has rank $m-1$ ($m$ is the number of classes). Our experiments show that they are both full-rank but sloppy, and we therefore analyze this instead.

\citet{bartlett2020benign} show that a minimum norm interpolating solution of over-parameterized linear regression can predict accurately if the data matrix has a long tail of small eigenvalues. Our notion of effective dimensionality is also seen in their calculations: roughly speaking, larger our sloppiness factor $c$ in~\cref{def:sloppy}, better the excess risk in their linear regression, which is consistent with~\cref{fig:intro} (bottom right). \cite{JustInterpolate} show similar results on the minimum-norm interpolating solution for kernel regression.

\section{Discussion}
\label{s:discussion}

We showed that for typical datasets, the sloppy decay pattern of eigenvalues of the input correlation matrix is mirrored in key quantities of the deep network, e.g., eigenspectra of the activation correlations, activation gradients, logit Jacobians, Hessian and the FIM. This suggests that the ``simplicity'', more precisely, the sloppiness, of inputs of high-dimensional datasets controls the representations learned by the network. We validated this hypothesis by providing non-vacuous PAC-Bayes generalization bounds for deep networks, including analytical ones. Our calculations also provided a simple definition of the effective dimensionality of a deep network and we showed how this number can be much smaller than the number of weights.



\section{Acknowledgments}
This work was supported by grants from the National Science Foundation (2145164) and the Office of Naval Research (N00014-22-1-2255), and cloud computing credits from Amazon Web Services.

\begin{small}
\bibliography{main}

\begin{thebibliography}{45}
\providecommand{\natexlab}[1]{#1}
\providecommand{\url}[1]{\texttt{#1}}
\expandafter\ifx\csname urlstyle\endcsname\relax
  \providecommand{\doi}[1]{doi: #1}\else
  \providecommand{\doi}{doi: \begingroup \urlstyle{rm}\Url}\fi

\bibitem[Achille et~al.(2018)Achille, Rovere, and Soatto]{achille2018critical}
Achille, A., Rovere, M., and Soatto, S.
\newblock Critical learning periods in deep networks.
\newblock In \emph{International Conference on Learning Representations}, 2018.

\bibitem[Achille et~al.(2019)Achille, Paolini, and
  Soatto]{achille2019information}
Achille, A., Paolini, G., and Soatto, S.
\newblock Where is the information in a deep neural network?
\newblock \emph{arXiv preprint arXiv:1905.12213}, 2019.

\bibitem[Amari(1998)]{6790500}
Amari, S.-i.
\newblock Natural gradient works efficiently in learning.
\newblock \emph{Neural Computation}, 10\penalty0 (2):\penalty0 251--276, 1998.
\newblock \doi{10.1162/089976698300017746}.

\bibitem[Amari et~al.(2002)Amari, Park, and Ozeki]{amari2002geometrical}
Amari, S.-i., Park, H., and Ozeki, T.
\newblock Geometrical singularities in the neuromanifold of multilayer
  perceptrons.
\newblock \emph{Advances in neural information processing systems}, 1:\penalty0
  343--350, 2002.

\bibitem[Ambroladze et~al.(2007)Ambroladze, {Parrado-Hernández}, and
  {Shawe-Taylor}]{ambroladze2007tighter}
Ambroladze, A., {Parrado-Hernández}, E., and {Shawe-Taylor}, J.
\newblock Tighter pac-bayes bounds.
\newblock \emph{Advances in neural information processing systems},
  19:\penalty0 9, 2007.

\bibitem[Bartlett et~al.(2017)Bartlett, Foster, and
  Telgarsky]{bartlett2017spectrally}
Bartlett, P.~L., Foster, D.~J., and Telgarsky, M.~J.
\newblock Spectrally-normalized margin bounds for neural networks.
\newblock In \emph{Advances in Neural Information Processing Systems}, pp.\
  6240--6249, 2017.

\bibitem[Bartlett et~al.(2020)Bartlett, Long, Lugosi, and
  Tsigler]{bartlett2020benign}
Bartlett, P.~L., Long, P.~M., Lugosi, G., and Tsigler, A.
\newblock Benign overfitting in linear regression.
\newblock \emph{Proceedings of the National Academy of Sciences}, 117\penalty0
  (48):\penalty0 30063--30070, 2020.

\bibitem[Bartlett et~al.(2021)Bartlett, Montanari, and
  Rakhlin]{bartlett2021deep}
Bartlett, P.~L., Montanari, A., and Rakhlin, A.
\newblock Deep learning: A statistical viewpoint.
\newblock \emph{arXiv preprint arXiv:2103.09177}, 2021.

\bibitem[Botev et~al.(2017)Botev, Ritter, and Barber]{ICML-2017-BotevRB}
Botev, A., Ritter, H., and Barber, D.
\newblock {Practical Gauss-Newton Optimisation for Deep Learning}.
\newblock In \emph{{Proceedings of the 34th International Conference on Machine
  Learning}}, pp.\  557--565. {PMLR}, 2017.

\bibitem[Brown et~al.(2004)Brown, Hill, Calero, Myers, Lee, Sethna, and
  Cerione]{brown2004statistical}
Brown, K.~S., Hill, C.~C., Calero, G.~A., Myers, C.~R., Lee, K.~H., Sethna,
  J.~P., and Cerione, R.~A.
\newblock The statistical mechanics of complex signaling networks: Nerve growth
  factor signaling.
\newblock \emph{Physical biology}, 1\penalty0 (3):\penalty0 184, 2004.

\bibitem[Chaudhari \& Soatto(2018)Chaudhari and
  Soatto]{chaudhari2017stochastic}
Chaudhari, P. and Soatto, S.
\newblock Stochastic gradient descent performs variational inference, converges
  to limit cycles for deep networks.
\newblock \emph{Proc. of International Conference of Learning and
  Representations (ICLR), Apr 30-May 3, 2018}, 2018.

\bibitem[Chaudhari et~al.(2017)Chaudhari, Choromanska, Soatto, LeCun, Baldassi,
  Borgs, Chayes, Sagun, and Zecchina]{chaudhari2016entropy}
Chaudhari, P., Choromanska, A., Soatto, S., LeCun, Y., Baldassi, C., Borgs, C.,
  Chayes, J., Sagun, L., and Zecchina, R.
\newblock Entropy-{{SGD}}: {{Biasing}} gradient descent into wide valleys.
\newblock In \emph{Proc. of the {{International Conference}} on {{Learning
  Representations}} ({{ICLR}})}, 2017.

\bibitem[Chizat et~al.(2019)Chizat, Oyallon, and Bach]{chizat2019lazy}
Chizat, L., Oyallon, E., and Bach, F.
\newblock On lazy training in differentiable programming.
\newblock \emph{Advances in Neural Information Processing Systems},
  32:\penalty0 2937--2947, 2019.

\bibitem[Dangel et~al.(2020)Dangel, Kunstner, and Hennig]{dangel2020backpack}
Dangel, F., Kunstner, F., and Hennig, P.
\newblock {{BackPACK}}: {{Packing}} more into backprop.
\newblock In \emph{International Conference on Learning Representations}, 2020.

\bibitem[Dziugaite(2020)]{dziugaite2020revisiting}
Dziugaite, G.~K.
\newblock \emph{Revisiting Generalization for Deep Learning: {{PAC}}-{{Bayes}},
  Flat Minima, and Generative Models}.
\newblock PhD thesis, University of Cambridge, 2020.

\bibitem[Dziugaite \& Roy(2017)Dziugaite and
  Roy]{dziugaiteComputingNonvacuousGeneralization2017}
Dziugaite, G.~K. and Roy, D.~M.
\newblock Computing {{Nonvacuous Generalization Bounds}} for {{Deep}}
  ({{Stochastic}}) {{Neural Networks}} with {{Many More Parameters}} than
  {{Training Data}}.
\newblock In \emph{Proc. of the {{Conference}} on {{Uncertainty}} in
  {{Artificial Intelligence}} ({{UAI}})}, 2017.

\bibitem[Dziugaite \& Roy(2018)Dziugaite and Roy]{dziugaite2018data}
Dziugaite, G.~K. and Roy, D.~M.
\newblock Data-dependent {{PAC}}-{{Bayes}} priors via differential privacy.
\newblock In \emph{Proceedings of the 32nd International Conference on Neural
  Information Processing Systems}, pp.\  8440--8450, 2018.

\bibitem[Fort \& Ganguli(2019)Fort and
  Ganguli]{fortEmergentPropertiesLocal2019}
Fort, S. and Ganguli, S.
\newblock Emergent properties of the local geometry of neural loss landscapes.
\newblock \emph{arXiv:1910.05929 [cs, stat]}, October 2019.

\bibitem[{Gur-Ari} et~al.(2018){Gur-Ari}, Roberts, and
  Dyer]{gur-ariGradientDescentHappens2018}
{Gur-Ari}, G., Roberts, D.~A., and Dyer, E.
\newblock Gradient {{Descent Happens}} in a {{Tiny Subspace}}.
\newblock \emph{arXiv:1812.04754 [cs, stat]}, December 2018.

\bibitem[Gutenkunst et~al.(2007)Gutenkunst, Waterfall, Casey, Brown, Myers, and
  Sethna]{gutenkunst2007universally}
Gutenkunst, R.~N., Waterfall, J.~J., Casey, F.~P., Brown, K.~S., Myers, C.~R.,
  and Sethna, J.~P.
\newblock Universally sloppy parameter sensitivities in systems biology models.
\newblock \emph{PLoS computational biology}, 3\penalty0 (10):\penalty0 e189,
  2007.

\bibitem[Hochreiter \& Schmidhuber(1997)Hochreiter and
  Schmidhuber]{hochreiterFlatMinima1997}
Hochreiter, S. and Schmidhuber, J.
\newblock Flat minima.
\newblock \emph{Neural Computation}, 9\penalty0 (1):\penalty0 1--42, 1997.

\bibitem[Karakida et~al.(2019)Karakida, Akaho, and
  Amari]{karakidaUniversalStatisticsFisher2019}
Karakida, R., Akaho, S., and Amari, S.-i.
\newblock Universal {{Statistics}} of {{Fisher Information}} in {{Deep Neural
  Networks}}: {{Mean Field Approach}}.
\newblock \emph{arXiv:1806.01316 [cond-mat, stat]}, October 2019.

\bibitem[Karoui(2010)]{Karoui2010TheSO}
Karoui, N.~E.
\newblock The spectrum of kernel random matrices.
\newblock \emph{Annals of Statistics}, 38:\penalty0 1--50, 2010.

\bibitem[Krizhevsky(2009)]{krizhevsky2009learning}
Krizhevsky, A.
\newblock \emph{Learning Multiple Layers of Features from Tiny Images}.
\newblock PhD thesis, Computer Science, University of Toronto, 2009.

\bibitem[Langford \& Caruana(2002)Langford and Caruana]{langford2002not}
Langford, J. and Caruana, R.
\newblock ({{Not}}) bounding the true error.
\newblock \emph{Advances in Neural Information Processing Systems}, 2:\penalty0
  809--816, 2002.

\bibitem[Langford \& Seeger(2001)Langford and Seeger]{langford2001bounds}
Langford, J. and Seeger, M.
\newblock Bounds for averaging classifiers.
\newblock 2001.

\bibitem[LeCun et~al.(1990)LeCun, Boser, Denker, Henderson, Howard, Hubbard,
  and Jackel]{lecun1990handwritten}
LeCun, Y., Boser, B.~E., Denker, J.~S., Henderson, D., Howard, R.~E., Hubbard,
  W.~E., and Jackel, L.~D.
\newblock Handwritten digit recognition with a back-propagation network.
\newblock In \emph{Advances in Neural Information Processing Systems}, pp.\
  396--404, 1990.

\bibitem[Liang \& Rakhlin(2018)Liang and Rakhlin]{JustInterpolate}
Liang, T. and Rakhlin, A.
\newblock Just interpolate: Kernel "ridgeless" regression can generalize.
\newblock \emph{CoRR}, abs/1808.00387, 2018.
\newblock URL \url{http://arxiv.org/abs/1808.00387}.

\bibitem[Martens \& Grosse(2016)Martens and
  Grosse]{martensOptimizingNeuralNetworks2016}
Martens, J. and Grosse, R.
\newblock Optimizing {{Neural Networks}} with {{Kronecker}}-factored
  {{Approximate Curvature}}.
\newblock \emph{arXiv:1503.05671 [cs, stat]}, May 2016.

\bibitem[McAllester(1999)]{mcallester1999pac}
McAllester, D.~A.
\newblock {{PAC}}-{{Bayesian}} model averaging.
\newblock In \emph{Proceedings of the Twelfth Annual Conference on
  {{Computational}} Learning Theory}, pp.\  164--170, 1999.

\bibitem[Neyshabur et~al.(2017)Neyshabur, Bhojanapalli, McAllester, and
  Srebro]{neyshaburExploringGeneralizationDeep2017}
Neyshabur, B., Bhojanapalli, S., McAllester, D., and Srebro, N.
\newblock Exploring {{Generalization}} in {{Deep Learning}}.
\newblock \emph{arXiv:1706.08947 [cs]}, July 2017.

\bibitem[Papyan(2018)]{papyan2018full}
Papyan, V.
\newblock The full spectrum of deepnet hessians at scale: {{Dynamics}} with sgd
  training and sample size.
\newblock \emph{arXiv preprint arXiv:1811.07062}, 2018.

\bibitem[Papyan(2019)]{papyan2019measurements}
Papyan, V.
\newblock Measurements of three-level hierarchical structure in the outliers in
  the spectrum of deepnet hessians.
\newblock \emph{arXiv preprint arXiv:1901.08244}, 2019.

\bibitem[{Parrado-Hernández} et~al.(2012){Parrado-Hernández}, Ambroladze,
  {Shawe-Taylor}, and Sun]{parrado2012pac}
{Parrado-Hernández}, E., Ambroladze, A., {Shawe-Taylor}, J., and Sun, S.
\newblock {{PAC}}-{{Bayes}} bounds with data dependent priors.
\newblock \emph{The Journal of Machine Learning Research}, 13\penalty0
  (1):\penalty0 3507--3531, 2012.

\bibitem[Pennington \& Bahri()Pennington and
  Bahri]{penningtonGeometryNeuralNetwork}
Pennington, J. and Bahri, Y.
\newblock Geometry of {{Neural Network Loss Surfaces}} via {{Random Matrix
  Theory}}.
\newblock pp.\ ~9.

\bibitem[Sagun et~al.(2016)Sagun, Bottou, and LeCun]{sagun2016singularity}
Sagun, L., Bottou, L., and LeCun, Y.
\newblock Singularity of the hessian in deep learning.
\newblock \emph{arXiv:1611:07476}, 2016.

\bibitem[Schwarz(1978)]{schwarz1978estimating}
Schwarz, G.
\newblock Estimating the dimension of a model.
\newblock \emph{The annals of statistics}, pp.\  461--464, 1978.

\bibitem[Springenberg et~al.(2015)Springenberg, Dosovitskiy, Brox, and
  Riedmiller]{springenbergStrivingSimplicityAll2015}
Springenberg, J.~T., Dosovitskiy, A., Brox, T., and Riedmiller, M.
\newblock Striving for {{Simplicity}}: {{The All Convolutional Net}}.
\newblock \emph{arXiv:1412.6806 [cs]}, April 2015.

\bibitem[Transtrum et~al.(2011)Transtrum, Machta, and
  Sethna]{transtrumGeometryNonlinearLeast2011}
Transtrum, M.~K., Machta, B.~B., and Sethna, J.~P.
\newblock The geometry of nonlinear least squares with applications to sloppy
  models and optimization.
\newblock \emph{Physical Review E}, 83\penalty0 (3):\penalty0 036701, March
  2011.
\newblock ISSN 1539-3755, 1550-2376.
\newblock \doi{10.1103/PhysRevE.83.036701}.

\bibitem[Waterfall et~al.(2006)Waterfall, Casey, Gutenkunst, Brown, Myers,
  Brouwer, Elser, and Sethna]{waterfallSloppyModelUniversalityClass2006}
Waterfall, J.~J., Casey, F.~P., Gutenkunst, R.~N., Brown, K.~S., Myers, C.~R.,
  Brouwer, P.~W., Elser, V., and Sethna, J.~P.
\newblock Sloppy-{{Model Universality Class}} and the {{Vandermonde Matrix}}.
\newblock \emph{Physical Review Letters}, 97\penalty0 (15):\penalty0 150601,
  October 2006.
\newblock \doi{10.1103/PhysRevLett.97.150601}.

\bibitem[Wilson(2020)]{wilson2020case}
Wilson, A.~G.
\newblock The case for bayesian deep learning.
\newblock \emph{arXiv preprint arXiv:2001.10995}, 2020.

\bibitem[Wu et~al.(2021)Wu, Zhu, Wu, Wang, and
  Ge]{wuDissectingHessianUnderstanding2021}
Wu, Y., Zhu, X., Wu, C., Wang, A., and Ge, R.
\newblock Dissecting {{Hessian}}: {{Understanding Common Structure}} of
  {{Hessian}} in {{Neural Networks}}.
\newblock \emph{arXiv:2010.04261 [cs, stat]}, June 2021.

\bibitem[Yin et~al.(2018)Yin, Pananjady, Lam, Papailiopoulos, Ramchandran, and
  Bartlett]{yin2018gradient}
Yin, D., Pananjady, A., Lam, M., Papailiopoulos, D., Ramchandran, K., and
  Bartlett, P.
\newblock Gradient diversity: A key ingredient for scalable distributed
  learning.
\newblock In \emph{International Conference on Artificial Intelligence and
  Statistics}, pp.\  1998--2007. {PMLR}, 2018.

\bibitem[Zagoruyko \& Komodakis(2016)Zagoruyko and
  Komodakis]{zagoruyko2016wide}
Zagoruyko, S. and Komodakis, N.
\newblock Wide residual networks.
\newblock In \emph{British Machine Vision Conference 2016}. {British Machine
  Vision Association}, 2016.

\bibitem[Zhou et~al.(2018)Zhou, Veitch, Austern, Adams, and
  Orbanz]{zhou2018non}
Zhou, W., Veitch, V., Austern, M., Adams, R.~P., and Orbanz, P.
\newblock Non-vacuous generalization bounds at the {{ImageNet}} scale: A
  {{PAC}}-{{Bayesian}} compression approach.
\newblock In \emph{International Conference on Learning Representations}, 2018.

\end{thebibliography}
\bibliographystyle{icml2022}
\end{small}

\newpage
\onecolumn

\renewcommand{\theequation}{S-\arabic{equation}}
\renewcommand{\thefigure}{S-\arabic{figure}}
\renewcommand{\thetable}{S-\arabic{table}}


\begin{appendix}

\section{Details of the experimental setup}
\label{s:app:setup}

\paragraph{Data}
We use the MNIST dataset for experiments on fully-connected networks and LeNet. We setup a binary classification problem (we map \{0,1,2,3,4\} to label 0 and \{5,6,7,8,9\} to label 1). We use 55000 samples from the training set to train the model and to optimize the PAC-Bayes bound. We set aside 5000 samples for calculating the FIM, which is used in Method 4 of PAC-Bayes bound optimization. Strictly speaking, it is not required to do so because a prior that depends upon the FIM is an expectation-prior (as discussed in~\citet{parrado2012pac}) but we set aside these samples to compare in a systematic manner to existing methods in the literature that use 55,000 samples. Test error of all models is estimated using the validation set of MNIST. We use the CIFAR-10 dataset for experiments using two architectures, an All-CNN network and a wide residual network. For CIFAR-10, we use 50, 000 samples for training and 10, 000 samples for estimating the test error. No data augmentation is performed for MNIST, for CIFAR-10 we randomly flip images (left to right) with probability 0.5 and select random crops of size 32$\times$32 after adding a padding of 4 pixels on the width and height.

\paragraph{Architectures} 
For experiments on MNIST, we use LeNet-5 (this is a network with two convolutional layers of 20 and 50 channels respectively, both of 5$\times$5 kernel size, and a fully-connected layer with 500 hidden neurons) and fully-connected net with one or two layers and 600 or 1200 neurons on each layer. The latter are denoted as FC-600-1, or FC-1200-2 in our experimental section. For CIFAR-10, we use ALL-CNN (in order to reduce the number of weights, we reduced the number of channels in the first set of blocks to 64, and in the second set of blocks to 128; this is down from 96 and 192 respectively in the original network) and wide residual net with depth 10 and a widening factor of 8. In the latter case, in order to reduce the number of weights which makes computing Hessian amenable, we reduce the number of channels in each block of the WRN to [4, 32, 64, 128], down from [16, 128, 256, 512] for a widen factor of 8.

\paragraph{Training procedure}
We train for 30 epochs on MNIST and for 100 epochs on CIFAR-10. The batch-size is fixed to 500 for both datasets. For all experiments with train with Adam and reduce the learning rate using a cosine annealing schedule starting from an initial learning rate of $10^{-3}$ and ending at a learning rate of $10^{-5}$.

\paragraph{Constructing the v2 model in~\cref{fig:overlap_fim_end_init}}
We construct the v2 model by training in two phases. The first phase proceeds as usual: we initialize the model at $w_0$ and train as discussed above to obtain the trained weights $w^1$. In the second phase, and training further for 20 epochs with an objective that is the sum of the original training objective and an addition term spring-force-like term:
\[
    \ehn + \a\norm{w-w_0}_2^2.
\]
The second term forces the weight updates the reduce the Euclidean distance with respect to $w_0$.  The coefficient $\a$ is set to be twice that of the learning rate.

\paragraph{Hyperparameters for optimizing the PAC-Bayes bound}
In Methods 2, 3, 4, we choose $b = 0.01$, $c = 0.1$ for the penalty of the scaling parameters in the prior. In method 1, we choose $b = 0.1$, $c = 0.05$. For all PAC-Bayes bound optimization experiments, we use confidence parameter $\d = 0.025$.

\paragraph{Optimizing the PAC-Bayes bound}
We use batch size of 1100, we draw 150 samples from the posterior $Q$ to estimate $\eqn$ for each weight update; see~\cref{s:app:working_efficiently_bayesian} for some more implementation details of how to compute a large number of samples efficiently. Adam is used to optimize the PAC-Bayes bound. For Methods 2, 3, 4, we first train for 100 epochs with learning rate $10^{-3}$ and train for another 150 epochs while decaying the learning rate by a multiplicative factor of 0.95 every 5 epochs. We found that for this problem, having a constant learning rate at the beginning is beneficial, instead of decaying the learning rate immediately, say using a cosine schedule. For the reproduction of the approach of~\citet{dziugaiteComputingNonvacuousGeneralization2017} (which we denote as $\diag(\S_q) = \L$), we train for 300 epochs for the second phase with decaying learning rate.

\paragraph{Atypical problems}
For atypical problems in, we constructed a training set of 50,000 samples and a validation set of 10,000 samples. Inputs $x_i \in \reals^{200}$ were generated from distribution $N(0, \L)$ where $\L = (\l_1, ..., \l_{200})$. We set $\l_i = b \exp(-c i)$ where and $b/c = 50$; fixing the ratio $b/c$ to be a constant keeps the trace of the data correlation matrix to be about the same for different values of $c$. Labels were generated by $y_i = \argmax_{y \in [m]} p^t_w(y | x_i)$, where $p^t_w$ is the teacher network randomly initialized with one hidden layer and ten output classes. We train fully-connected networks on these synthetic datasets for 50 epochs; Adam is used with a batch-size of 500 and a cosine learning rate schedule with learning rate that ranges from $10^{-3}$ to $10^{-5}$.

We constructed datasets of Gaussian inputs of varying degrees of sloppiness by selecting decay patterns for the eigenvalues of a diagonal data correlation matrix. For $n^{-1} \diag(X X^\top) = \L$ where $\L = (\l_1, \ldots, \l_d)$ are eigenvalues in descending order, we set $\l_i = b \exp(-c i)$ where $b, c$ are constants. The trace of this correlation matrix is roughly $b/c$ which we keep constant for different datasets. Larger the value of the ``sloppy factor'' $c$, more sharp the decay for the eigenspectrum of the data matrix. We randomly initialize a two layer fully-connected neural network with 10 output classes (called the teacher) and use it to label a dataset of such inputs. Note that since the teacher's weights in the first layer multiply the inputs, the correlation matrix of the first layer activations is non-diagonal and we are not being unduly restrictive in picking a diagonal data correlation matrix. We then fit student networks (fully-connected networks with two layers) on this data until they interpolate on the training dataset. Our goal is to study (i) how the various quantities discussed in this paper, e.g., the Hessian, FIM, activations, activation gradients, logit Jacobians , depend upon the sloppiness of the data matrix; (ii) whether the student can interpolate on sloppy datasets without over fitting. \cref{fig:intro} shows the results of the experiment.

\section{Calculation of the effective dimensionality of a deep network}
\label{s:app:effective_dim}

\subsection[PAC-Bayes bounds]{PAC-Bayes bounds}



\begin{theorem}[PAC-Bayes generalization bound~\citet{mcallester1999pac, langford2001bounds}]
For every $\delta > 0$, $n \in \naturals$, distribution $D$ on $\reals^k \times \cbr{0, 1}^m$, and distribution $P$ on $\HH$, with probability at least $1-\delta$ over $D_n \sim D^n$, for all distributions $Q$ on $\HH$,
\[
    \kl(\hat{e}(Q, D_n), e(Q)) \leq \frac{\KL(Q , P) + \log \frac{n}{\delta}}{n-1}
\]
\end{theorem}

We have the following lower-bound from Pinksker's inequality on the KL-divergence between two Bernoulli random variables:
\[
    2(q-p)^2 \leq \kl(q,p).
\]
We can invert this inequality to get
\[
    \kl^{-1} (q, p) \leq q + \sqrt{p/2}.
\]
When this is substituted into the above PAC-Bayes bound~\cref{eq:pac_bayes}, we have
\[
    e(Q) \leq \hat{e}(Q) + \sqrt{\f{\KL(Q, P) + \log (\f{n}{\delta})}{2(n-1)}}.
\]
Since
\[
    \ind{y_i \neq \argmax_y(p_w(y \mid x_i))} \leq -\frac{1}{\log 2} \log p_w(y_i|x_i)
\]
we also have
\[
    \hat{e}(Q) \leq \breve{e}(Q).
\]
Now set $\e = c \exp(j/b)$, for $j \in \mathbb{N}$ and for a fixed $b, c \geq 0$, by the calculations in~\cref{s:penalty}, we see that
\[
    e(Q) \leq \Breve{e}(Q) + \sqrt{\frac{\KL(Q, P) + 2\log(b\log \frac{c}{\e}) + \log (\frac{\pi^2 n}{6\delta})}{2(n-1)}},
\]
holds with probability $1-\delta$.

\subsection[Calculation inverse posterior covariance]{Calculation for the closed form expression for eigenvalues of the inverse posterior covariance in~\cref{eq:lbar_4nm1}}
\label{s:app:lbar_4nm1}

The KL-divergence between two multivariate Gaussians $Q = N(\mu_q, \S_q)$, $P = N(\mu_p, \S_p)$ be two multivariate Gaussians is
\beq{
\label{eq:KL_gaussian}
    \KL(Q,P) = \frac{1}{2} \left( \tr(\S_p^{-1}\S_q)-p + (\mu_p-\mu_q)^\top \S_p^{-1}(\mu_p-\mu_q) + \log(\frac{\det \S_p }{\det \S_q}) \right).
}
In order to compute the inverse posterior covariance that minimizes the right-hand side of the PAC-Bayes bound, we would like to solve the problem
\[
    \aed{
    \text{minimize} \quad L(\S_q) &:= \breve{e}(Q, D_n) + \f{\KL(Q,P)}{2(n-1)}\\
    \text{such that} \quad Q &= N(w, \S_q)\\
    \text{and}\qquad \S_q &\succeq 0.
    }
\]
Observe that
\[
    \aed{
    \Breve{e}(h_{w'}, D_n) &= \lhn + \f{1}{2} \inner{w'-w}{H_w (w'-w)}.\\
    P &= N(w_0, \e^{-1}I).
    }
\]
For $P_w = N(w, \e^{-1} I)$, we have
\[
    \KL(Q,P) = \KL(Q, P_w) + \frac{\e}{2} \norm{w-w_0}^2
\]
Hence,
\[
    \aed{
        L(\S_q) &= \int Q(w') \Breve{e}(h_{w'}, D_n) \dd{w'} + \frac{1}{2(n-1)} \int Q(w') \log \frac{Q(w')}{P_w(w')} \dd{w'} + \frac{\e}{4(n-1)} \norm{w-w_0}^2  \\
        &= \frac{1}{2(n-1)} \int \rbr{-\log \exp (-2(n-1)\Breve{e}(w', D_n)) + \log \frac{Q(w')}{P_w(w')}} Q(w') \dd{w'} + \frac{\e}{4(n-1)} \norm{w-w_0}^2 \\
        &= \frac{1}{2(n-1)} \int \rbr{\log \frac{Q(w')}{\exp(-2(n-1)\Breve{e}(w', D_n))P_w(w') / Z} - \log Z} Q(w') \dd{w'} + \frac{\e}{4(n-1)} \norm{w-w_0}^2 \\
        &= \frac{1}{2(n-1)} \rbr{\KL(Q, B) - \log Z} + \frac{\e}{4(n-1)} \norm{w-w_0}^2,
    }
\]
where we have defined
\[
    \aed{
    B(w') &= \exp(-2(n-1)\Breve{e}(w', D_n))P_w(w')/Z, \qq{and}\\
    Z &= \int \exp(-2(n-1)\Breve{e}(w', D_n))P_w(w') \dd{w'}.
    }
\]
We can now see that $L(\S_q)$ attains a minimum when 
\beq{\label{eq:optimal_non_quadratic}
Q = B \propto \exp(-2(n-1)\Breve{e}(w', D_n)) P_w(w')
}
or $\S_q^{-1} = 2(n-1)H_w + \e\ I$, in other words,
\[
    \S_q = U_w (\Lbar)^{-1} U_w^\top,
\]
where
\[
    \lbar_i = 2(n-1) \l_i + \e\quad \forall i \leq p.
\]

\subsection[Calculation for (15)]{Calculation for~\cref{eq:effective_dim_order}}
\label{s:app:effective_dim_calculation}

Recall that the effective dimensionality of a model at a local minimum $w$ is the number of eigenvalues of the Hessian with magnitude at least $\f{\e}{2(n-1)}$, i.e.,
\[
    p(n, \e) = \sum_{i=1}^p \ind{\abr{\l_i} \geq \f{\e}{2(n-1)}},
\]
The strength of the model at $w$ is

\[
    s(n, \e) = \sum_{i=1}^{p(n, \e)} 1+\log \rbr{\frac{2(n-1) \l_i}{\e} + 1}.
\]
We assume that $c(H_w, p(n, \e)) >0$. i.e., denote $c(H_w, p(n, \e))$ as $c(n, \e)$
\beqs{
    \l_i \leq \f{\e}{2(n-1)} \exp(-c(n, \e) (i - p(n,\e)))
}
We can also assume a weaker version of this decay pattern,
\[
    \sum_{i = p(n, \e)+1}^p \l_i = \frac{\e}{2(n-1)c(n, \e)}.
\]
We approximate the training objective in the neighborhood of $w$ as 
\[
    \Breve{e}(h_{w'}, D_n) = \Breve{e}(h_w, D_n) + \f 1 2 \inner{w'-w}{H_w(w'-w)}.
\]
and we assume that the model at $w$ is a interpolation solution.
In~\cref{s:pac_bayes_bounds}, for the posterior $Q = N(w, \S_q)$ that maximizes the loose version of the PAC-Bayes bound \cref{eq:pac_bayes},  where
\[
    \aed{
        \S_q &= U_w {\Lbar}^{-1} {U_w}^\top,\\
        \lbar_i &= 2(n-1) \l_i + \e.
    }
\]
We can now calculate
\[
    \aed{
        \lqn - \lhn &= \f{1}{2} \sum_{i=1}^p \f{\l_i}{\lbar_i}\\
        &\leq \f{p(n,\e) + 1/c(n, \e)}{4(n-1)}, \text{ and}
    }
\]
\[
    \aed{
        \f{\KL(Q, P)}{2 (n-1)} &= \f{1}{4(n-1)} \rbr{\e \norm{w-w_0}^2 - p + \sum_{i=1}^p \log \f{\lbar_i}{\e} + \f{\e}{\lbar_i}}\\
        &\leq \f{1}{4(n-1)} \rbr{\e \norm{w-w_0}^2 + \sum_{i=1}^{p(n,\e)} \log \rbr{\f{2(n-1)\l_i}{\e} +1} +
                \sum_{i=p(n,\e)+1}^p \f{2(n-1) \l_i}{\e}}\\
       &\leq \f{1}{4(n-1)} \rbr{\e \norm{w-w_0}^2 + \sum_{i=1}^{p(n,\e)} \log \rbr{\f{2(n-1)\l_i}{\e} +1} +
                \f{1}{c(n, \e)}}, \text{ hence}
    }
\]
\[
    \breve{e}(Q, D_n) + \frac{\KL(Q, P)}{2(n-1)} \leq \frac{s(n, \e) + 2/c(n, \e) + \e \norm{w-w_0}^2}{4(n-1)}.
\]
For the KL-term, in the first inequality we have used the fact that $\log(1+x) \leq x$ to split the first summation into two parts; in the second inequality we have used the assumption that the eigenspectrum is sloppy to sum the series from $i=p(n, \e)+1$; the latter is also used in the inequality for the gap in the loss.

\section[Proofs of Lemmas in 3.1]{Proofs of Lemmas in~\cref{s:sloppy_input_sloppy_fim_hessian}}
\label{s:app:proofs}

We use $\E$ to denote the expectation over inputs $x$. The following lemmas holds for all distribution of $x$. In particular, we can choose the distribution of $x$ to be the point mass distribution on the dataset $D_n$, i.e. $x\sim \frac{1}{n} \sum_{i=1}^n \delta_{x_i}$, in this case, $\mathbb{E}\sbr{xx^\top} = \frac{1}{n}XX^\top \in \mathbb{R}^{d\times d}$ is the input corelation matrix.

The following lemma bounds the trace of the activation correlations and the norm of the gradient of each logit with respect to the activations.
\begin{lemma}[Bounding the trace of the correlations of activations and norm of activation gradients]
\label{lem:bound_xcorr_norm}
We have
\beq{
    \textstyle \tr \rbr{\E \sbr{h^k {h^k}^\top}} \leq a^2 \norm{w^{k-1}}_2^2 \tr \rbr{ \E \sbr{h^{k-1} {h^{k-1}}^\top}},
    \label{eq:trace_activation_correlations}
}
and
\beq{
    \norm{\dv{z_i}{h^k}}_2 \leq a \norm{\dv{z_i}{h^{k+1}}}_2 \norm{w^k}_2.
    \label{eq:norm_activation_gradients}
}
\end{lemma}

\begin{proof}[Proof of~\cref{lem:bound_xcorr_norm}]
\label{s:app:proof:lem:bound_xcorr_norm}
For the first inequality in~\cref{eq:trace_activation_correlations}, observe that
\[
    \aed{
        \tr \rbr{\E \sbr{h^k {h^k}^\top}} &\leq \sum_{j=1}^{d_k} \E \sbr{\s(u^k_j)^2}\\
        &\leq a^2 \sum_{j=1}^{d_k} \E \sbr{(u^k_j)^2}\\
        &= a^2 \tr \rbr{ \E\sbr{u^k {u^k}^\top}}\\
        &= a^2 \tr \rbr{ \E \sbr{ \rbr{ w^{k-1} h^{k-1}} \rbr{w^{k-1} {h^{k-1}}^\top}}}\\
        &= a^2 \tr \rbr{w^{k-1} \E\sbr{h^{k-1} {{h^{k-1}}^\top}} {w^{k-1}}^\top}\\
        &\leq a^2 \norm{w^{k-1}}^2_2 \tr \rbr{ \E \sbr{h^{k-1} {{h^{k-1}}^\top}} }.
    }
\]

For the second inequality in~\cref{eq:norm_activation_gradients}, observe that
\[
    \aed{
        \dv{z_i}{h^k} &= \dv{z_i}{u^{k+1}} w^k\\
        &= a \rbr{\dv{z_i}{h^{k+1}} \ones_{u^{k+1} \geq 0}} w^k\\
        \implies \norm{\dv{z_i}{h^k}}_2 &\leq a \norm{\dv{z_i}{h^{k+1}}}_2 \norm{w^k}_2.
    }
\]
where $\ones_{\text{cond}}$ is a vector of 1s at elements where the condition is true.
\end{proof}

The above inequalities can be used in~\cref{lem:outer_product_logit_gradients} to bound the trace of the gradient correlation of any logit $z_i$ with respect to weights of a layer $w^k$.
\begin{lemma}[Bounding the trace of the correlation sum-of-logit Jacobian]
\label{lem:outer_product_logit_gradients}
For logit $z_i$, $i = 1, ..., m$
\beq{
    \tr \rbr{ \E \sbr{ \dv{z_i}{w^k} \dv{z_i}{w^k}^\top }} \leq a^{2 L}  \tr \rbr{ \E \sbr {x x^\top} } \prod_{j=0, j \neq k}^{L} \norm{w^j}^2_2.
    \label{eq:outer_product_logit_gradients}
}
for $k = 0, ..., L$. As a result, 
\[
    \tr \rbr{\E \sbr{ \dv{z_i}{w} \dv{z_i}{w}^\top }} \leq a^{2 L} \tr \rbr{ \E \sbr {x x^\top} } \prod_{j=0}^{L} \norm{w^j}^2_2 \rbr{\sum_{j=0}^{L} \frac{1}{\norm{w^j}_2^2}}.
\]
\end{lemma}

\begin{proof}[Proof of~\cref{lem:outer_product_logit_gradients}]
\label{s:app:proof:lem:outer_product_logit_gradients}
The proof follows via an application of~\cref{lem:bound_xcorr_norm}. For $k = 0, 1, ..., L-1$,
\[
    \aed{
        \tr \rbr{ \E \sbr{ \dv{z_i}{w^k} \dv{z_i}{w^k}^\top}} &= \tr\rbr{\E \sbr{\dv{z_i}{u^{k+1}}\dv{z_i}{u^{k+1}}^\top \otimes h^k {h^k}^\top}} \\
        &= \E \sbr{ \tr \rbr{\dv{z_i}{u^{k+1}}\dv{z_i}{u^{k+1}}^\top} \tr\rbr{h^k {h^k}^\top}} \\
        &\leq a^2 \norm{\dv{z_i}{h^{k+1}}}_2^2 \tr \rbr{ \E \sbr{ h^k {h^k}^\top}}\\
        &\leq a^2 \norm{\dv{z_i}{h^L}}_2^2 \rbr{\prod_{j=k+1}^{L-1} \norm{w^j}^2_2} a^{2(L-k-1)}\\
        &\qquad a^{2k} \prod_{j=0}^{k-1} \norm{w^j}^2_2 \tr \rbr{ \E \sbr{x x^\top}} \\
        &\leq a^{2 L}  \tr \rbr{ \E \sbr {x x^\top} } \prod_{j=0, j \neq k}^{L} \norm{w^j}^2_2.
    }
\]
The third line comes from the fact that the matrix $\dv{z_i}{u^{k+1}} \dv{z_i}{u^{k+1}}^\top$ is rank one and its trace is the same as 2-norm. The last inequality comes from the fact that $\norm{w_i^L}_2 \leq \norm{w^L}_2$. For $k=L$,
\[
    \aed{
        \tr \rbr{ \E \sbr{ \dv{z_i}{w^L} \dv{z_i}{w^L}^\top}} &= \tr \rbr{ \E \sbr{ \dv{z_i}{w_i^L} \dv{z_i}{w_i^L}^\top}} \\
        &= \tr\rbr{\E \sbr{h^L {h^L}^\top}} \\
        &\leq a^{2 L}  \tr \rbr{ \E \sbr {x x^\top} } \prod_{j=0}^{L-1} \norm{w^j}^2_2.
    }
\]
\end{proof}

\begin{proof}[Proof of~\cref{lem:sloppy_input_sloppy_fim_hessian}]
\label{s:app:proof:lem:sloppy_input_sloppy_fim_hessian}
We first calculate an inequality for the Fisher Information Matrix (FIM)
\beqs{
    \aed{
        F&= \E \sbr{ \sum_{y=1}^m p_w(y \mid x) (\partial_w \log p_w(y \mid x)) (\partial_w \log p_w(y \mid x))^\top }\\
        &= \E \sbr{\partial_w z \sbr{\sum_{y=1}^m p_w(y\mid x) \dv{\log p_w(y\mid x)}{z} \dv{\log p_w(y\mid x)}{z}^\top} \partial_w z^\top}
    }
    \label{eq:app:f}
}
For an output distribution $p_w(y \mid x)$ obtained using the softmax operator on the logits $z_y$
\[
    p_y \equiv p_w(y \mid x) = \f{e^{z_y}}{\sum_{y'} e^{z_{y'}}}
\]
we have
\[
    \dv{z}\log p_w(y\mid x) = e_y-p
\]
where $e_y$ is the one-hot vector of the class $y$ and $p = [p_1, ..., p_m]$.
\[
    \aed{
        \sum_{y=1}^m p_w(y\mid x) \dv{\log p_w(y\mid x)}{z} \dv{\log p_w(y\mid x)}{z}^\top
        &\preceq \sum_{y=1}^mp_w(y\mid x) \norm{\dv{\log p_w(y\mid x)}{z}}_2^2 I \\
        &= (1 - \norm{p}_2^2)I \\
        &\preceq I
    }
\]
Hence we have 
\[
    F \preceq \E\sbr{\rbr{\partial_w z} \rbr{\partial_w z}^\top}.
\]

In the case of the Hessian for the cross-entropy loss we make a similar calculation following the calculation of~\citet{fortEmergentPropertiesLocal2019}. For the calculation of Hessian, the expectation $\E$ denotes the expectation with respect to inputs and labels in the training set. We write
\[
    \aed{
   (\log2) H &\approx \E\sbr{\rbr{\partial_w z} \nabla_z^2 (-\log p_w(y \mid x)) \rbr{\partial_w z}^\top}
    \label{eq:app:h_ab}\\
    &=  \E \sbr{\rbr{\partial_w z} \rbr{\text{diag}(p) - pp^\top} \rbr{\partial_w z}^\top}\\
    & \preceq \E \sbr{\rbr{\partial_w z} \rbr{\text{diag}(p)} \rbr{\partial_w z}^\top}\\
    & \preceq \E \sbr{\rbr{\partial_w z}  \rbr{\partial_w z}^\top}.
    }
\]
In the above calculation, we have kept only the so-called G-term of the Hessian and neglected an additional H-term.
\[
    \E \sbr{ \sum_{i=1}^m
            \rbr{y_i-p_i} \pdv{z_i}{w_\a}{w_\b}}
\]
which is typically small in practice for a well-trained network because the terms $1 - p_i$ are close to zero for all logits~\citep{papyan2019measurements,sagun2016singularity}($\mathbb{E}[\sum_{i=1}^m|y_i-p_i|$ is $5.32\times 10^{-8}$ for FC-600-2 on MNIST). Hence, both $\tr\rbr{F}$ and $(\log2)\tr\rbr{H}$ can be bounded by
\beq{
    \label{eq:trace_bound_HF}
    \tr \rbr{F}, (\log2)\tr \rbr{H} \leq \sum_{i=1}^m \E\sbr{\dv{z_i}{w} \dv{z_i}{w}^\top} \leq
    m a^{2 L} \tr \rbr{ \E \sbr {x x^\top} } \prod_{j=0}^{L} \norm{w^j}^2_2 \rbr{\sum_{j=0}^{L} \frac{1}{\norm{w^j}_2^2}}.
}
Notice that the $\log2$ factor in front of $\tr(H)$ comes from the rescaling factor in the definition of $\Breve{e}(h_w, D_n)$.
\end{proof}

\begin{remark}
The G-term is always positive semi-definite since the output distribution $p \in \mathbb{R}^C$ is always convex on the logits $z \in \mathbb{R}^C$, i.e., $\rbr{- \log \rbr{\frac{e^{z_y}}{\sum_{y'e^{z_{y'}}}}}}_{y=1}^C$ is convex in $z$.
\end{remark}


\begin{remark}
Empirically, the trace of FIM and Hessian at the end of training (\cref{fig:full_vs_kfac}) is usually much smaller than the trace of correlation matrix of logit Jacobians (\cref{fig: logit_compare}). In this case, the prediction of the bound in~\cref{eq:trace_bound_HF} seems very loose. However from the above calculation, we also know that
\[
    \aed{
    \tr(F) &\leq (1-\norm{p}_2^2)\ \tr \rbr{ \sum_{i=1}^m \E \sbr{\dv{z_i}{w} \dv{z_i}{w}^\top}},\\
    \tr(H) &\leq \tr \rbr{ \E \sbr{\rbr{\partial_w z} \rbr{\diag (p) - pp^\top} \rbr{\partial_w z}^\top}}.
    }
\]
For trained network that predicts accurately, we usually get the probabilities $p$ that are very close to one-hot vectors of the correct classes. In this case, both $1-\norm{p}_2^2$ and $\diag(p) - p p^\top$ are close to zero. This explains why in our experiments the trace of $F$ and $H$ at the end of training are much smaller than that of logit Jacobians.
\end{remark}

\begin{proof}[Proof of~\cref{lem:fim_sloppy}]
\label{s:app:proof:lem:fim_sloppy}

The proof depends upon Weyl's inequality to control the eigenvalues of the sum of Hermitian matrices. It states that for Hermitian matrices $A, B, C \in \reals^{p \times p}$, if $C = A + B$, then
\beq{
    \aed{
        \scalemath{1}
        {\l_{i+j-1}(C) \leq \l_i(A) + \l_j(B)}, \quad
        {\l_{p-i-j}(C) \geq \l_{p-i}(A) + \l_{p-j}(B)
        }
    }
    \label{eq:weyls_inequality}
}
for all $ 1 \leq i, j \leq p$. In particular if $B \succeq 0$, then $\l_i(C) \geq \l_i(A)$ for all $i\leq p$.

We can now write,
\[
    \aed{
    \E \sbr{ \dv{z_i}{w^k} {\dv{z_i}{w^k}}^\top} &= \E \sbr{ \rbr{\dv{z_i}{h^{k+1}} \odot \dv{h^{k+1}}{u^{k+1}}} \rbr{\dv{z_i}{h^{k+1}} \odot \dv{h^{k+1}}{u^{k+1}}}^\top \otimes h^k{h^k}^\top}\\
    &\preceq \E\sbr{a^2\norm{\dv{z_i}{h^{k+1}}}^2 I_{d_{k+1}} \otimes h^k{h^k}^\top}\\
    &= a^2\norm{\dv{z_i}{h^{k+1}}}^2 I_{d_{k+1}} \otimes \E \sbr{h^k{h^k}^\top}\\
    &= a^{2(L-k)} \rbr{\prod_{j=k+1}^L ||w_j||^2} I_{d_{k+1}} \otimes \E \sbr{h^k{h^k}^\top}
    }
\]
Hence, by~\cref{eq:weyls_inequality}
\[
    \spec \rbr{\E \sbr{\dv{z_i}{w_k} \dv{z_i}{w_k}^\top} }  \preceq \spec\rbr{a^{2(L-k)} \prod_{j=k+1}^L ||w_j||^2 I_{d_{k+1}} \otimes \E \sbr{h^k{h^k}^\top}}
\]
so we have
\[
    \spec \rbr{\E \sbr{\dv{z_i}{w_k} \dv{z_i}{w_k}^\top} }
    \preceq a^{2(L-k)} \prod_{j=k+1}^L ||w_j||^2 \spec \rbr{I_{d_{k+1}}} \otimes \spec(\E \sbr{h^k{h^k}^\top})
\]
\end{proof}

\begin{remark}[Modification using sloppiness of activation gradients]
\label{rem:activation_gradients}
\cref{fig:intro} shows that the slope of decay of FIM and the activations are essentially the same.  However, in~\cref{eq:fim_sloppy} if $\spec\rbr{\E\sbr{h^k {h^k}^\top}}$ decays as $\OO(\exp(-c i))$, the decay of $\spec \rbr{\E \sbr{\dv{z_i}{w^k} {\dv{z_i}{w^k}}^\top}}$ is $\OO(\exp(-c i/d_{k+1}))$. This is a loose bound, especially when $d_{k+1}$ is large, e.g., the spectrum could decay much more faster. But note that if we can write a KFAC-approximation
\[
    \aed{
    \E \sbr{\dv{z_i}{w^k} {\dv{z_i}{w^k}}^\top}
    \approx \E \sbr{\dv{z_i}{u^{k+1}} {\dv{z_i}{u^{k+1}}}^\top}}
    \otimes \E\sbr{h^k {h^k}^\top}.
\]
then we obtain a stronger decay for the logit gradient when $d_{k+1}$ is large, if we assume that the activations \emph{gradients} are sloppy. If $\spec \rbr{ \E \sbr{\dv{z_i}{u^{k+1}} \dv{z_i}{u^{k+1}}^\top}}$ decays as $\exp{-c_1i}$ and $\spec\rbr{\E\sbr{h^k{h^k}^\top}}$ decays as $\exp{-c_2j}$, then the $(i+j)^2$th largest eigenvalue of $\E \sbr{\dv{z_i}{w^k} \dv{z_i}{w^k}^\top}$ is smaller than $\exp(-\min\{c_1, c_2\}(i+j))$, hence the $k$th largest eigenvalue of $\E \sbr{\dv{z_i}{w^k} \dv{z_i}{w^k}^\top}$ is smaller than $\exp(-\min\{c_1, c_2\}\sqrt{k})$. Hence, the decay rate of $\spec \rbr{\E \sbr{\dv{z_i}{w^k} \dv{z_i}{w^k}^\top}}$ is $\OO \rbr{\exp(-\min\{c_1, c_2\}\sqrt{k})}$.
\end{remark}

\begin{corollary}
\label{cor: fim_sloppy}
Denote the FIM and Hessian with respect to the $k$th layer $F(w_k), H(w_k)$ respectively, then we have,
\[
   \spec \rbr{F(w_k)}, \spec \rbr{H(w_k)} \preceq 2 m a^{2(L-k)} \prod_{j=k+1}^L \norm{w^j}_2^2 \spec (I_{d_{k+1}}) \otimes \spec \rbr{\mathbb{E}\sbr{h^k{h^k}^\top}}.
\]
As in ~\cref{lem:fim_sloppy}, $\prod_{j=L=1}^L \norm{w^j}_2^2 = 1$.
\end{corollary}

\begin{proof}
From ~\cref{lem:outer_product_logit_gradients} we know that 
\[
F(w_k), H(w_k) \preceq 2 \E \sbr{(\partial_{w_k}z)(\partial_{w_k}z)^\top}
\]
Let $s = \sum_{i=1}^m z_i$ be the sum of logits, then we have
\[
\aed{
F(w_k),  H(w_k) &\preceq 2 \mathbb{E} \sbr{\rbr{\dv{s}{w_k}}\rbr{\dv{s}{w_k}}^\top}\\
& \preceq 2 m a^{2(L-k)} \prod_{j=k+1}^L \norm{w^j}_2^2 \spec (I_{d_{k+1}}) \otimes \spec \rbr{\mathbb{E}\sbr{h^k{h^k}^\top}}
}
\]
The second inequality comes from a similar calculation as in ~\cref{lem:fim_sloppy} for network with one added layer where $h_{L+1} = u_{L+1} = z$, $u_{L+2} = w_{L+1}h_{L+1}$, and $w_{L+1} = [1, ..., 1]$, $\norm{w_{L+1}}_2^2 = m$.
\end{proof}

\section{Technical details of different methods for optimizing the PAC-Bayes bound}
\label{s:app:technical_details_pac_bayes}

We optimize the problem,
\beq{\label{eq:pac-bayes optim}
    \min \breve{e}(Q, D_n) + \sqrt{\frac{\KL(Q, P) + \varphi}{2(n-1)}}
}
where $Q$, $P$ are multivariate normal distribution, $\varphi$ is the penalty we added for including a trainable parameter in prior (say its scale), and $n$ is the number of samples. For Gaussian distributions on the weight space $Q, P$, as we saw in ~\cref{eq:KL_gaussian}, the KL-divergence is
\[
    \aed{
        \f 1 2 \rbr{\tr(\S_p^{-1} \S_q) - p + (w-w_0)^\top \S_p^{-1} (w-w_0) +
        \log (\det \S_p /\det \S_q)}.
    }
\]
The penalty for the case when $P = N(0, \e^{-1} I)$ comes from the union bound over the set $\e = c \exp(j/b)$ for $j \in \naturals$ and is given by
\[
    \varphi = 2 \log \rbr{b \log (c/ \e)} + \log (\pi^2 n/ (6 \delta))
\]
Note that for Method 4, we need more than one trainable parameters for the prior, and the penalty $\varphi$ should also be modified according to ~\cref{s:penalty}. We calculate $\Breve{e}(Q, D_n)$ using Monte Carlo samples from $Q$. After the optimization process, we calculate the PAC-Bayes bound on $\eq$ using
\beq{
    \kl(\hat{e}(Q, D_n), e(Q)) \leq \frac{\KL(Q , P) + \varphi}{n-1},
}\label{eq:pac-bayes evaluate}
which involves finding an approximation of $\kl^{-1}(b, a) := \sup\{a' \in [0, 1]: \kl(b, a') \leq a\}$ (see \cite{dziugaiteComputingNonvacuousGeneralization2017} for details).
We next discuss the various methods for calculating PAC-Bayes bounds developed in the paper and provide their implementation details.

\subsection{Method 1}
\label{s:app:method_1}

The tightest bound in this case is obtained using the v2 model described in~\cref{fig:overlap_fim_end_init} and~\cref{s:app:setup}. To recall, this involves a second post-training phase where the trained model is updated to be closer to the initialization $w_0$. In the context of the PAC-Bayes upper bound, this reduces the distance between the means of the Gaussian prior and posterior. We choose $\S_q$ as in ~\cref{eq:S_q_basic} and~\cref{eq:lbar_4nm1}. For $\e = c \exp(j / b)$ and $j = 1, \ldots, 60$, we evaluate $\KL(Q, P)$ by using ~\cref{eq:KL_gaussian}, and $\eqn$ is estimated by sampling. The covariance $\S_q$ is approximated by the top eigenvalues and eigenvectors of the Hessian as discussed in~\cref{s:app:top_spaces}. The PAC-Bayes bound is calculated by~\cref{eq:pac-bayes optim} and we choose the smallest bound among all choices of $\e$.

We also set $\S_q = U_w \Lbar U_w^\top$ and calculate $\Bar{\L}$ by directly minimizing~\cref{eq:pac-bayes optim} where the variables of optimization are $\Bar{\l}_i$ for $i \leq k$ using nonlinear optimization in scipy (using the BFGS algorithm), and the PAC-Bayes bound is calculated in the same way as above. This is denoted as Method 5 (Numerical) in~\cref{tab:pac_bayes_full}.

For comparison, we also choose $\S_q = \e^{-1} I$ and calculate the PAC-Bayes bound. This is denoted as Method 6 (Isotropic) in~\cref{tab:pac_bayes_full}.

\subsection{Methods 2 and 3}
\label{s:app:methods_2_3}

We choose $P$ and $Q$ as described in ~\cref{s:non_analytical_pac_bayes}. We set $\e^{-1} = \exp(2 \rho)$ , $\bar{\L}_w = \exp(2 \xi)$. The parameters $\rho$, $w$, $\xi$ are optimized while optimizing the PAC-Bayes upper bound. We initialize $\e^{-1}$ at $\exp(-6)$ and $\bar{\L}_w$ at $(\L^F + \e^{-1}) / 10$ where $\L^F$ are the eigenvalues of the FIM at initialization. For fully-connected networks and LeNet, we evaluate $\eqn$ using the methods described in ~\cref{s:KFAC} and~\cref{s:app:top_spaces} respectively.

We use the Gauss-Newton matrix as an approximation of the FIM for Method 2.

\subsection{Method 4}

We choose $P$ and $Q$ as described in ~\cref{s:non_analytical_pac_bayes}. We set $a = \exp(2 \rho_1)$, $\e^{-1} = \exp(2 \rho_2)$, $\s = \exp(2 \xi)$ and train parameters $\rho_1$, $\rho_2$, $w$, $\xi$. In our experiments, $\e^{-1}$ is initialized to $\exp(-6)$, $a$ is initialized to $\exp(-1)$ and $\S_q$ is initialized to be $(a F_{w_0} + \e) / 10$. In this case,
\[
    \KL(Q, P) = \frac{1}{2}\rbr{\sum_i\frac{\sigma_i}{a\l^F_i + \e^{-1}} - d + (w-w_0)^\top (a F_{w_0} + \e^{-1})^{-1}(w-w_0) + \sum_i \log \frac{a\l^F_i + \e^{-1}}{\s_i}}
\]
where $\l_i^F$ are eigenvalues of $F_{w_0}$. For fully-connected networks and LeNet, we approximate $(w-w_0)^\top (a F_{w_0} + \e^{-1})^{-1}(w-w_0)$ using the methods described in~\cref{s:KFAC} and~\cref{s:app:top_spaces} respectively.

We use the Gauss-Newton matrix as an approximation of the FIM for Method 4.

\subsection{Computing the PAC-Bayes term that corresponds to the distance from initialization}
In Method 4, we need to calculate
\[
    E = (w-w_0)^\top (a F_{w_0} + \e^{-1})^{-1}(w-w_0).
\]
In Methods 2 and 3, we need to sample from a posterior of the form $N(0, U \L U^\top)$ for various different values of $U$ and $\L$. Doing either of these is not easy for high-dimensional weight spaces. We employ two different methods to deal with this problem. For fully-connected networks we use a KFAC approximation of the Hessian/FIM while for LeNet which has much fewer weights, we approximate these matrices using their top few eigenvalues and eigenvectors.

\subsubsection{KFAC approximation of the FIM and Hessian}
\label{s:KFAC}

We approximate the Hessian/FIM by a variation of Kronecker decomposition of block-diagonal Hessian/FIM (KFRA, \cite{ICML-2017-BotevRB}). We use the BACKPACK library for implementing this~\citep{dangel2020backpack}. For the weight of the $k^{\text{th}}$ layer $w_k \in \mathbb{R}^{d_{k+1} \times d_{k}}$, the KFRA approximation of the corresponding block in the Hessian/FIM which is denoted by $F^k$ or $H^k$ can be written as $A_k \otimes B_k$. Denote by $U_{A_k}, U_{B_k}$ the eigenspaces of $A_k$ and $B_k$. To estimate $E$, we can first decompose $E$ as the summation where each term is for a particular layer $k$
\[
    E = \sum_{k=0}^{L} E^k
\]
where
\[
    \aed{
    E^k &= (w^k-w_0^k)^\top (a \rbr{F_{w_0}}^k + \e^{-1})^{-1}(w^k-w^k_0)\\
    &= (w^k-w^k_0)^\top U^k (a \L^k + \e^{-1})^{-1} {U^k}^\top (w^k-w_0^k)\\
    &= \rbr{(w^k-w^k_0)^\top U^k \rbr{a\L^k + \e^{-1}}^{-1/2}} \rbr{(w^k-w^k_0)^\top\ U^k \rbr{a\L^k + \e^{-1}}^{-1/2} }^\top
    }
\]
$(E^k)^{1/2}$ can be calculated by
\[
    \aed{
    {E^k}^{1/2} &= (w^k-w^k_0)^\top U^k \rbr{a\L^k + \e^{-1}}^{-1/2}\\
    &= (U_{A_k}^\top (w^k-w^k_0) U_{B_k}) \odot \rbr{a \L^k + \e^{-1}}^{-1/2}
    }
\]
where in the last line, $(w^k-w^k_0) \in \reals^{d_{k+1} \times d_k}$. We use $\odot$ to denote element wise multiplication. $U_{A_k}^T (w^k-w^k_0) U_{B_k}$ can now be easily calculated using the KFAC factors.

To sample from the posterior $N(w, U\L U^\top)$, we can concatenate the samples of the weights of each layer. We first sample $r^k \sim N(0, I_{d_k})$, then calculate $\sqrt{\L^k} \odot r_k$ and thereby
\[
    \nu_k := U_k \rbr{\sqrt{\L^k} \odot r_k} = U_{A_k} \rbr{\sqrt{\L^k} \odot r_k} U_{B_k}^\top.
\]
The final sample is therefore $w + [\nu_1, ..., \nu_k]$ which is distributed as $N(w, U \L U^\top)$.

\subsubsection{Approximate FIM and Hessian using its top eigenvalues and eigenvectors}
\label{s:app:top_spaces}

For symmetric $\S$ with orthogonal decomposition $\S = U\L U^\top$, $U = [U_1, U_2]$, $\L = \diag(\L_1, \L_2)$, we have
\[
    \aed{
    \S &= U_1 \L_1 U_1^\top + U_2 \L_2 U_2^\top\\
    \text{where}\ I &= U_1U_1^\top + U_2U_2^\top.
    }
\]
In this case, to calculate $E$, we approximate $a F_{w_0} + \e^{-1}$ by
\[
    a F_{w_0} + \e^{-1} = U_1(a\L_1 + \e_1^{-1}) U_1^\top + \e_2^{-1} U_2 U_2^\top
\]
where $\L_1, U_1$ are the stiff (largest $k$) eigenvalues and corresponding eigenvectors for $F_{w_0}$ and $U_2, \L_2$ are the sloppy ones. Notice that we use two scalar parameters $\e_1$ and $\e_2$ to set the additive constant in the prior covariance.
\[
    \aed{
        E &= (w-w_0)^\top U_1 (a\L_1 + \e_1^{-1})^{-1} U_1^\top (w-w_0) + \e_2 (w-w_0)^\top U_2 U_2^\top (w-w_0)\\
        &= (w-w_0)^\top U_1(a\L_1 + \e_1^{-1})^{-1} U_1^\top (w-w_0) + \e_2 \rbr{\norm{w-w_0}_2^2-(w-w_0)^\top U_1 U_1^\top (w-w_0)}
    }
\]
Notice that the term $(w-w_0)^\top U_1$ is not hard to calculate because $U_1 \in \reals^{p \times k}$ and since we are choosing the top few eigenvalues of the Hessian/FIM, the value of $k$ is small (about 300).

To sample from the posterior $N(w, U\L U^\top)$, we first set $\L = \diag (\L_1, \L_2)$ where $\L_1$ are the top $k$ stiff eigenvectors and $\L_2$ are the $p-k$ other eigenvectors. Correspondingly, we have $U = [U_1, U_2]$. We use an isotropic variance for the sloppy subspace and set $\L_2 = \e^{-1} I_{p-k}$. We first sample $r \sim N(0, I_k)$, then calculate
\[
    \aed{
        \nu_1 &= U_1 \sqrt{\L_1} U_1^\top r\\
        \nu_2 &= \e^{-1/2} U_2 U_2^\top r = \e^{-1/2} (r - U_1 U_1^\top r)
    }
\]
Notice that $U_1 U_1^\top r$, and $U_1 \sqrt{\L_1} U_1^\top r$ are easy to calculate. The result $w+ [\nu_1, \nu_2]$ is distributed as $N(w, U\L U^\top)$.

For cases when we recompute the FIM/Hessian while optimizing the PAC-Bayes bound (Method 2 and 3 respectively), we recompute the eigenvalues $\L_1$ and the corresponding eigenvectors $U_1$. Note that the parameter $\e$ in the covariance of the posterior is also optimized when we optimize the PAC-Bayes bound.

\subsection{Optimizing parameters of the prior in the PAC-Bayes bound}
\label{s:penalty}

The prior should be fixed before looking at the training set, but for all methods above, we optimize the scale of the prior. We do this by adding an additional penalty in the KL term. Assume that $a^i$ for $i = 1, \ldots, m'$ are the number of parameters in the prior that we can select, we choose $a^i = (1/c^i) \exp(-j^i / b^i)$ for $j^i \in \mathbb{N}$. We reindex $j^i$ as a single index $k = (\sum_i j^i)^{m'}$, then if the PAC-Bayes bound for each index $k$ is designed to hold with probability at least $1-\frac{6 \delta}{\pi^2k^2}$, then by union bound, it will hold uniformly for all $k \in \mathbb{N}$ with probability at least $1-\delta$.
For a bound that holds with probability $1-\delta'$, the penalty we should add is $\log \frac{n}{\delta'}$, hence, using the relation
\[
    a^i = (1/c^i) \exp(j^i / b^i), \quad \delta' = \frac{6 \delta}{\pi^2 k^2}, \quad k = (\sum_i j^i)^{m'}
\]
we add the penalty
\[
    \varphi(a^1, ..., a^{m'}) = 2 {m'} \log (\sum_i b_i \log (c^ia^i)) + \log \frac{\pi^2 n}{6 \delta}
\]
Similarly, for any positive or negative integer $j^i$, we can set $k = (\sum_i 2|j|^i)^{m'}$ to get the penalty
\[
    \varphi(a^1, ..., a^{m'}) = 2 {m'} \log (2\sum_i \abr{b_i \log(c^i a^i)}) + \log \frac{\pi^2 n}{6 \delta}
\]
In Methods 1, 2, 3 we choose $a^1 = \e^{-1}$, in Method 4, we choose $a^1 = a$ and $a^2 = \e^{-1}$.

\section{Working efficiently with Bayesian deep networks}
\label{s:app:working_efficiently_bayesian}

Typically, a Bayesian neural network is implemented by programming Bayesian variants of standard layers in deep learning. For instance, one defines a BayesianLinear layer which maintains two sets of parameters, the mean weight vector and a standard deviation for each weight. At each forward pass, the layer samples a weight vector using the reparameterization trick to compute the activations. This is a reasonably efficient way to implement a Bayesian neural network but it is cumbersome because code for complex deep network architectures has to be rewritten from scratch to accommodate these Bayesian layers. We noticed that we can use the following trick (this is likely specific to PyTorch) to create a wrapper around any existing deep network code and construct its Bayesian variant.
All our experiments use 150 samples from $Q$ before each update; in comparison typical implementations use 1 sample~\citep{dziugaiteComputingNonvacuousGeneralization2017,wuDissectingHessianUnderstanding2021}. This strategy is potentially useful for other problems as well, e.g., for estimating the prediction uncertainty.

\begin{figure}[!htpb]
\centering
\includegraphics[scale=0.6]{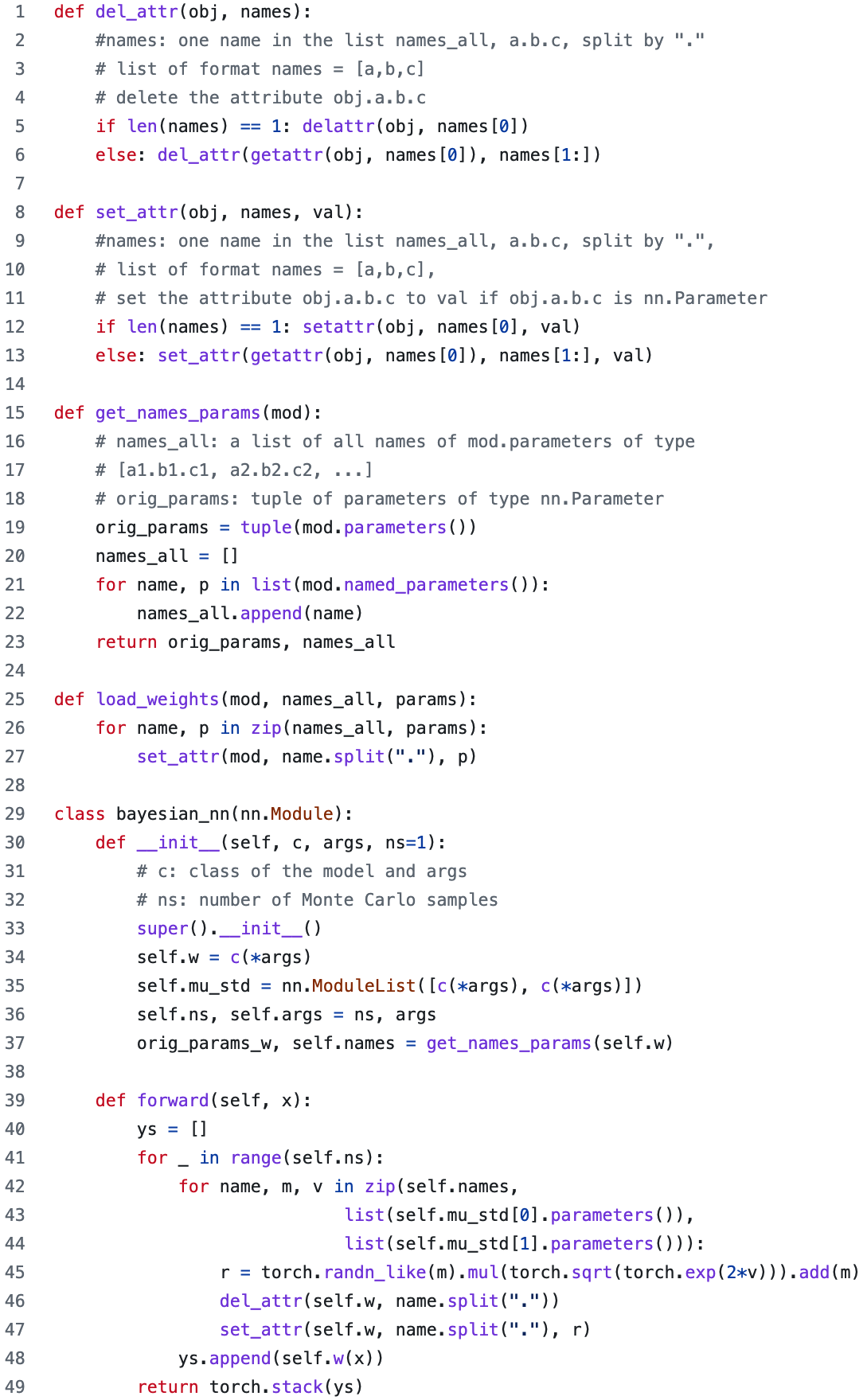}
\label{fig:bayes_nn_code}
\caption{Code for Bayesian neural networks}
\end{figure}

The code shown in~\cref{fig:bayes_nn_code} is adapted from\newline
\href{https://github.com/pytorch/pytorch/blob/master/benchmarks/functional_autograd_benchmark/utils.py}{https://github.com/pytorch/pytorch/blob/master/benchmarks/functional\_autograd\_benchmark/utils.py} and works
by first calculating the reparameterization trick (Line 45) using the mean and (logarithm of the) standard deviation of the weights (self.mu\_std) and then swapping the weight of the actual model (self.w) that performs the forward propagation using the sampled weights.

\section{Full results of PAC-Bayes generalization bounds and effective dimensionalities}
\label{s:app:pac_bayes_full}
We display the extended version of the results of PAC-Bayes bound optimization in~\cref{tab:pac_bayes_full}. Methods 1 and 5 give bounds that are similar to each other: this shows that our analytical expression~\cref{eq:S_q_basic} for the optimal posterior using a loose PAC-Bayes bound under the assumption that the loss is quadratic at the weights at the end of training is an accurate estimate of the optimal posterior~\cref{eq:pac_bayes}. The bound calculated by these two methods is smaller than that of Method 6, which shows that the sloppiness of the Hessian at the end of training ($H_w$) is effective at providing non-vacuous generalization bounds. Using an isotropic posterior in Method 6 produces a remarkably good bound because almost all eigenvalues of $H_w$ for MNIST are small; even the largest eigenvalue is quite small in its magnitude as shown in~\cref{fig:full_vs_kfac}). Methods 1, 5, 6 (which are the three methods that compute a bound without any optimization using the training dataset) give worse bounds than Methods 2, 3, 4 and also the method of~\citet{dziugaiteComputingNonvacuousGeneralization2017}. This is because the approximation
\[
    \Breve{e}(h_{w'}, D_n) = \lhn + \f 1  2 \inner{w'-w}{H_w (w'-w)}.
\]
as we discussed in Method 1 may not be an accurate estimate of $\Breve{e}(w', D_n)$ in the neighborhood of $w$. As we see in ~\cref{s:app:effective_dim}, the posterior that optimizes the loose PAC-Bayes bound \emph{without} the approximation of the quadratic loss instead looks like~\cref{eq:optimal_non_quadratic}. Methods 2--4 which involve optimization of the PAC-Bayes bound capture the optimal posterior better than the one corresponding to the quadratic assumption leads to a tighter PAC-Bayes bound. Method 4 gives the tightest bound since the training predominantly takes place in the stiff subspace of FIM at initialization, and the prior with covariance proportional to FIM puts less penalty than the isotropic prior on the stiff directions. Using posterior with $\text{E}(\S_q) = H_w$ (Method 3, which is similar to~\citet{wuDissectingHessianUnderstanding2021}) works better than a diagonal posterior $\text{E}(\S_q) = \L$ (which is the method in~\citet{dziugaiteComputingNonvacuousGeneralization2017}); this coincides with our calculation in Method 1 (see~\cref{s:pac_bayes_bounds}) that the eigenvectors of the optimal posterior is the same as that of the Hessian $H_w$.

We also calculated the effective dimensionalities, strength and sloppy factor of different models using $\epsilon$ derived in Method 3 (the $\e$ calculated by PAC-Bayes bound optimization can be regarded as a sound choice), the results are displayed in the 4th block of \cref{tab:pac_bayes_full}.

{
\renewcommand{\arraystretch}{0.8}
\begin{table}
    \centering
    \tiny
    \rowcolors{1}{}{black!5}
    \resizebox{0.8\linewidth}{!}{
    \begin{tabular}{lrrrrrr}\toprule
    Quantity/Model &FC-600-1 &FC-600-2 &FC-1200-1 &FC-1200-2 & LeNet \\
    \midrule

    \multicolumn{6}{c}{\textbf{Training and validation error of the trained model}}\\[0.25em]
    $\hat{e}(h_w, D_n)$ &0.0000 &0.0000 &0.0000 &0.0000 &0.0000 \\
    $\log 2 * \Breve{e}(h_w, D_n)$ &0.0008 &0.0000 &0.0010 &0.0000 &0.0000 \\
    $e(h_w)$ &0.0150 &0.0143 &0.0146 &0.0139 &0.0111 \\
    $\log 2 * \Breve{e}(h_w)$ &0.0641 &0.0956 &0.0584 &0.0977 &0.0669 \\
    \midrule

    \multicolumn{6}{c}{\textbf{Analytic (Method 1)}}\\[0.25em]
    $\hat{e}(Q, D_n)$ &0.0901 &0.0766 &0.0534 &0.0678 &0.0074 \\
    $\log 2 * \Breve{e}(Q, D_n)$ &0.2299 &0.1997 &0.1410 &0.1776 &0.0263 \\
    $e(Q)$ &0.0897 &0.0827 &0.0553 &0.0729 &0.0167 \\
    $\log 2 * \Breve{e}(Q)$ &0.2384 &0.2314 &0.1492 &0.2015 &0.0927 \\
    PAC-Bayes bound &0.3241 &0.3794 &0.3509 &0.3915 &0.0572 \\
    $\KL(Q, P)$ &8512.5098 &13417.4023 &14088.1738 &15308.4170 &1965.8048 \\
    $\e$ &199.4836 &401.7107 &328.8929 &443.9590 &36.4424 \\
    \midrule

    \multicolumn{6}{c}{\textbf{$\ev(\S_q)=\ev(F_{w_0})$ (Method 2)}}\\[0.25em]
    $\hat{e}(Q, D_n)$ &0.0309 &0.0288 &0.0267 &0.0298 &0.0053 \\
    $\log 2 * \Breve{e}(Q, D_n)$ &0.0895 &0.0798 &0.0742 &0.0829 &0.0160 \\
    $e(Q)$ &0.0346 &0.0331 &0.0327 &0.0348 &0.0147 \\
    $\log 2 * \Breve{e}(Q)$ &0.0995 &0.0959 &0.0947 &0.0995 &0.0590 \\
    PAC-Bayes bound  &0.1590 &0.1767 &0.1523 &0.2017 &0.0099 \\
    $\KL(Q, P)$ &4772.4854 &5953.1523 &4841.5972 &7268.2832 &46.5822 \\
    \midrule

    \multicolumn{6}{c}{\textbf{$\text{E}(\S_q)=\text{E}(H_w)$ (Method 3, our implementation)}}\\[0.25em]
    $\hat{e}(Q, D_n)$ &0.0202 &0.0165 &0.0169 &0.0178 &0.0043 \\
    $\log 2 * \Breve{e}(Q, D_n)$ &0.0556 &0.0451 &0.0466 &0.0487 &0.0133 \\
    $e(Q)$ &0.0268 &0.0253 &0.0245 &0.0249 &0.0141 \\
    $\log 2 * \Breve{e}(Q)$ &0.0781 &0.0781 &0.0761 &0.0742 &0.0564 \\
    PAC-Bayes bound  &0.1357 &0.1540 &0.1515 &0.1817 &0.0188 \\
    $\KL(Q, P)$ &4645.1128 &6122.5703 &5919.6455 &7589.6387 &430.4026 \\
    $\epsilon$ & 46 & 101 & 53 & 172 & 1360 \\
    $p(n, \epsilon)$ &2301 (0.487\%) & 2429 (0.292 \%) & 2315 (0.245 \%) & 2287 (0.095 \%) & 82 (0.184 \%) \\
    $s(n, \epsilon)$ & 6435 & 6810 & 6420 & 6280 & 231\\
    $1/c(n, \epsilon)$ & 2236 & 2497 & 2604 & 2841 & 38\\
    \midrule

    \multicolumn{6}{c}{\textbf{$\S_p = aF_{w_0} + \e^{-1}, \text{E}(\S_q)=\text{E}(F_{w_0})$ (Method 4)}}\\[0.25em]

    $\hat{e}(Q, D_n)$ &0.0237 &0.0218 &0.0226 &0.0220 &0.0048 \\
    $\log 2 * \Breve{e}(Q, D_n)$ &0.0663 &0.0611 &0.0631 &0.0614 &0.0147 \\
    $e(Q)$ &0.0270 &0.0265 &0.0266 &0.0264 &0.0145 \\
    $\log 2 * \Breve{e}(Q)$ &0.0806 &0.0956 &0.0789 &0.0801 &0.0573 \\
    PAC-Bayes bound &0.1323 &0.1397 &0.1486 &0.1702 &0.0092 \\
    $\KL(Q, P)$ &4090.7241 &4679.0293 &5074.4102 &6369.7505 &23.2886 \\
    \midrule

    \multicolumn{6}{c}{\textbf{$\text{diag}(\S_q) = \L$ (our implementation)}}\\[0.25em]
    $\hat{e}(Q, D_n)$ &0.0283 &0.0249 &0.0284 &0.0285 &0.0079 \\
    $\log 2 * \Breve{e}(Q, D_n)$ &0.0795 &0.0700 &0.0795 &0.0797 &0.0236 \\
    $e(Q)$ &0.0330 &0.0311 &0.0326 &0.0331 &0.0161 \\
    $\log 2 * \Breve{e}(Q)$ &0.0942 &0.0923 &0.0940 &0.0963 &0.0637 \\
    PAC-Bayes bound  &0.1707 &0.1846 &0.1886 &0.2167 &0.0131 \\
    $\KL(Q, P)$ &5674.5186 &6854.7871 &6668.9023 &8332.5869 &37.5598 \\
    \midrule

    \multicolumn{6}{c}{\textbf{Numerical optimization of Method 1 calculations (Method 5)}}\\[0.25em]
    $\hat{e}(Q, D_n)$ &0.0711 &0.0630 &0.0805 &0.0580 &0.0087 \\
    $\log 2 * \Breve{e}(Q, D_n)$ &0.1805 &0.1673 &0.2072 &0.1510 &0.0331 \\
    $e(Q)$ &0.0717 &0.0683 &0.0800 &0.0644 &0.0168 \\
    $\log 2 * \Breve{e}(Q)$ &0.1902 &0.1925 &0.2092 &0.1811 &0.0955 \\
    PAC-Bayes bound &0.3182 &0.3917 &0.3539 &0.4366 &0.0792 \\
    $\KL(Q, P)$ &9920.2510 &15908.7813 &11271.7246 &20162.2891 &2941.7917 \\
    $\e$ &243.6499 &490.6506 &269.2748 &599.2820 &54.3656 \\
    \midrule

    \multicolumn{6}{c}{\textbf{Isotropic Posterior (Method 6)}}\\[0.25em]
    $\hat{e}(Q, D_n)$ &0.0473 &0.0879 &0.0653 &0.0638 &0.0094 \\
    $\log 2 * \Breve{e}(Q, D_n)$ &0.1266 &0.2538 &0.1661 &0.1757 &0.0385 \\
    $e(Q)$ &0.0524 &0.0937 &0.0677 &0.0697 &0.0191 \\
    $\log 2 * \Breve{e}(Q)$ &0.1409 &0.2935 &0.1759 &0.2057 &0.1076 \\
    PAC-Bayes bound &0.3694 &0.5461 &18533.2422 &0.5490 &0.1146 \\
    $\KL(Q, P)$ &16261.0205 &26160.2773 &0.4288 &30034.0645 &4807.3564 \\
    $\e$ &401.7107 &808.9461 &443.9590 &894.0237 &89.6338 \\
    \midrule

    \multicolumn{6}{c}{\textbf{$\text{diag}(\S_q) = \L$ (from~\citet{dziugaiteComputingNonvacuousGeneralization2017})}}\\[0.25em]
    $e(h_w)$ &0.018 &0.016 &0.018 & 0.015 & - \\
    $e(Q)$ &0.034 &0.033 &0.035 & 0.035 & - \\
    PAC-Bayes bound  &0.161 &0.186 &0.179 &0.223 & - \\
    $\KL(Q, P)$ & 5144 & 6534 & 5977 & 8558 & - \\
    \midrule

    \multicolumn{6}{c}{\textbf{$\text{E}(\S_q)=\text{E}(H_w)$ (from \cite{wuDissectingHessianUnderstanding2021})}}\\[0.25em]
    $e(h_w)$ &0.0153 &0.0148 &0.0161 & - & - \\
    $e(Q)$ &0.02347 &0.02523 &0.02316 & - & - \\
    PAC-Bayes bound  &0.1198 &0.1443 &0.1413 &- & - \\
    $\KL(Q, P)$ & 3766.1 & 4956.8 & 5021.1 & - & - \\
    \bottomrule
    \end{tabular}
}
\caption{\textbf{Comparison of PAC-Bayes bounds on MNIST for different methods.} This table is an expansion of ~\cref{tab:pac_bayes}. The 6th block is our reproduction of  \cite{dziugaiteComputingNonvacuousGeneralization2017}, the first, 7th and 8th block corresponds to the three methods of constructing posterior for PAC-Bayes bound without training described in ~\cref{s:app:method_1}.}

\label{tab:pac_bayes_full}
\end{table}
}

\section{Further experimental studies}
\label{s:app:further_expt}

\subsection{Additional results on the sloppiness of different architectures and datasets}
\label{s:app:additional:sloppiness}

\begin{figure}[htpb]
\centering
\includegraphics[width=0.5\linewidth]{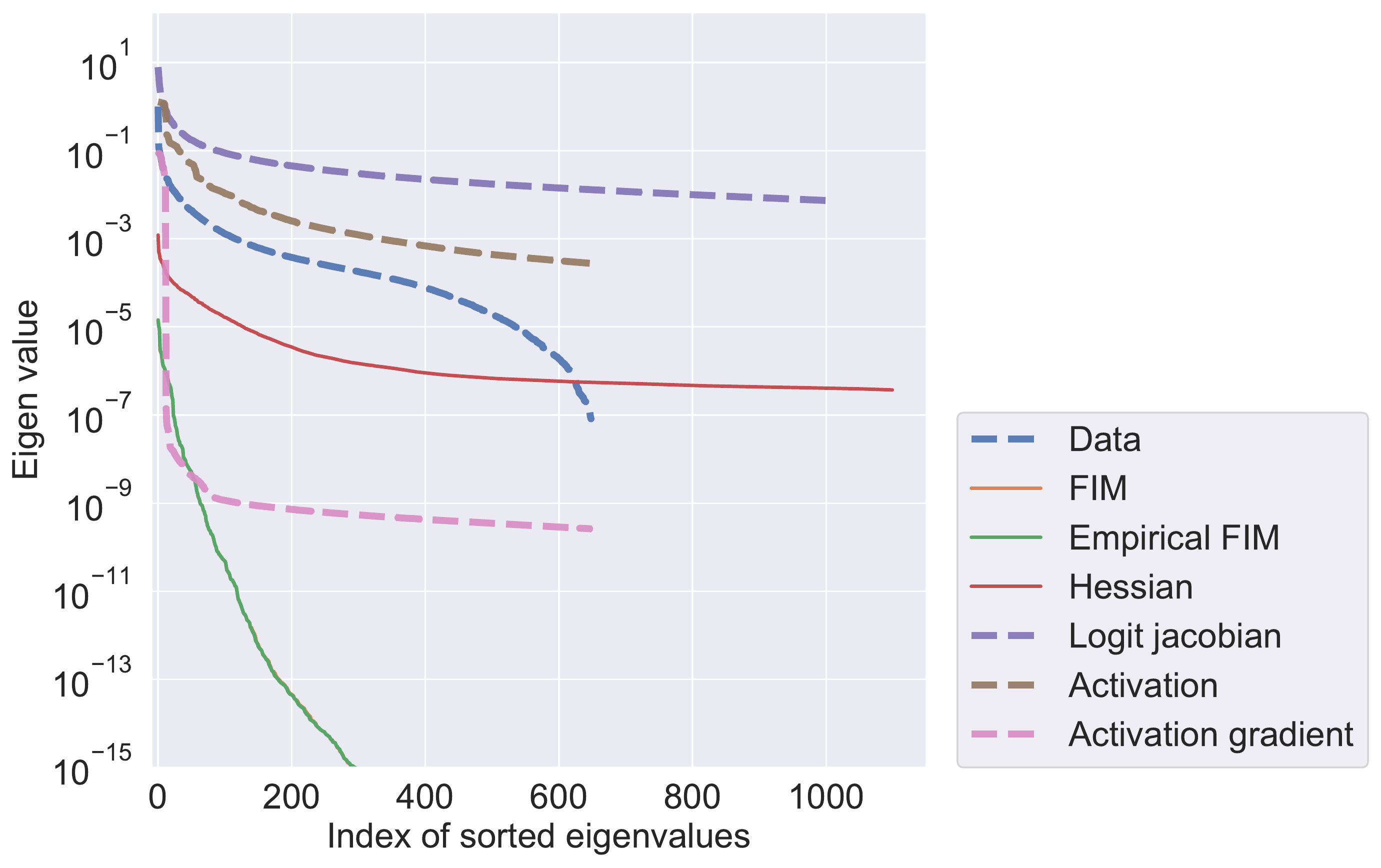}
\caption{
Eigenspectra for a two-layer fully-connected network on MNIST. The eigenspectra are qualitatively the same as those of~\cref{fig:intro}, e.g., there is a sharp drop at the beginning and a long, linear tail of small eigenvalues follows. Slopes of the eigenspectra of activations, activation gradients, Jacobians and Hessian mirror those of the data. In contrast to~\cref{fig:intro}, the slope of the FIM is quite different here. The Empirical FIM and FIM overlaps with each other since the model is trained to nearly perfect train and validation error.
}
\label{fig:eig_mnist_fc_600_2}
\end{figure}

MNIST in spite of its lower dimensionality has roughly the same range of eigenvalues but it has a very small threshold $r$ in~\cref{def:sloppy} which indicates that data has a lower number of effective dimensions than CIFAR-10. The FIM (empirical FIM is essentially the same line) shows a very strong decay for MNIST; since the trace of the FIM has been used as an indicator of the information stored in the weights~\citep{achille2018critical}, this indicates that the weights have to store very little information to predict MNIST well. The Hessian and FIM have very different eigenvalues for MNIST but as~\cref{fig:full_vs_kfac} indicates the two matrices have a larger overlap in their top eigenvectors. Eigenspectra of other networks on MNIST are similar to~\cref{fig:eig_mnist_fc_600_2} while those of CIFAR-10 are similar to~\cref{fig:intro}.

\begin{figure}[htpb]
\centering
\begin{subfigure}[c]{0.4\linewidth}
\centering
\includegraphics[width=\linewidth]{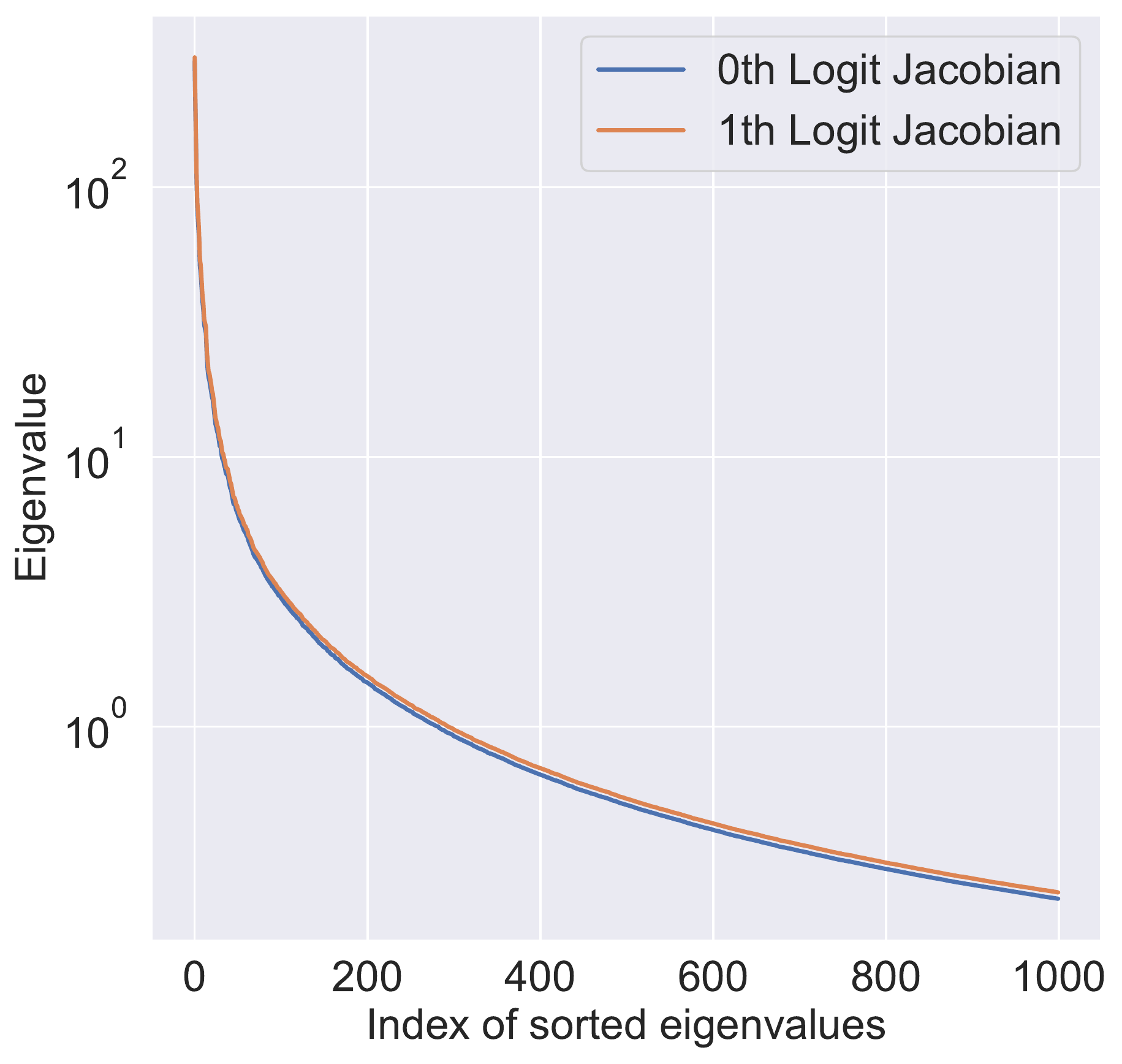}
\end{subfigure}
\begin{subfigure}[c]{0.4\linewidth}
\centering
\includegraphics[width=\linewidth]{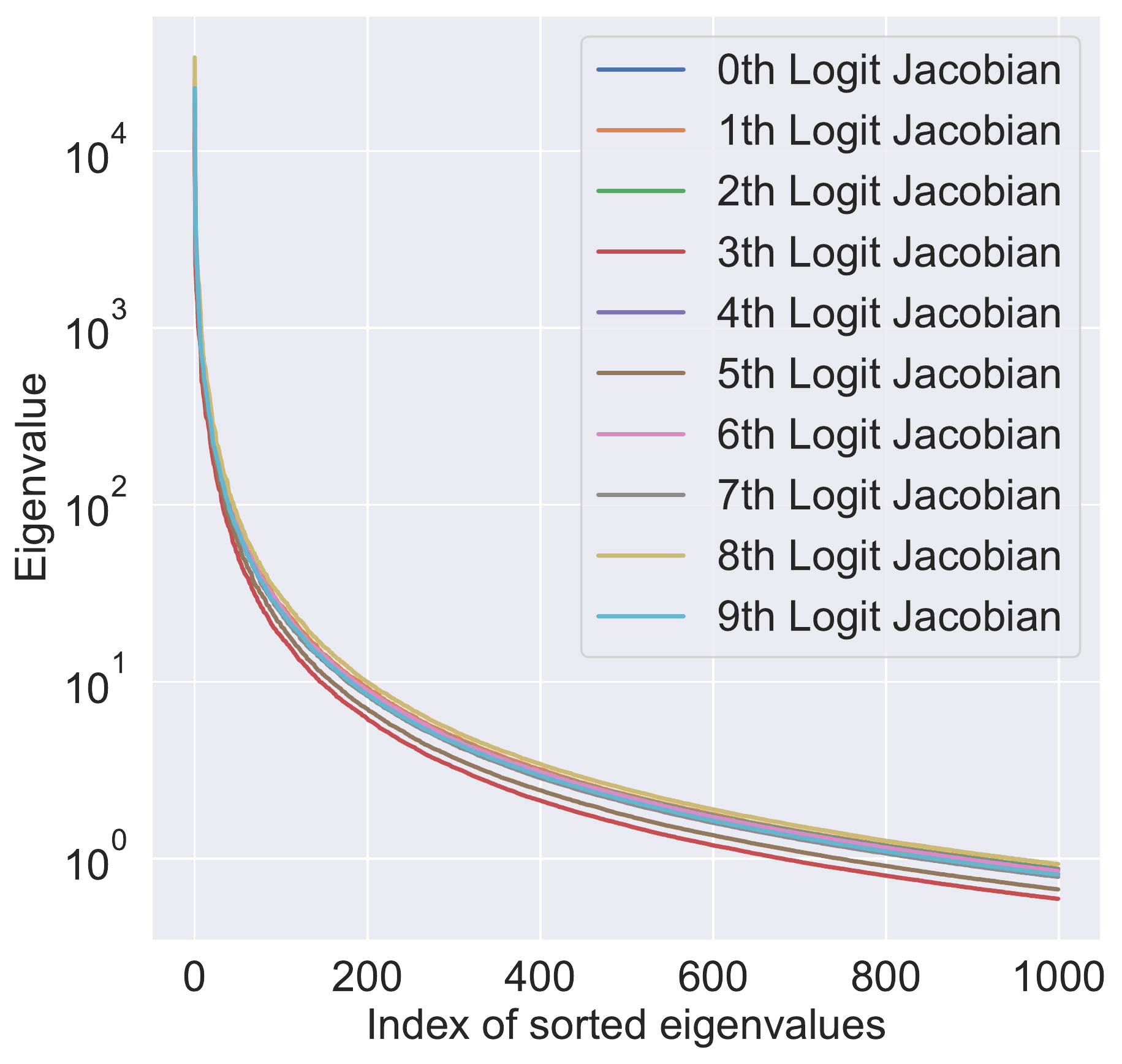}
\end{subfigure}
\caption{Eigenspectra of the correlation matrices of Jacobian of logits for FC-600-2 on MNIST (Left) and wide residual net on CIFAR-10 (Right). The eigenspectra are similar for different logits.}
\label{fig: logit_compare}
\end{figure}

\begin{figure}[htpb]
\centering
\begin{subfigure}[c]{0.4 \linewidth}
\centering
\includegraphics[width=\linewidth]{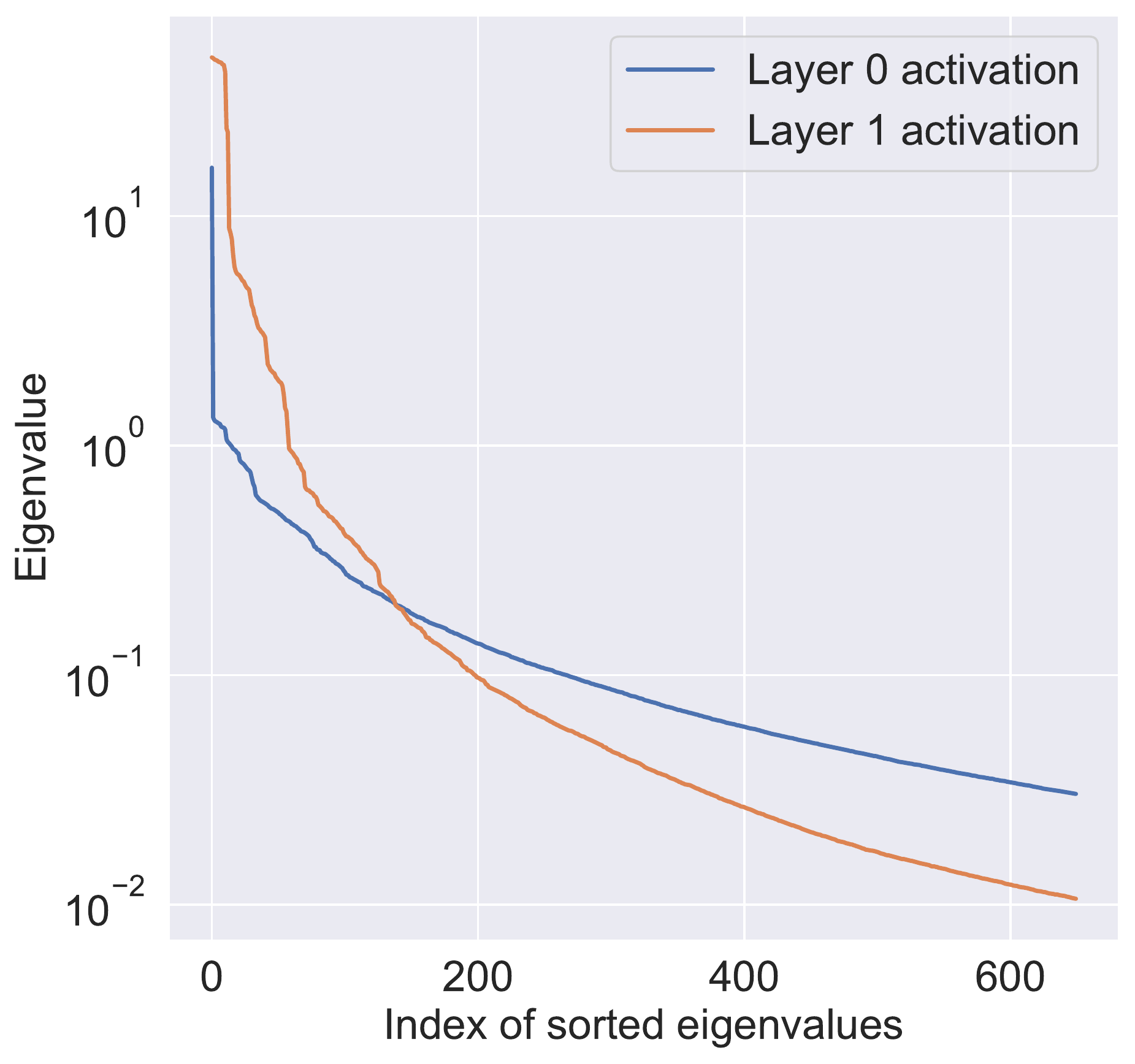}
\end{subfigure}
\begin{subfigure}[c]{0.4 \linewidth}
\centering
\includegraphics[width=\linewidth]{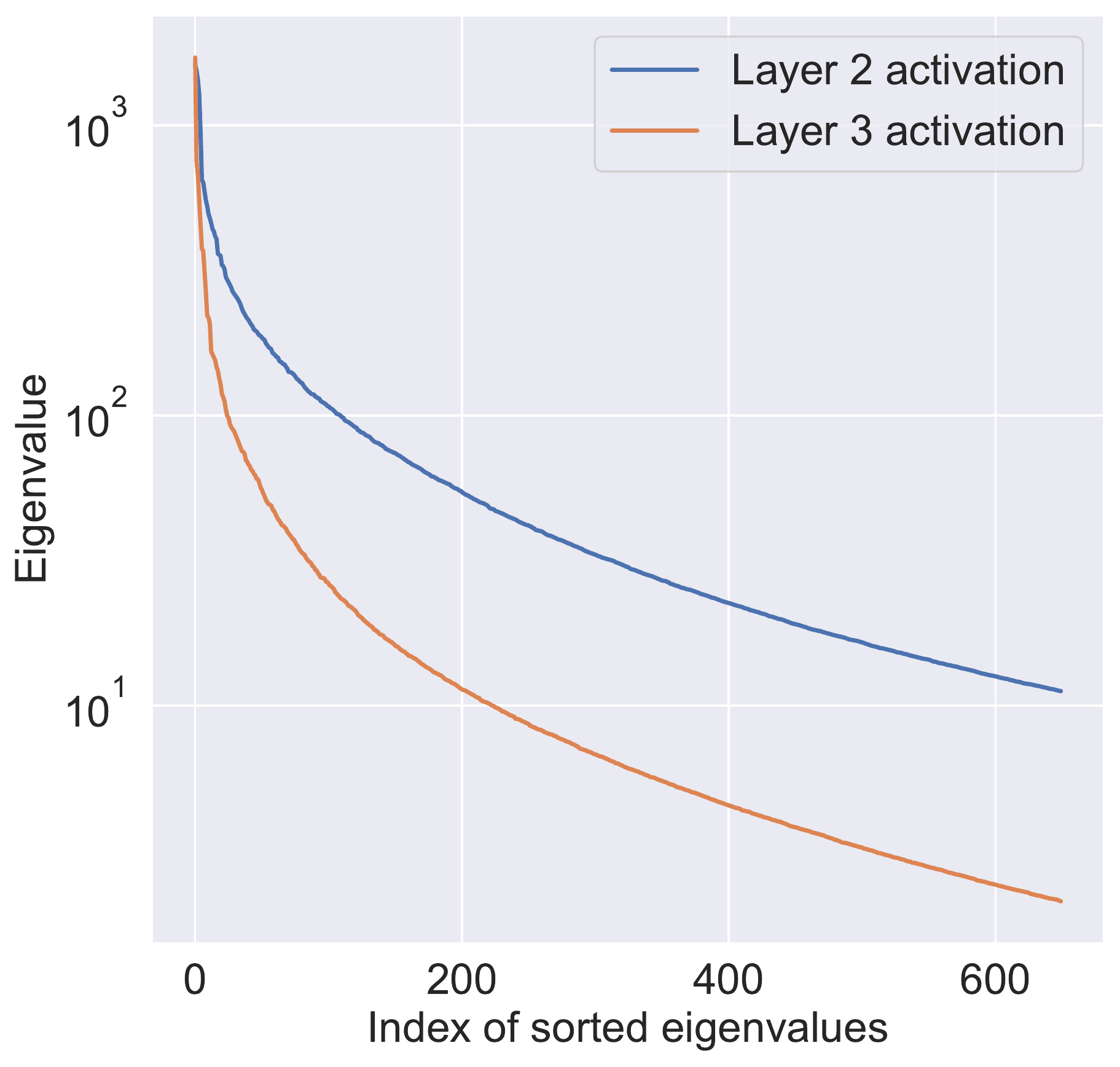}
\end{subfigure}
\caption{Eigenspectra of the correlation matrices of activations of different layers for FC-600-2 on MNIST (Left) and wide residual net on CIFAR-10 (Right). For different layers, the eigenspectra are similar.}
\label{fig: correlation_compare}
\end{figure}

\begin{figure}[htpb]
\centering
\begin{subfigure}[c]{0.4 \linewidth}
\centering
\includegraphics[width=\linewidth]{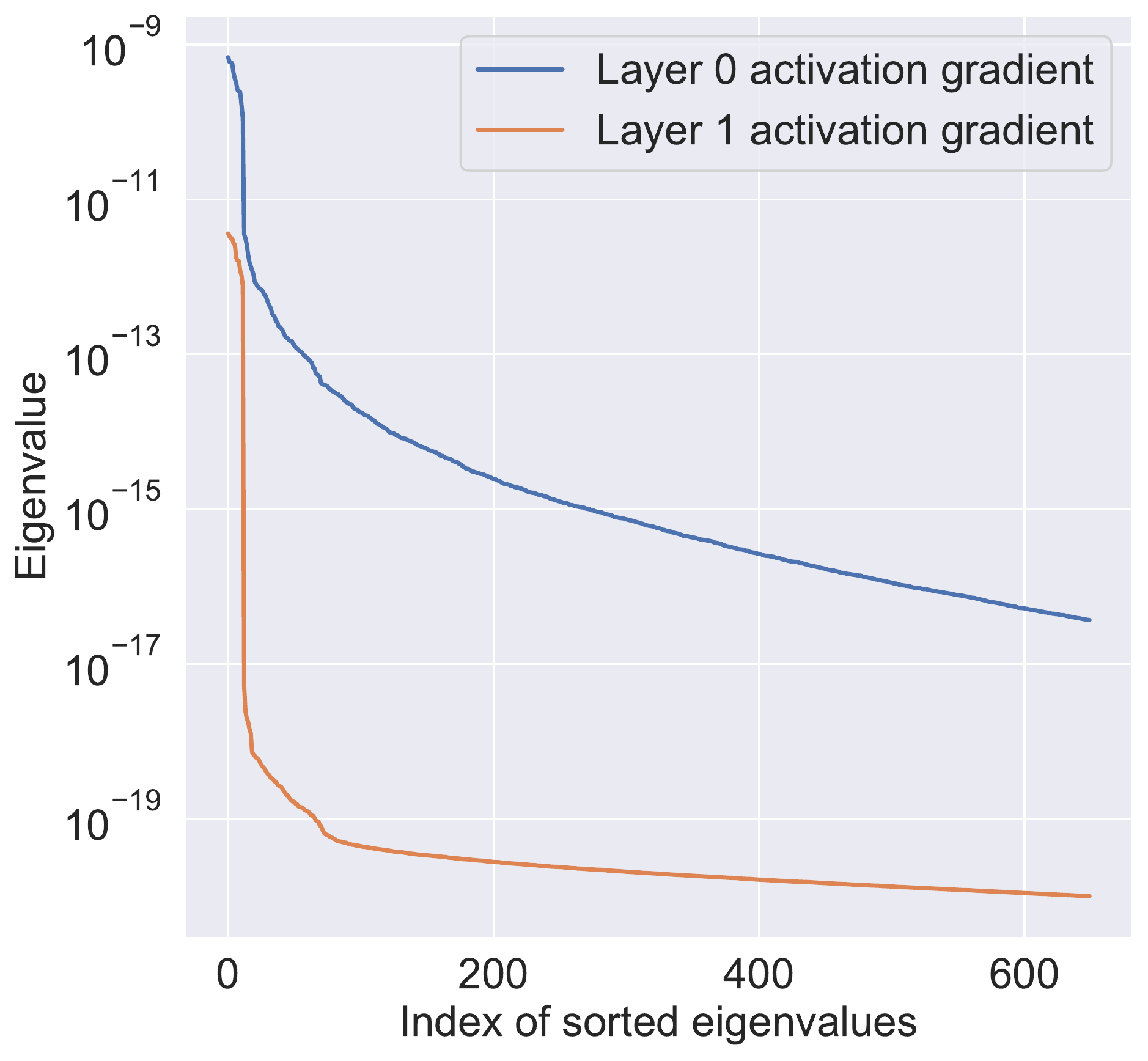}
\end{subfigure}
\begin{subfigure}[c]{0.4 \linewidth}
\centering
\includegraphics[width=\linewidth]{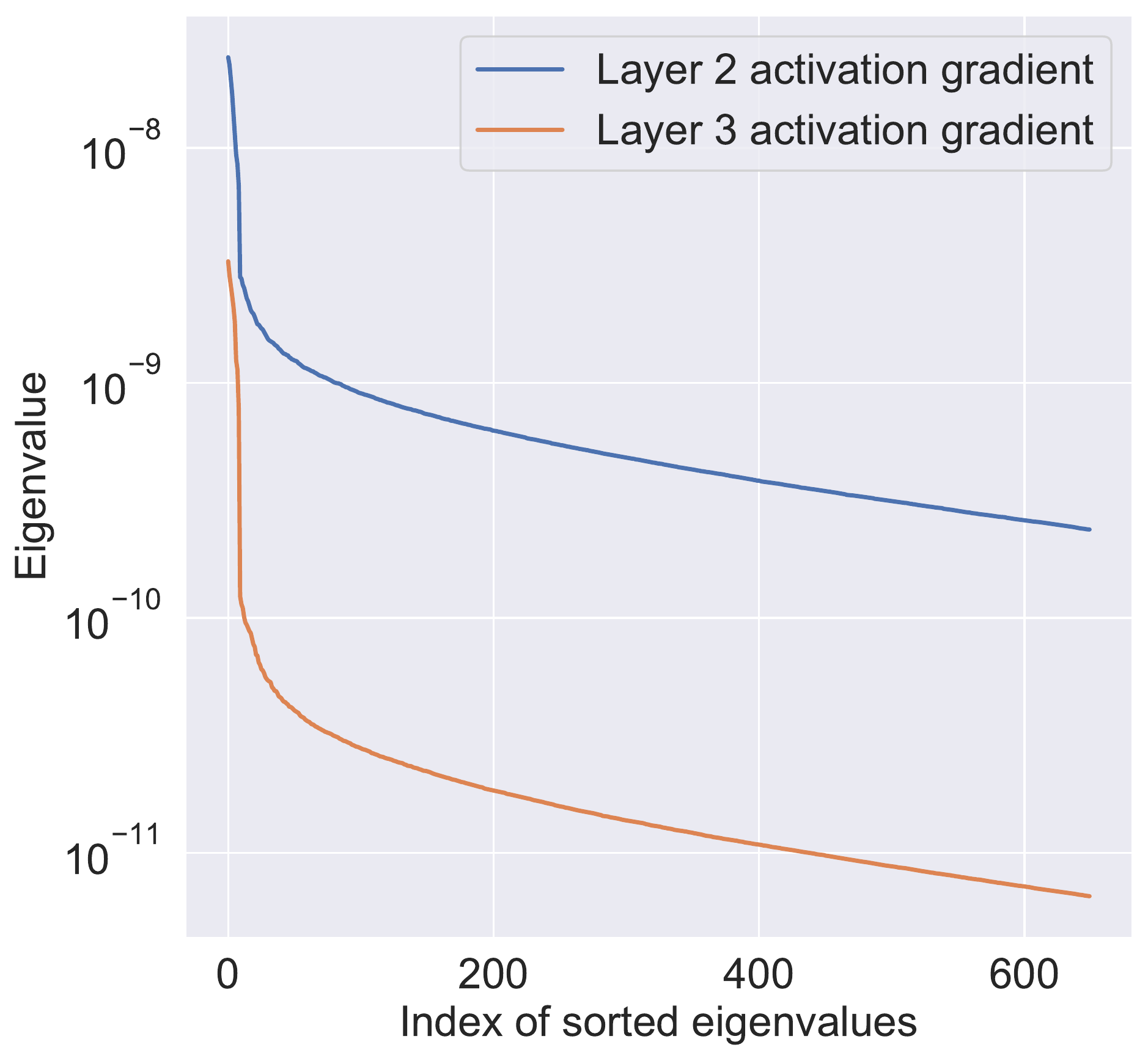}
\end{subfigure}
\caption{Eigenspectra of the correlation matrices of gradients with respect to the activations of different layers for FC-600-2 on MNIST (Left) and wide residual net on CIFAR-10 (Right). For different layers, the eigenspectra are qualitatively similar, and as we move into higher layers of neural networks, the eigenvalues becomes smaller for gradient of activations.}
\label{fig: correlation_grad_compare}
\end{figure}

\begin{figure}[htpb]
\centering
\includegraphics[width=0.5\linewidth]{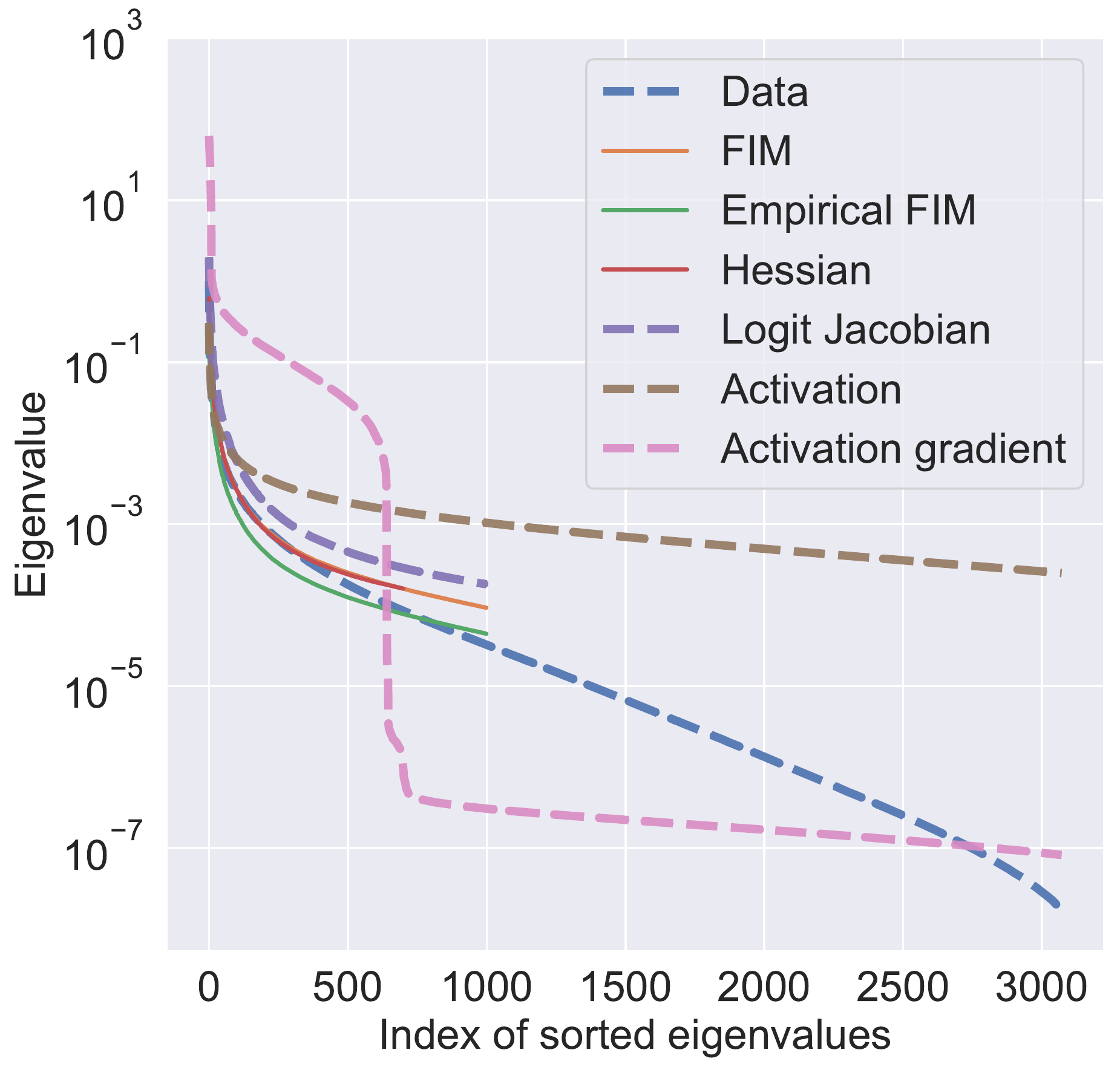}
\caption{Eigenspectra for ALL-CNN on CIFAR-10. The eigenspectra are qualitatively the same as those of ~\cref{fig:intro} for a wide residual network on CIFAR-10.}
\label{fig: eig_all_cnn}
\end{figure}

\begin{figure}[htpb]
\centering
\includegraphics[width=0.6\linewidth]{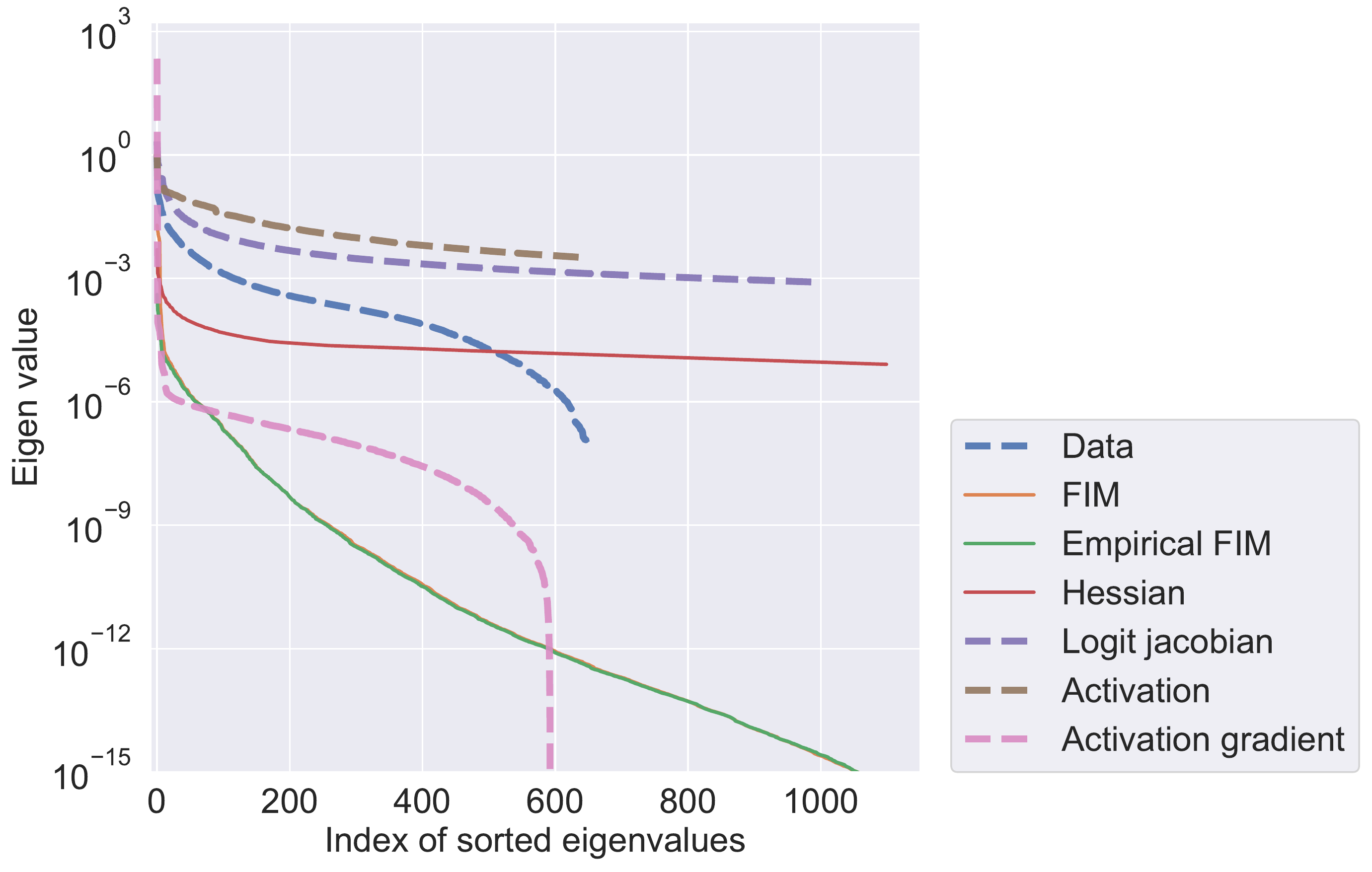}
\caption{Eigenspectra for FC-1200-1 on MNIST. The eigenspectra are qualitatively the same as those of~\cref{fig:eig_mnist_fc_600_2} for FC-600-2 on MNIST.}
\label{fig: eig-fc-1200-1}
\end{figure}

\begin{figure}[htpb]
\centering
\begin{subfigure}[c]{0.45 \linewidth}
\centering
\includegraphics[width=\linewidth]{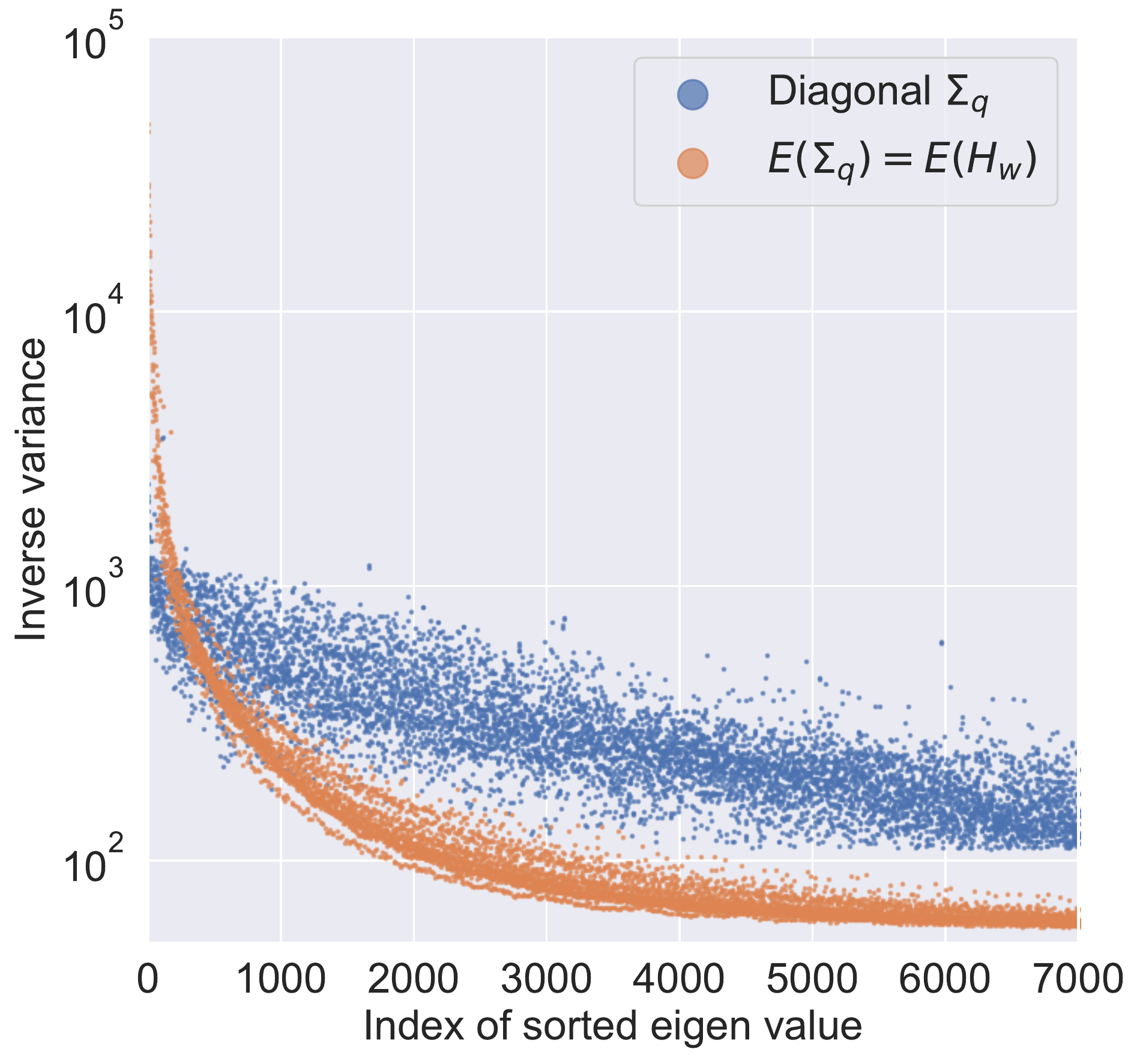}
\end{subfigure}
\begin{subfigure}[c]{0.45 \linewidth}
\centering
\includegraphics[width=\linewidth]{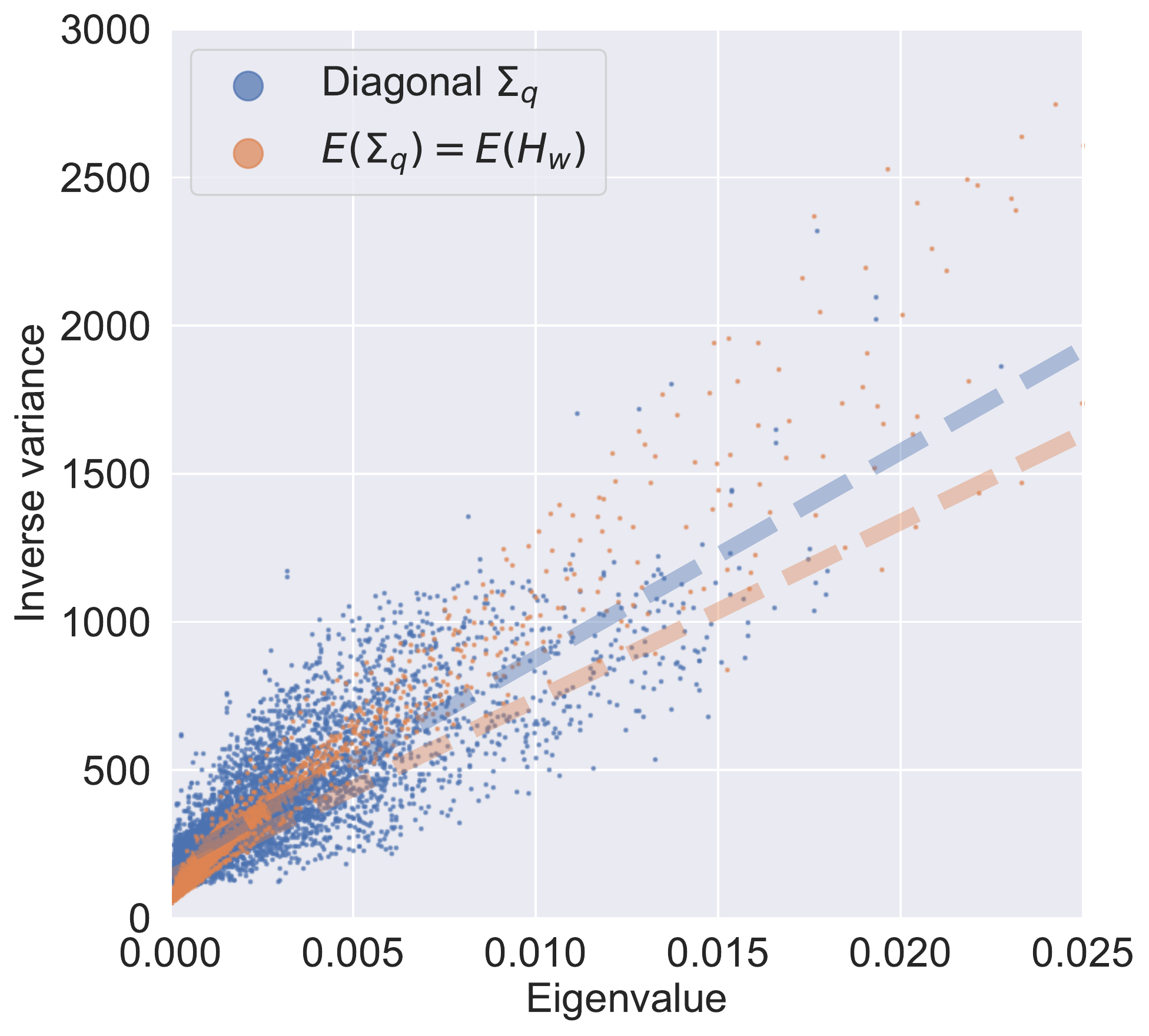}
\end{subfigure}
\caption{\textbf{Posterior covariance computed by optimizing PAC-Bayes bound is aligned with sloppy directions} This plot is a reproduction of~\cref{fig:posterior_scatter} for FC-1200-1 (\cref{fig:posterior_scatter} is for FC-600-2). We get $n \sim 30000$ (true $n$ = 55000) and $\e \sim 138.2$ (true $\e$ = 53.3)}
\label{fig: posterior scatter fc-1200-1}
\end{figure}

In~\cref{fig: logit_compare}, we compare the correlation matrices of logit Jacobian for different logits, which shows that the eigenspectra for different logits are similar. In~\cref{fig: correlation_compare} and~\cref{fig: correlation_grad_compare} we compare the correlation matrices of activations and their gradients. From the figures, we can see that the eigenspectra are similar for different layers, which shows that the sloppiness is preserved as we getting into higher layers of neural network. In \ref{fig: eig_all_cnn} and \ref{fig: eig-fc-1200-1} we ploted the eigenspectra for different networks. The similarity of eigenspectra of matrices calculated on same dataset but different architectures strongly indicates that the sloppiness of Hessian, FIM, correlations of logit Jacobians, activations and gradients of activations are all inherited from the sloppiness of the data set. ~\cref{fig: posterior scatter fc-1200-1} is a reproduction of ~\cref{fig:posterior_scatter} using FC-1200-1 on MNIST.

\subsection{Weights of a trained network can come back towards the initialization in the sloppy subspace even if they evolved in the stiff subspace}
\label{s:app:pruned}

\begin{figure}[htpb]
\centering
\begin{subfigure}[c]{0.4 \linewidth}
\centering
\includegraphics[width=\linewidth]{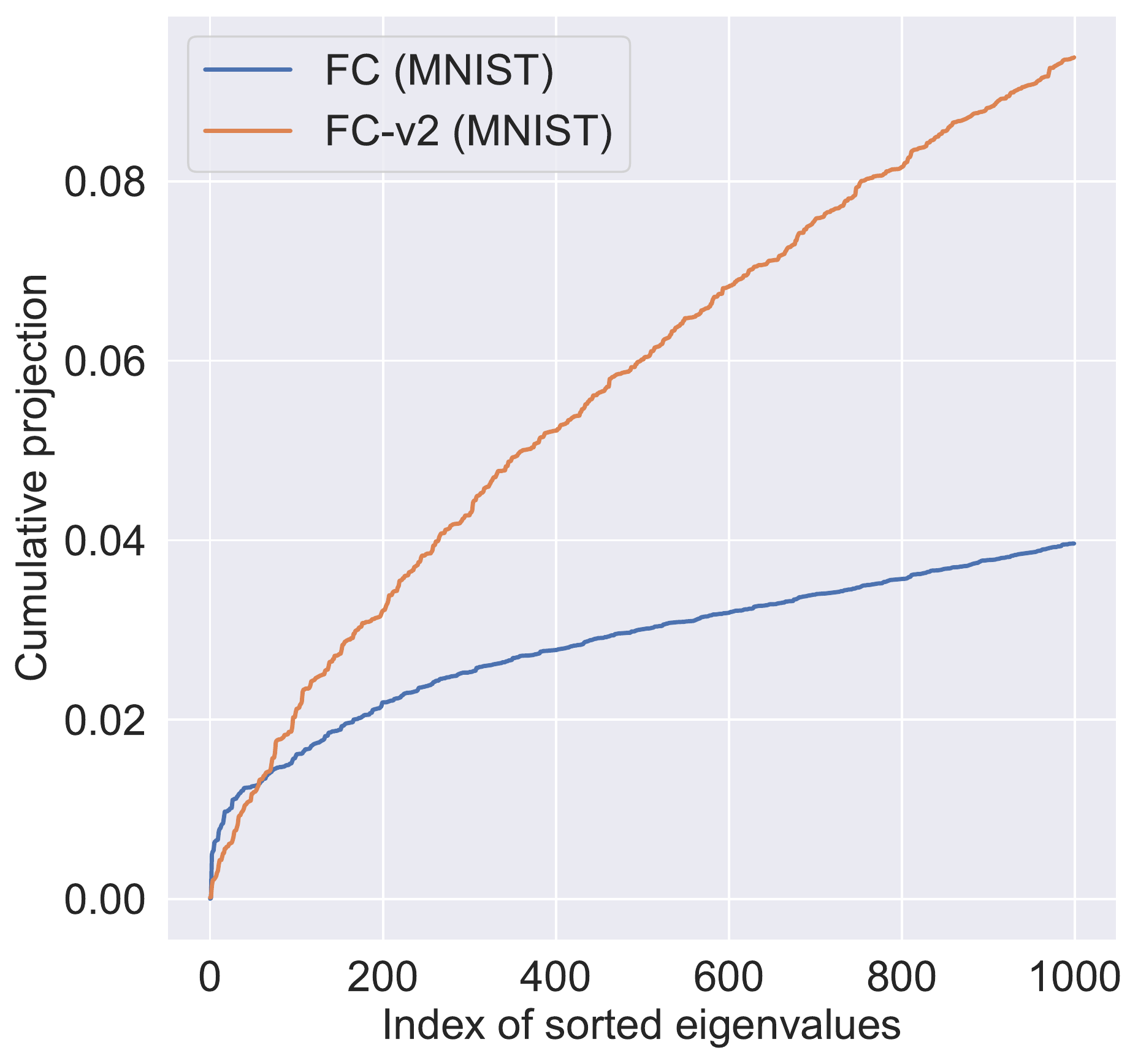}
\end{subfigure}
\begin{subfigure}[c]{0.4 \linewidth}
\centering
\includegraphics[width=\linewidth]{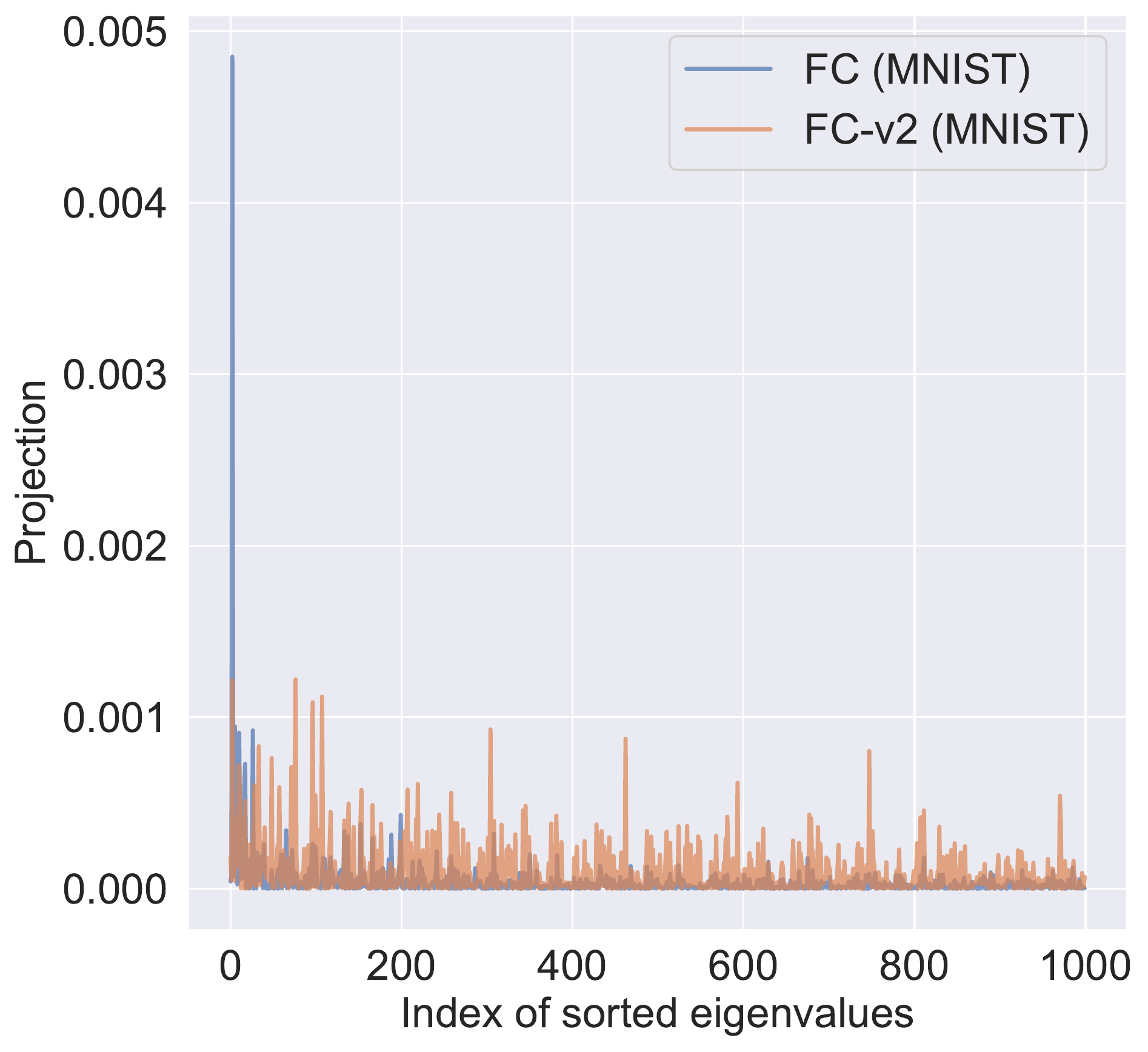}
\end{subfigure}
\caption{(Left) Cumulative projection $\norm{E_k(F_{w_0})\Delta w}_2^2 / \norm{\Delta w}_2^2$ of the change in weights during training (where $\Delta w = w-w_0$) onto the eigenspace of the top $k$th eigenvalues of $F_{w_0}$. (Right) Projection $\norm{\nu_k(F_{w_0})\Delta w}_2^2 / \norm{\Delta w}_2^2$ of the change in weights during training onto the eigenvector $\nu_k(F_{w_0})$ of $k$th largest eigenvalue of $F_{w_0}$, which is the derivative of the curve in the left plot. We use FC-600-2 on MNIST for this experiment.}
\label{fig:pruning}
\end{figure}

\cref{fig:pruning} shows that the projection of change of weights of the model ($w-w_0$) for the v2 model (which has a second phase of training with a penalty $\propto ||w-w_0||_2^2$) onto the stiff directions is larger than that of original model (FC). This indicates that the projection onto the sloppy directions of model v2 is smaller than that of the original model because the projection onto orthogonal decompositions of the parameter space sums to one. This indicates that weights can effectively come back towards the initialization in the sloppy subspace without affecting the accuracy of the model even if the model predominantly evolves in the stiff subspace during training.

\subsection{PAC-Bayes generalization bounds and effective dimensionality for synthetic data sets} \label{s:app:synthetic}
In \cref{tab:pac_bayes_synthetic}, we show the results for PAC-Bayes bound optimization for synthetic data sets introduced in \cref{s:data_hessian_fim_sloppy}. Using the $\epsilon$ derived by Method 3, we calculate the effective dimensionality, strength, and sloppy factor of Hessian at the end of training (showed in the third block of  \cref{tab:pac_bayes_synthetic}), which shows that for non-sloppy data set, we have heavier tails and more stiff directions, resulting in worse generalization.

{
\renewcommand{\arraystretch}{1}
\begin{table}
\centering
\tiny
\rowcolors{1}{}{black!5}
\resizebox{0.5\linewidth}{!}{

\begin{tabular}{lrrr}\toprule
    Quantity/Model & Random-0.1 & Random-0.001 \\
    \midrule
    
\multicolumn{3}{c}{\textbf{Training and validation error of the trained model}}\\[0.25em]
$\hat{e}(h_w, D_n)$ &0.0000 &0.0749 \\
$\log 2 * \Breve{e}(h_w, D_n)$ &0.0869 &0.5745 \\
$e(h_w)$ &0.1150 &0.5035 \\
$\log 2 * \Breve{e}(h_w)$ &0.2983 &1.3797 \\
\midrule

\multicolumn{3}{c}{\textbf{$\ev(\S_q)=\ev(F_{w_0})$ (Method 2)}}\\[0.25em]
$\hat{e}(Q, D_n)$ &0.0155 &0.0117 \\
$\log 2 * \Breve{e}(Q, D_n)$ &0.0509 &0.2368 \\
$e(Q)$ &0.1191 &0.4596 \\
$\log 2 * \Breve{e}(Q)$ &0.3965 &1.2574 \\
PAC-Bayes bound &0.3560 &0.6781 \\
$\KL(Q, P)$ &18468.6914 &53052.6094 \\
\midrule

\multicolumn{3}{c}{\textbf{$\text{E}(\S_q)=\text{E}(H_w)$ (Method 3, our implementation)}}\\[0.25em]
$\hat{e}(Q, D_n)$ &0.0102 &0.0128 \\
$\log 2 * \Breve{e}(Q, D_n)$ &0.0420 &0.2508 \\
$e(Q)$ &0.1133 &0.4614 \\
$\log 2 * \Breve{e}(Q)$ &0.3793 &1.2556 \\
PAC-Bayes bound &0.2986 &0.6769 \\
$\KL(Q, P)$ &15311.6416 &52592.7852 \\
$\epsilon$ & 1.60 & 0.266 \\
$p(n, \epsilon)$ &542 (0.256\%) & 1730 (0.820 \%)\\
$s(n, \epsilon)$ & 2325 & 4962\\
$1/c(n, \epsilon)$ & 209 & 452\\
\midrule

\multicolumn{3}{c}{\textbf{$\S_p = aF_{w_0} + \e^{-1}, \text{E}(\S_q)=\text{E}(F_{w_0})$ (Method 4)}}\\[0.25em]

$\hat{e}(Q, D_n)$ &0.0093 &0.0083 \\
$\log 2 * \Breve{e}(Q, D_n)$ &0.0262 &0.2217 \\
$e(Q)$ &0.1199 &0.5035 \\
$\log 2 * \Breve{e}(Q)$ &0.6493 &1.3797 \\
PAC-Bayes bound &0.2514 &0.6577 \\
$\KL(Q, P)$ &12346.2207 &50933.4063 \\
\midrule

\multicolumn{3}{c}{\textbf{$\text{diag}(\S_q) = \L$ (our implementation)}}\\[0.25em]
$\hat{e}(Q, D_n)$ &0.0094 &0.0083 \\
$\log 2 * \Breve{e}(Q, D_n)$ &0.0275 &0.0452 \\
$e(Q)$ &0.1325 &0.4599 \\
$\log 2 * \Breve{e}(Q)$ &0.6041 &2.1837 \\
PAC-Bayes bound &0.5096 &0.7512 \\
$\KL(Q, P)$ &32949.1836 &66678.6016 \\
\bottomrule
\end{tabular}

}

\caption{\textbf{Comparison of PAC-Bayes bounds on synthetic data sets.} The table shows the PAC-Bayes bound optimization results of the synthetic data sets introduced in \cref{s:data_hessian_fim_sloppy}. The first 3 blocks corresponds to our Methods 2-4 described in ~\cref{s:pac_bayes_bounds}. The 4th block is our reproduction of  \cite{dziugaiteComputingNonvacuousGeneralization2017} on synthetic data sets.}
\label{tab:pac_bayes_synthetic}
\end{table}
}

\subsection{Eigenspectra at the middle of training}
\label{s:app:eig_mid}

\begin{figure}[H]
\centering
\begin{subfigure}[c]{0.4 \linewidth}
\centering
\includegraphics[width=\linewidth]{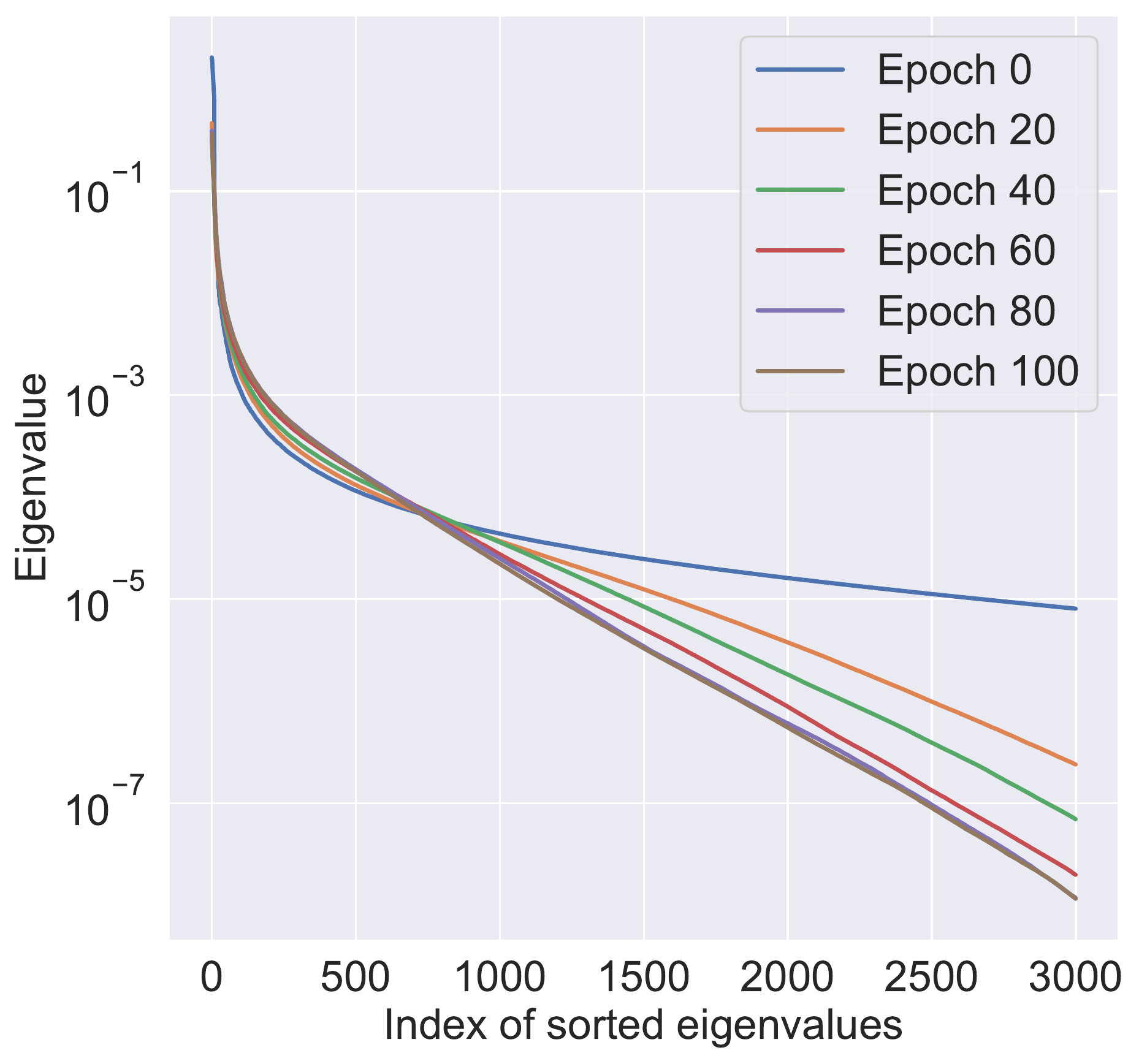}
\end{subfigure}
\begin{subfigure}[c]{0.4 \linewidth}
\centering
\includegraphics[width=\linewidth]{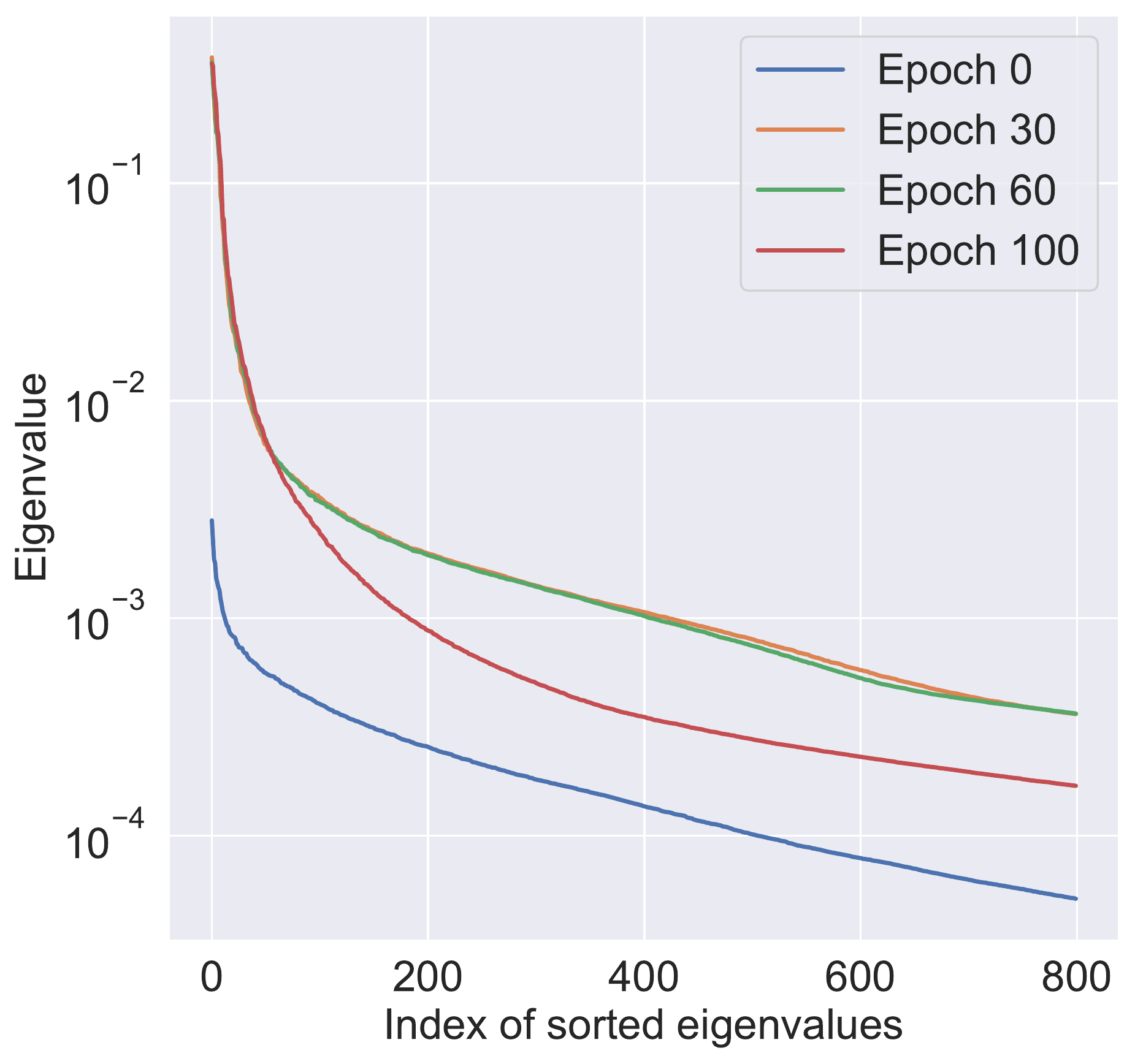}
\end{subfigure}
\caption{(Left) Eigenspectra for FIM of WRN on CIFAR-10 throughout training. The eigenspectrum for epoch 0 are scaled up by $10^3$ to bring it to this scale. (Right) Eigenspectra for Hessian of WRN on CIFAR-10 throughout training. We take the absolute value of the eigenvalues. The eigenspectra are qualitatively the same.}
\label{fig: eig_mid}
\end{figure}

\cref{fig: eig_mid} shows the eigenspectra for FIM and Hessian of a wide residual net on CIFAR-10 during training, which shows that the eigenspectra are qualitatively the same throughout training.

\subsection{Eigenspectrum at the end of training for data sets with random labels}
\label{s:app:eig_random_end}

\begin{figure}[htpb]
\centering
\begin{subfigure}[c]{0.5 \linewidth}
\centering
\includegraphics[width=\linewidth]{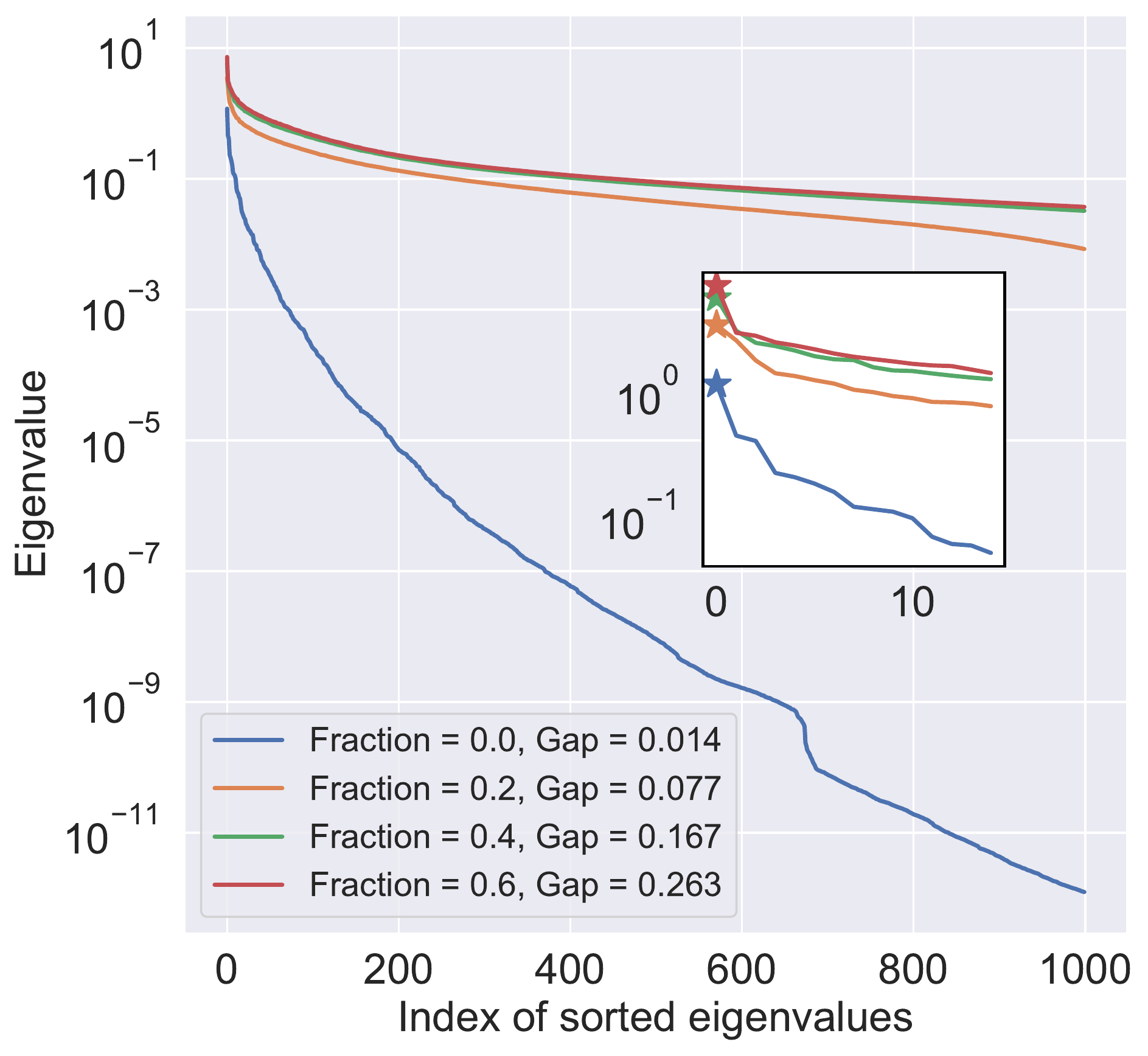}
\end{subfigure}
\caption{\textbf{Eigenspectra at the end of training for data sets with random labels.} The plots shows the eigenspectra of empirical FIM at the end of training. The experiment is done on MNIST using fully connected net FC-600-2. The label "Fraction=$a$" indicates the data set with random label of fraction $a$. The inset plot shows the top 15 eigenvalues. The line for Fraction=0.0 are scaled up by $10^7$. The plots shows that the FIM at the end of training is less sloppy for data sets with random labels, and the top eigenvalues increases as we increase the fraction of random labels.}
\label{fig: random label}
\end{figure}

\cref{fig: random label} shows the eigenspectra of empirical FIM at the end of training, in which the eigenspectra is less sloppy for data sets with random labels, and the top eigenvalues increases as we increase the fraction of random labels. This shows that even if we have the same input data set, the sloppiness and the top few eigenvectors can still be affected by the task. The eigenspectra becomes less sloppy and has larger head when the the model is used to learn more difficult tasks (more random labels).

\end{appendix}
\end{document}